\documentclass[10pt,twocolumn,letterpaper]{article}

\usepackage[pagenumbers]{cvpr} 

\usepackage{graphicx}
\usepackage{amsmath}
\usepackage{amssymb}
\usepackage{booktabs}

\usepackage{placeins}
\usepackage{xfrac}
\usepackage{nicefrac}
\usepackage{array}

\usepackage[pagebackref,breaklinks,colorlinks]{hyperref}

\usepackage[capitalize]{cleveref}
\crefname{section}{Sec.}{Secs.}
\Crefname{section}{Section}{Sections}
\Crefname{table}{Table}{Tables}
\crefname{table}{Tab.}{Tabs.}

\def\cvprPaperID{5932} 
\def\confName{CVPR}
\def\confYear{2022}

\begin{document}

\title{Rotation Equivariant 3D Hand Mesh Generation from a Single RGB Image}

\author{Joshua Mitton\\
University of Glasgow\\
Glasgow, UK\\
{\tt\small j.mitton.1@research.gla.ac.uk}
\and
Chaitanya Kaul\\
University of Glasgow\\
Glasgow, UK\\
{\tt\small chaitanya.kaul@glasgow.ac.uk}
\and
Roderick Murray-Smith\\
University of Glasgow\\
Glasgow, UK\\
{\tt\small roderick.murray-smith@glasgow.ac.uk}
}
\maketitle

\begin{abstract}
   We develop a rotation equivariant model for generating 3D hand meshes from 2D RGB images. This guarantees that as the input image of a hand is rotated the generated mesh undergoes a corresponding rotation. Furthermore, this removes undesirable deformations in the meshes often generated by methods without rotation equivariance. By building a rotation equivariant model, through considering symmetries in the problem, we reduce the need for training on very large datasets to achieve good mesh reconstruction.
   
   The encoder takes images defined on $\mathbb{Z}^{2}$ and maps these to latent functions defined on the group $C_{8}$. We introduce a novel vector mapping function to map the function defined on $C_{8}$ to a latent point cloud space defined on the group $\mathrm{SO}(2)$. Further, we introduce a 3D projection function that learns a 3D function from the $\mathrm{SO}(2)$ latent space. Finally, we use an $\mathrm{SO}(3)$ equivariant decoder to ensure rotation equivariance. Our rotation equivariant model outperforms state-of-the-art methods on a real-world dataset and we demonstrate that it accurately captures the shape and pose in the generated meshes under rotation of the input hand.
\end{abstract}

\section{Introduction}
\label{sec:intro}
Vision based models for 3D hand mesh generation are currently receiving a lot of interest, as they facilitate a wide range of applications including virtual reality (VR), augmented reality (AR), and sign language recognition. Currently, two main approaches exist to tackle the difficulties of 3D mesh generation from 2D RGB images. One approach uses deformable models which exploit a prior distribution in 3D shape shape. These models can be trained on relatively small datasets, although typically they struggle to generalise to a diverse range of hand meshes. As such, they often do not accurately generate hand meshes of correct shape and pose. The second approach is to use neural network models, which are more general and do not exploit a 3D shape space prior. The neural network model is generally in the form of a convolutional encoder and graph or mesh neural network decoder. These models are able to more accurately predict the correct shape and pose of the hands, although these models currently require large amounts of data to train. This often requires the use of synthetic data as a pre-training step. Despite this, use of synthetic data often results in a model that does not generalise well to real-world examples. On the other hand, collecting large datasets of real-world data is expensive and time consuming. The lack of robustness presented from using synthetic data or the expense of gathering real-world data may limit the number of applications of the approach. 

In this work we overcome the issue of requiring large amounts of data by building a neural network model with a suitable inductive bias. We quantitatively demonstrate the effectiveness of this approach on real-world data, outperforming previous state-of-the-art methods. Further, we qualitatively demonstrate the benefit of our choice of inductive bias and compare to alternative models. We follow the approach of using a neural network based decoder, rather than a deformable model with 3D shape prior, to overcome the issue with generalisation. In addition to building a model than can generalise, we utilise suitable inductive biases in each component of the model. This improved weight sharing in the model reduces the need for pre-training on large synthetic datasets and generalises well training only on a small real-world dataset. The resulting model is rotation equivariant, which is a suitable inductive bias for generating 3D meshes from RGB images due to the symmetries present in the data. To the best of our knowledge this is the first work proposing rotation equivariance for the generation of 3D meshes from RGB images. 

Previous approaches use a convolutional encoder and graph based decoder to generate meshes. We introduce a rotation-equivariant convolutional encoder and an $\mathrm{SO}(3)$-equivariant decoder. In addition, we introduce two new novel components into the framework for 3D mesh generation from RGB images which we term a vector mapping function and 3D projection function. Instead of encoding images into a latent vector where no structure is enforced, as is done in previous works, we introduce a vector mapping function which maintains the rotation equivariance enforced in the encoder while projecting the data into a latent vector space. This guarantees that a rotation of the input image maps to a rotation of the latent space and therefore enforces some structure in the latent space. Furthermore, we introduce a 3D projection function that maintains the rotation equivariance enforced in the prior components of the model by utilising an $\mathrm{SO}(2)$ equivariant map. This maps from a vector representation of the group $\mathrm{SO}(2)$, $\rho_{1}$, to the direct sum of a vector and scalar representation of the group $\mathrm{SO}(2)$, $\rho_{1} \oplus \rho_{0}$. This produces a third inferred dimension which we treat as the third dimension of a 3D space and feeds into the $\mathrm{SO}(3)$-equivariant decoder. 

Current leading models use some form of graph neural network as the decoder. Using a graph neural network (GNN) model enforces locality in the model as the features of each node in the graph are updated from the features of nodes within some neighbourhood. Current approaches choose a small sized neighbourhood for the update functions, although do not consider a larger neighbourhood. This locality is lost if the graph chosen is a fully connected graph. Current methods allow longer range dependency between nodes in the graph by stacking multiple GNN layers. No attention appears to be paid to the impact of this choice and whether there is a benefit to using larger neighbourhoods during update functions or even fully connected graphs. Further, it can be noted that using an MLP layer instead of fully connected GNN could be seen as an alternative approach to allowing long range dependency on node features. In this work we compare models with both an MLP and GNN decoder to explore the benefit of using a decoder with locality as an inductive bias. While this highlights some advantages of using a GNN decoder, there are also some disadvantages and this motivates us to build a MLP based model with rotational equivariance.


\section{Related Work}
\label{sec:relwork}

Equivariance in neural networks has been used for many years in the form of translation equivariance \cite{lecun1989backpropagation}, which led to many successes with image data \cite{krizhevsky2012imagenet}. More recently, equivariance has been considered for rotations of images. Initially this was achieved by looking at group actions for the discrete groups $p4$ and $p4m$ of rotations by $90^{\circ}$ and also reflections \cite{cohen2016group}. The theory of steerable CNNs describes rotation and reflection equivariant convolutions on the image plane $\mathbb{R}^{2}$ \cite{cohen2016steerable, weiler20183d, cohen2018intertwiners, cohen2018general, cohen2019gauge}. Steerable CNNs have a feature space that is defined as a space of feature fields, which is characterised by a group representation. This group representation defines the transformation law of the feature space under transformation of the input. A range of group equivariant convolution neural networks have been proposed \cite{cohen2016group, weiler2018learning, hoogeboom2018hexaconv, bekkers2018roto, dieleman2016exploiting, kondor2018generalization, weiler2019general}, which consider equivariance to the group $\mathrm{E}(2)$ and its subgroups. It has been shown that rotation equivariance is a useful choice of inductive bias in a range of applications where there is a known symmetry of the input domain \cite{weiler2019general, bekkers2018roto, veeling2018rotation, mitton2021rotation, gupta2021rotation}.

Similarly to the developments of equivariance for image data on the plane $\mathbb{R}^{2}$, equivariance has been considered for graphs and point clouds in 3D space. In general for 3D point clouds or graphs the group considered is that of 3D rotations, $\mathrm{SO}(3)$, 3D rotations and translations, $\mathrm{SE}(3)$, or 3D rotations, reflections and translations, $\mathrm{E}(3)$ \cite{thomas2018tensor, fuchs2020se3transformers, satorras2021n, kohler2019equivariant, schutt2017schnet}. The group representations for $\mathrm{SO}(3)$ are orthogonal matrices that can be decomposed as $\rho (g) = Q^{T} \left [ \oplus_{l} D_{l} (g) \right ] Q$, where $Q$ is a change-of-basis matrix and $D_{l}$ is a Wigner-D matrix \cite{fuchs2020se3transformers}. In work on $\mathrm{E}(3)$ equivariant models the message passing graph neural network \cite{gilmer2017neural} is modified such that edge updates use the squared distance and position is updated using a vector field in a radial direction to guarantee equivariance to the group $\mathrm{E}(n)$. In addition to 3D point clouds, an equivariance constraint can be placed over a network for 2D rotational groups, such as $\mathrm{SO}(2)$ or $\mathrm{SE}(2)$ \cite{finzi2020generalizing}. 

Following this, work has been conducted on a more general method of constructing equivariant models \cite{lang2020wigner, ravanbakhsh2017equivariance}. However these methods still require considerable mathematical work limiting there general application. On the other hand, a general framework for constructing rotation reflection equivariant convolutional networks for the group $\mathrm{E}(2)$ including subgroups was presented in \cite{weiler2019general}. Further, a general framework for solving equivariance constraints for a range of different groups was presented for a multilayer perceptron (MLP) based architecture in \cite{finzi2021practical}.

GNNs were first introduced as a means to learn on graph structured data using neural networks where the data is irregularly structured \cite{gori2005new, scarselli2008graph, li2015gated, duvenaud2015convolutional}. GNNs become more widely used when they were developed to scale better with the size of input graph \cite{kipf2016semi, velivckovic2017graph, bronstein2017geometric, gilmer2017neural, defferrard2016convolutional}. Since then GNNs have been used to learn about graphs \cite{chen2020simple, simonovsky2017dynamic, morris2019weisfeiler, xu2018powerful, mitton2021graph}, point clouds \cite{wu2019pointconv, hertz2020pointgmm, zhao2021point, fuchs2020se3transformers, li2019point}, and meshes \cite{feng2019meshnet, hanocka2019meshcnn, hanocka2020point2mesh}. Graph- mesh- and point-convolutions are relevant here as they can be used in the decoder as an inductive bias to build into the model. They provide the inductive bias of locality in the model, meaning that features are updated only based on neighbouring nodes. On the other hand, using an MLP does not provide any locality inductive bias in the model, as the features at every node are updated based on the features of every other node.

Generating meshes from 2D images has seen increased attention in recent years \cite{chatzis2020comprehensive}. Using a convolutional encoder is the approach taken in all prior works to embed the image into a latent space, with differing choices of convolutional architecture. The main difference between prior works comes in the form of the decoder. One approach is to use a deformable model which has been fit to landmarks, exploiting a prior distribution in 3D shape space \cite{baek2019pushing, boukhayma20193d, hasson2019learning, zhang2019end, kulon2019single}. This has the advantage that the model will generate realistic looking hands. On the other hand, this approach has a weakness that the model will tend to generate hands only from the prior distribution and therefore the generated meshes may not accurately reflect the input image. Another approach is to use more general models as the decoder with no prior distribution guiding the model to generate hands. One such method uses a ResNet-50 encoder and spacial mesh convolutional decoder \cite{kulon2020weakly}. Another method uses a stacked hourglass convolutional encoder and graph convolutional decoder utilising Chebyshev polynomials \cite{ge20193d}. While these approaches more accurately generate hand meshes with the correct shape and pose of the input image, they generally require large amounts of synthetic data as a pre-training step. This is not advantageous as it requires significant effort in creating a synthetic dataset. In addition, these methods are not robust at generating real-world meshes due to the reliance on large datasets.


\section{Methodology}
\label{sec:method}

\subsection{Overview}
Our work follows the general structure of previous works on learning to generating 3D hand meshes from RGB images in that it uses an encoder to embed the image into a latent space and a decoder to generate the 3D meshes. Unlike previous works we consider the symmetries in the problem and create a rotation equivariant encoder and decoder. In addition, we add two new modules a vector mapping function and a 3D projection function. The vector mapping function maps from a latent image space on the group $C_{8}$ to an $\mathrm{SO}(2)$ representation space. The 3D projection function maps from an $\mathrm{SO}(2)$ vector space to an $\mathrm{SO}(2)$ vector space and scalar space which allows us to introduce the third inferred dimension of the 3D space in a controlled way. By ensuring rotation equivariance at each stage of the model a 2D rotation of the input image corresponds to a 2D rotation of the output mesh about the third axis. Ensuring rotation equivariance within the model reduces the dependency on very large data sets and removes undesirable deformations of the mesh under rotation of the image. 

The input to the model is RGB images of hands and the model outputs predicted meshes of the hand. The decoder predicts the point positions of a mesh, where we follow previous work \cite{ge20193d, kulon2020weakly} and output the points of the mesh in a fixed order which makes use of a predefined topology. In this work, we train directly on a small dataset of real-world data and do not have any expensive pre-training step on synthetic data.

\subsection{Graphs}
The state-of-the art prior methods utilise a GNN decoder with node neighbourhoods chosen to be relatively small in comparison to the entire set of vertices in the output mesh. No consideration is given to the neighbourhood size used in the graph decoder and whether a global update could improve upon the local updates of a GNN. A MLP could be viewed as a fully connected GNN update with separate update on each edge. The main drawback of an MLP is the fixed size outputs of the model and in general GNNs are used due to the variable size of the data. Here when generating 3D meshes the output is of fixed size and hence one benefit of using a GNN is moot. The remaining advantage of using a GNN over an MLP is that the GNN has an inductive bias of locality. We experimentally compare using a GNN and MLP decoder in Table~\ref{tab:MLPvGNN} and note that the MLP decoder achieves a lower vertex and Laplacian loss indicating the MLP model learns to generate more accurate and smoother meshes. We further look at the validation mesh error for both the GNN and MLP decoders in Table~\ref{tab:mesherror}. This shows that the MLP produces a lower mesh error on the fixed orientation images, but a larger mesh error on the rotated hands. This indicates that the inductive bias of locality marginally reduces over fitting over the MLP model. The GNN decoder therefore predicts less accurate and less smooth meshes. As we seek to build the inductive bias of rotational equivariance into our model, which aims to improve the generalisation of the vertex position prediction, we conclude that an MLP decoder is desirable given its improved performance. 

\subsection{Encoder}
\begin{figure}[h]
  \centering
  \includegraphics[width=\linewidth]{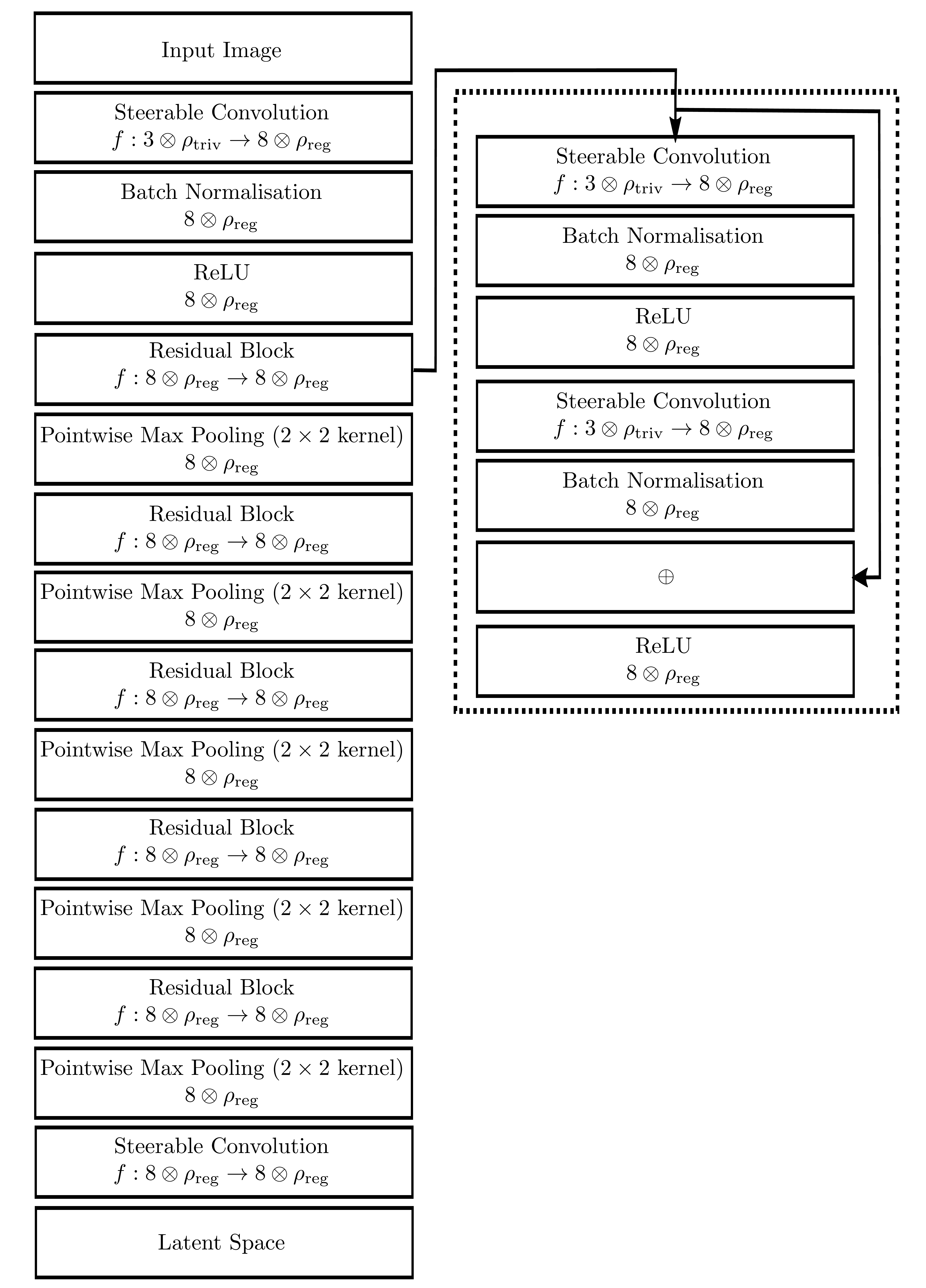}
  \caption{Diagram of the encoder inside the model. This shows the representation space used at each layer in the model and the total number of layers. The right hand column details the components of the residual block, while the left hand column shows the entire structure of the encoder.}
  \label{fig:convexpl}
\end{figure}

For the encoder we use a steerable convolutional neural network (CNN) model which is translation-rotation equivariant. We enforce equivariance to the group $C_{8}$, the group of rotations by $45^{\circ}$, and use the regular representation, $\rho_{\mathrm{reg}}$, to define the transformation law of the feature fields in the network. The steerable CNN has a feature space of steerable fields $f : \mathbb{R}^{2} \rightarrow \mathbb{R}^{8}$ which associate an $8$-dimensional vector $f(x) \in \mathbb{R}^{8}$ with each point $x$ in the base space. The transformation law of the feature fields in the convolutional model is given by $f(x) := \rho(g) \cdot f(g^{-1} (x-t))$, which says that it transforms the feature fields by moving the feature vectors from $g^{-1} (x-t)$ to their new position $x$ and acts on them by $\rho(g)$ which permutes the order of the feature vector depending on the group element $g$. For the encoder we define a residual block, which comprises of two steerable convolution layers, two batch normalisation layers \cite{ioffe2015batch}, and two ReLU layers \cite{nair2010rectified} with skip connection \cite{he2016deep}. The encoder comprises of a steerable convolution, batch normalisation, and ReLU layer followed by five residual blocks and final steerable convolution. The encoder is detailed in Figure~\ref{fig:convexpl} and Figure~\ref{fig:model-a}.

\subsection{Vector Mapping}
The vector mapping layer converts latent functions defined on the group $C_{8} \rtimes \mathbb{Z}_{n^{2}}$ to a latent vector space on the group $\mathrm{SO}(2)$. We achieve this by first inverting the action of the group on the base space, which, noting that the transformation law of the feature fields in the encoder is given by $f(x) := \rho(g) \cdot f(g^{-1} (x-t))$, involves applying the transformation $f(x) := f(g(x))$, where the group action is $g \in C_{8}$. We then interpret the latent space as a 2D latent space of point vectors, which is implicitly done in prior works, and apply the corresponding group action from the group $\mathrm{SO}(2)$. This is done by using representations of the group $\rho(g) \in \mathbb{R}^{2 \times 2}$. Finally, we apply a permutation invariant function as we specify that the output graph should be of fixed topology. This permutation invariant function ensures the topology of the output graph is not dependent on the permutations of the encoder's transformation law. In tensor notation the vector mapping function maps from $T_{1}^{C_{8} \rtimes \mathbb{Z}_{n^{2}}} \rightarrow T_{1}^{\mathrm{SO}(2)}$. The vector mapping function is detailed in Figure~\ref{fig:model-b}.

\subsection{3D Projection}
The 3D projection function takes a latent vector space as the input, where the input representations are representation of the group $\mathrm{SO}(2)$, in tensor notation $T_{1}^{\mathrm{SO}(2)}$. This function maps to both a vector representation and scalar representation of the group $\mathrm{SO}(2)$, in tensor notation $T_{1}^{\mathrm{SO}(2)} \oplus T_{0}^{\mathrm{SO}(2)}$. Therefore, the 3D projection function maps from $T_{1}^{\mathrm{SO}(2)} \rightarrow T_{1}^{\mathrm{SO}(2)} \oplus T_{0}^{\mathrm{SO}(2)}$. We treat the scalar representation as the third inferred dimension of the 3D mesh and reshape the output into a 3D vector space. The 3D projection function is detailed in Figure~\ref{fig:model-c}.

\subsection{Decoder}
\begin{figure}[h]
  \centering
  \includegraphics[width=0.478\linewidth]{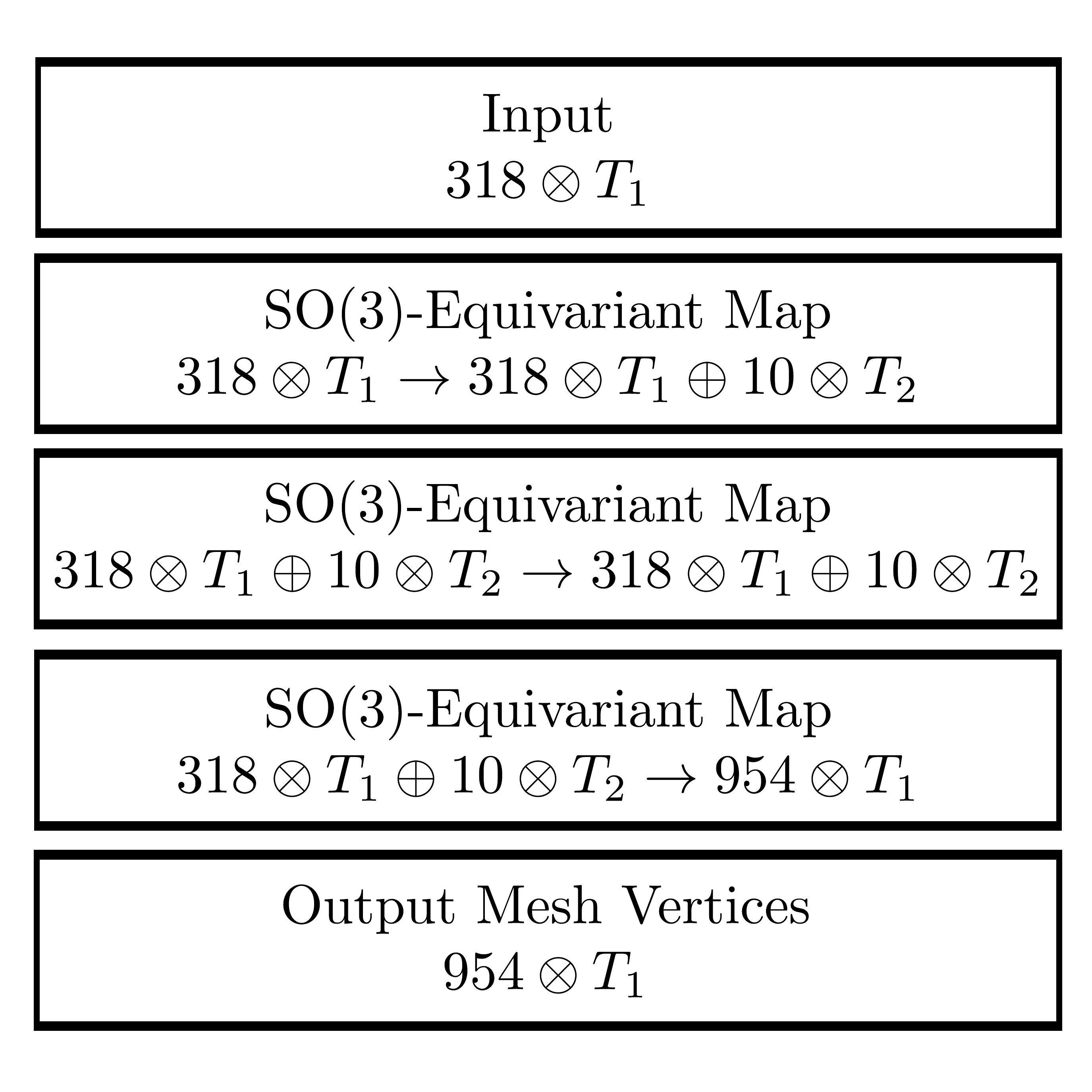}
  \caption{Diagram of the decoder used inside the model. This shows the representation space used at each layer in the decoder.}
  \label{fig:decexpl}
\end{figure}

\begin{figure*}
  \centering
  \begin{subfigure}{1.0\linewidth}
    \includegraphics[width=\linewidth]{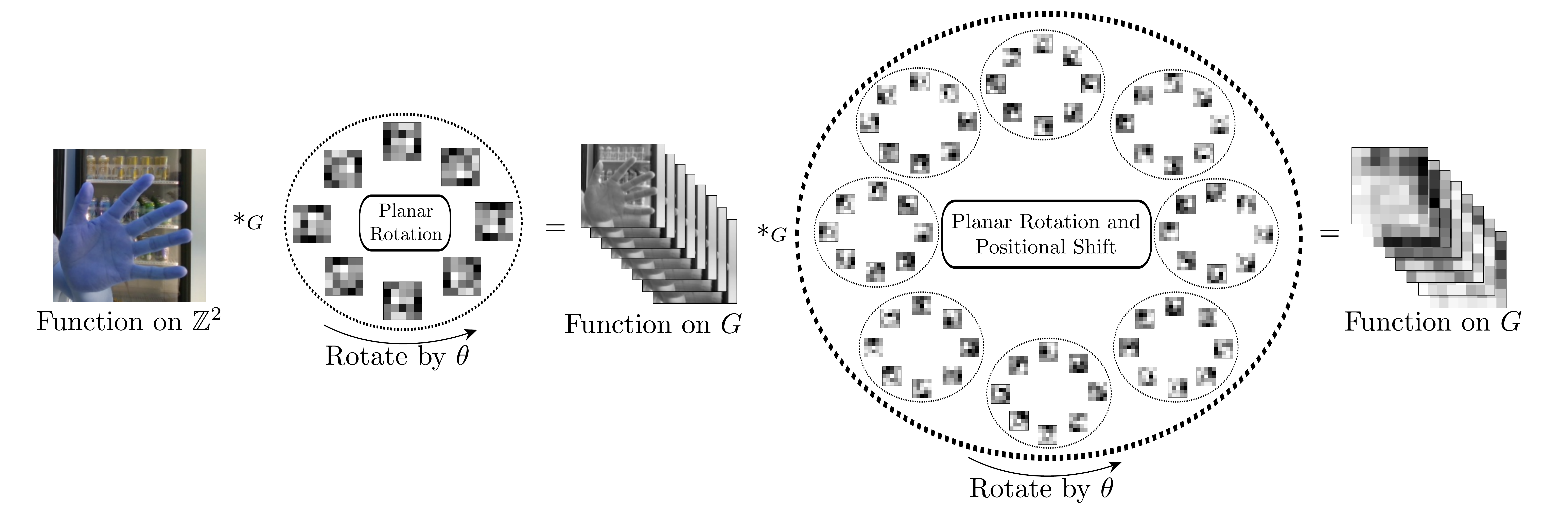}
    \vspace{-20pt}
    \caption{Encoder.}
    \label{fig:model-a}
  \end{subfigure}
  \begin{subfigure}{0.849\linewidth}
    \vspace{-0.5cm}
    \includegraphics[width=\linewidth]{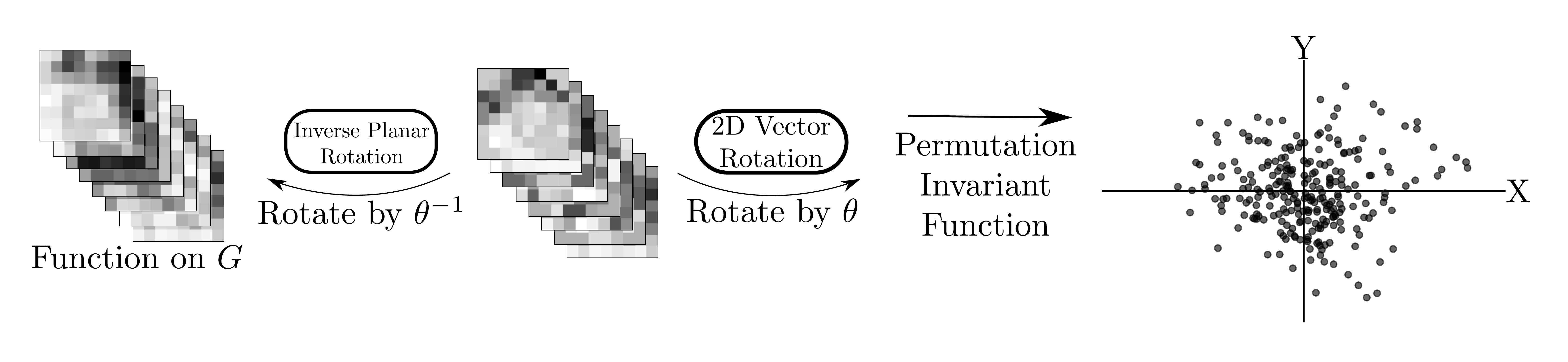}
    \vspace{-20pt}
    \caption{Vector Mapping.}
    \label{fig:model-b}
  \end{subfigure}
  \begin{subfigure}{0.645\linewidth}
    \vspace{-0.5cm}
    \includegraphics[width=\linewidth]{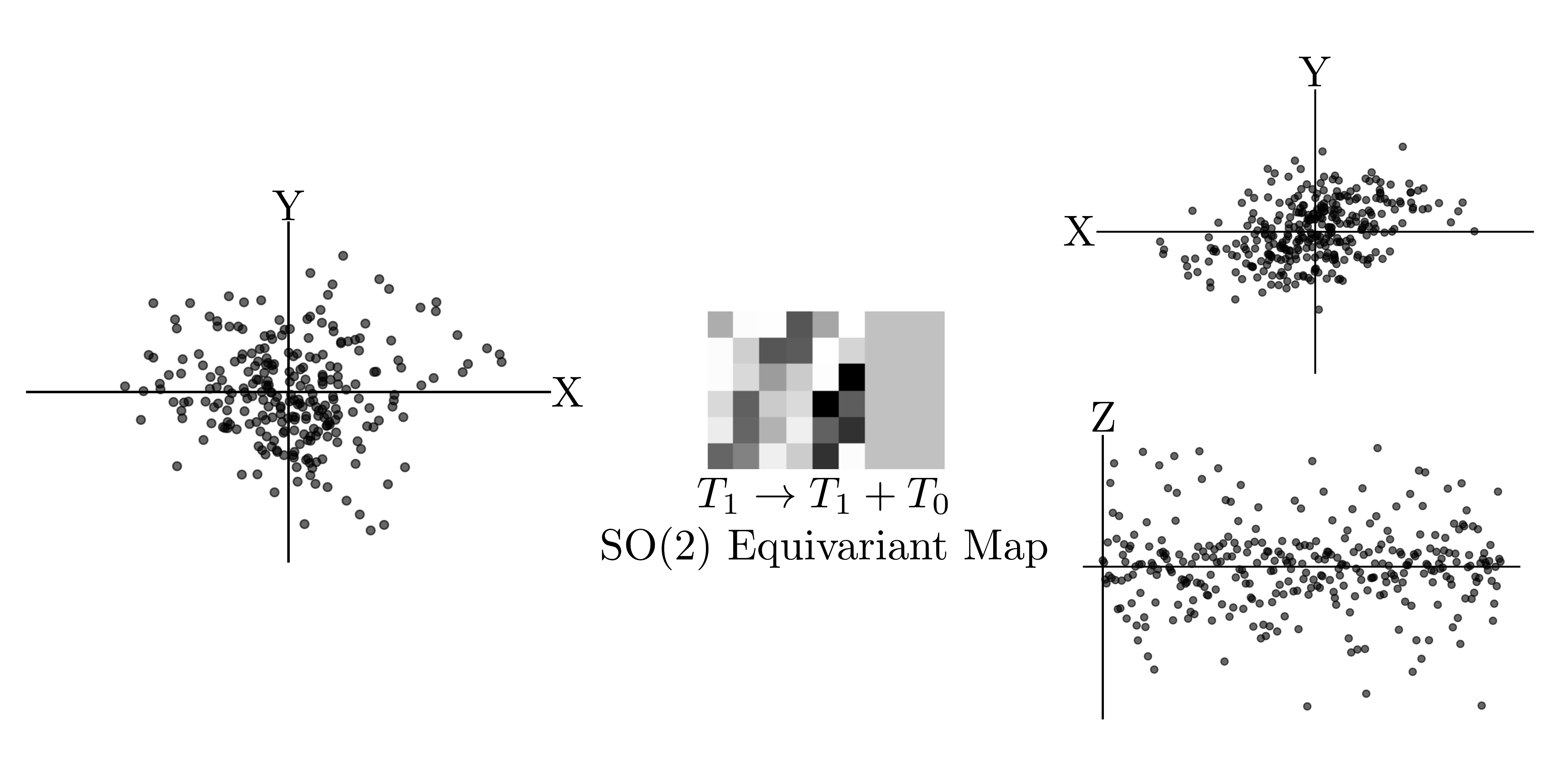}
    \vspace{-25pt}
    \caption{3D Projection.}
    \label{fig:model-c}
  \end{subfigure}
  \begin{subfigure}{0.774\linewidth}
    \vspace{-0.8cm}
    \includegraphics[width=\linewidth]{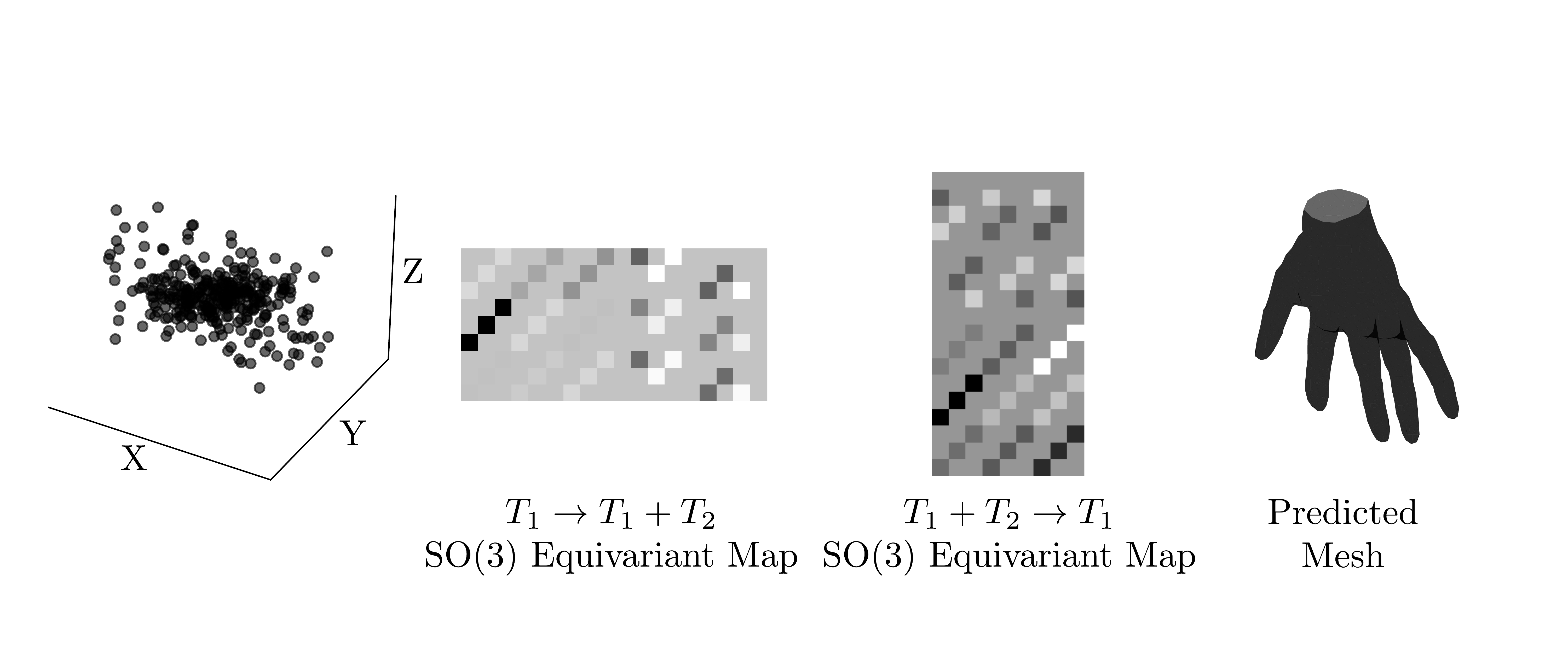}
    \vspace{-20pt}
    \caption{Decoder.}
    \label{fig:model-d}
  \end{subfigure}
  \caption{The 3D hand mesh generation model can be broken down into four key components, namely an encoder, vector mapping, 3D projection, and decoder. (a) The encoder is a rotation equivariant convolutional neural network that takes as input 2D RGB images of hands and outputs a latent space that comprises of functions on the group. This means that a rotation of the input image corresponds to a rotation and permutation of the latent space. (b) The vector mapping function takes as input the latent functions and outputs 2D vector spaces. (c) The 3D projection function takes as input 2D vector spaces and outputs both a 2D vector space and a 1D vector, which are interpreted as the X-Y and Z dimensions in 3D space respectively. This function uses representations of the group $\mathrm{SO}(2)$ to map from a vector representation, $T_{1}$, to both a vector representation, $T_{1}$, and scalar representation, $T_{0}$. (d) The decoder is a $\mathrm{SO}(3)$ equivariant function that maps from 3D points to 3D points, where the output is the hand mesh.}
  \label{fig:model}
\end{figure*}

The decoder maps the latent function, which is a vector representation of the group $\mathrm{SO}(3)$, into the vertex points of the output mesh such that the decoder is $\mathrm{SO}(3)$-equivariant. The decoder comprises of three layers mapping between representations of the group $\mathrm{SO}(3)$ with an output space of $954V^{1}$, which reflects that the output mesh has $954$ position vectors in 3D space as the vertex set. The architecture of the decoder is presented in Figure~\ref{fig:decexpl} and Figure~\ref{fig:model-d}.

\subsection{Training}
The loss function consists of a vertex reconstruction loss and a Laplacian loss. The vertex reconstruction loss is a mean squared error loss and enforces that points in the predicted mesh are close to points in the true mesh:

\begin{equation}
  \mathcal{L}_{v} = \dfrac{1}{N} \sum^{N}_{i=1} \left \lVert v^{\mathrm{3D}}_{i} - \hat{v}^{\mathrm{3D}}_{i} \right \rVert^{2}_{2},
  \label{eq:vertexloss}
\end{equation}

where $v^{\mathrm{3D}}_{i}$ and $\hat{v}^{\mathrm{3D}}_{i}$ are the ground truth and predicted vertex locations respectively. The Laplacian loss is introduced to preserve the local surface smoothness of the mesh:

\begin{equation}
    {{\mathcal{L}}_{l}} = \dfrac{1}{N} \sum\nolimits_{i = 1}^N {\left\| {{{\delta} _i} - {{\sum\nolimits_{{{v}_k} \in {\mathcal{N}}\left( {{{v}_i}} \right)} {{{\delta} _k}} } \mathord{\left/
				{\vphantom {{\sum\nolimits_{k \in N\left( {{v_i}} \right)} {{{\delta} _k}} } {B_i}}} \right.
				\kern-\nulldelimiterspace} {B_i}}} \right\|_2^2},
    \label{laplaceloss}
\end{equation}

where $\delta_{i} = v^{\mathrm{3D}}_{i} - \hat{v}^{\mathrm{3D}}_{i}$ is the offset between the ground truth and predicted vertex locations, $\mathcal{N}\left( {{v}_i} \right)$ is the set of neighbouring vertices of $v_{i}$, and $B_{i}$ is the number of vertices in the set $\mathcal{N}\left( {{v}_i} \right)$. We use hyperparameters $\lambda_{v}=1$ and $\lambda_{l}=10$ in our implementation.

The model is trained with the Adam optimizer for $1000$ epochs with an initial learning rate of $10^{-5}$, decayed by a factor of $0.5$ every $100$ epochs. We do not use data augmentation as in general the typical data augmentation applied to images does not have a corresponding augmentation on the mesh. Therefore, using data augmentation where the input data is defined on the group $\mathbb{Z}^{2}$ and the output data is defined on the group $\mathrm{SO}(3)$ is ill-defined. Further, ignoring this and applying data augmentation to the input images, such as rotations, without applying the corresponding rotation to the output mesh will guide the learning process of the model to being invariant to such transformations, which is undesirable. We train on a Titan Xp with a batch size of 8 with input image size of $256 \times 256$. 

\section{Experiments}
\label{sec:exper}

We experiment on the real-world dataset from \cite{ge20193d} which we split into $500$ training examples of images and meshes, and $83$ validation examples of images and meshes. We report the vertex and Laplacian losses and the mesh error as evaluation metrics, where the mesh error is the average error in Euclidean space between corresponding vertices in each generated 3D mesh and its ground truth 3D mesh.

We first experiment with the choice of decoder in the model by considering an MLP and GNN decoder. As presented in Table~\ref{tab:MLPvGNN}, the MLP decoder achieves lower vertex and Laplacian loss indicating it is predicting more accurate and smoother meshes. This result indicates that the inductive bias of locality introduced by using a GNN decoder has a negative impact upon the accuracy and smoothness of generated meshes, while updating each vertex position based on the positions of all other vertices through the use of an MLP generates more accurate and smoother meshes. We also compare to a rotation equivariant model, EMLP. The EMLP model performs worse than the MLP model for the vertex loss, but outperforms the GNN model. Despite the worse vertex prediction accuracy of the EMLP model, it achieves almost comparable performance to the MLP model for Laplacian loss in Table~\ref{tab:MLPvGNN} indicating it is predicting equally smooth meshes. The better performance of the MLP model is due to the model having more trainable parameters than the other two models and being unconstrained through the addition of no inductive bias, which allows the model to over-fit to the training data.

\begin{table}[h]
  \centering
  \begin{tabular}{p{0.12\linewidth} p{0.29\linewidth} p{0.33\linewidth}}
    \toprule
    Method & vertex loss $\mathcal{L}_{v}$ & Laplacian loss $\mathcal{L}_{l}$ \\
    \midrule
    MLP  & $1e^{-6}$ & $1e^{-6}$ \\
    GNN  & $1e^{-5}$ & $4e^{-5}$ \\
    EMLP  & $5e^{-6}$ & $3e^{-6}$ \\
    \bottomrule
  \end{tabular}
  \caption{Training loss split into vertex loss $\mathcal{L}_{v}$ and Laplacian loss $\mathcal{L}_{l}$ for an MLP and GNN decoder in a non-rotation equivariant model and a EMLP model that has rotation equivariance build into the entire model for the real-world dataset \cite{ge20193d}.}
  \label{tab:MLPvGNN}
\end{table}

Secondly, we consider three different models two non-rotation equivariant models, named MLP and GNN, and a rotation equivariant model, named EMLP. All three models have an identical training scheme and they are tested on the $83$ validation examples. As presented in Table~\ref{tab:rotations} the model without rotation equivariance using an MLP decoder achieves the best validation mesh error on the fixed orientation validation dataset. The GNN decoder performs more poorly indicating worse generalisation to the validation dataset. Further, the rotation equivariant model performs worse still. This result is due to the constraints placed over the GNN and EMLP models via a choice of inductive bias, which results in them having less trainable parameters in the models. In addition to comparing to the fixed orientation validation set, which is likely to contain hands in similar orientations to the training dataset, we consider a rotated orientation validation dataset, where the hands and meshes are correspondingly rotated. Evaluation on this validation dataset highlights the failure of the non-rotation equivariant models, where the rotation equivariant EMLP model drastically outperforms the other methods. Overall, the consistency of the rotation equivariant model across all validation datasets and the improved generalisability results in it outperforming other models. This indicates our choice our choice of model for generating 3D meshes is beneficial and rotation equivariant is a useful inductive bias to build into the model. 

\begin{table}[h]
  \centering
  \begin{tabular}{p{0.12\linewidth} p{0.29\linewidth} p{0.33\linewidth}}
    \toprule
    Method & Fixed orientation mesh error (mm) & Rotated orientation mesh error (mm) \\
    \midrule
    MLP  & 5.33 & 143.10 \\
    GNN  & 6.02 & 138.5 \\
    EMLP & 9.67 & 10.41 \\
    \bottomrule
  \end{tabular}
  \caption{Average mesh error tested on the validation set of the real-world dataset \cite{ge20193d} in the column fixed orientation mesh error. Also, a rotated version of this validation dataset in the column rotated orientation mesh error, where the rotations used are $90^{\circ}$, $180^{\circ}$, and $270^{\circ}$.}
  \label{tab:rotations}
\end{table}

\begin{figure*}
  \centering
  \begin{subfigure}{0.09\linewidth}
    \includegraphics[width=\linewidth]{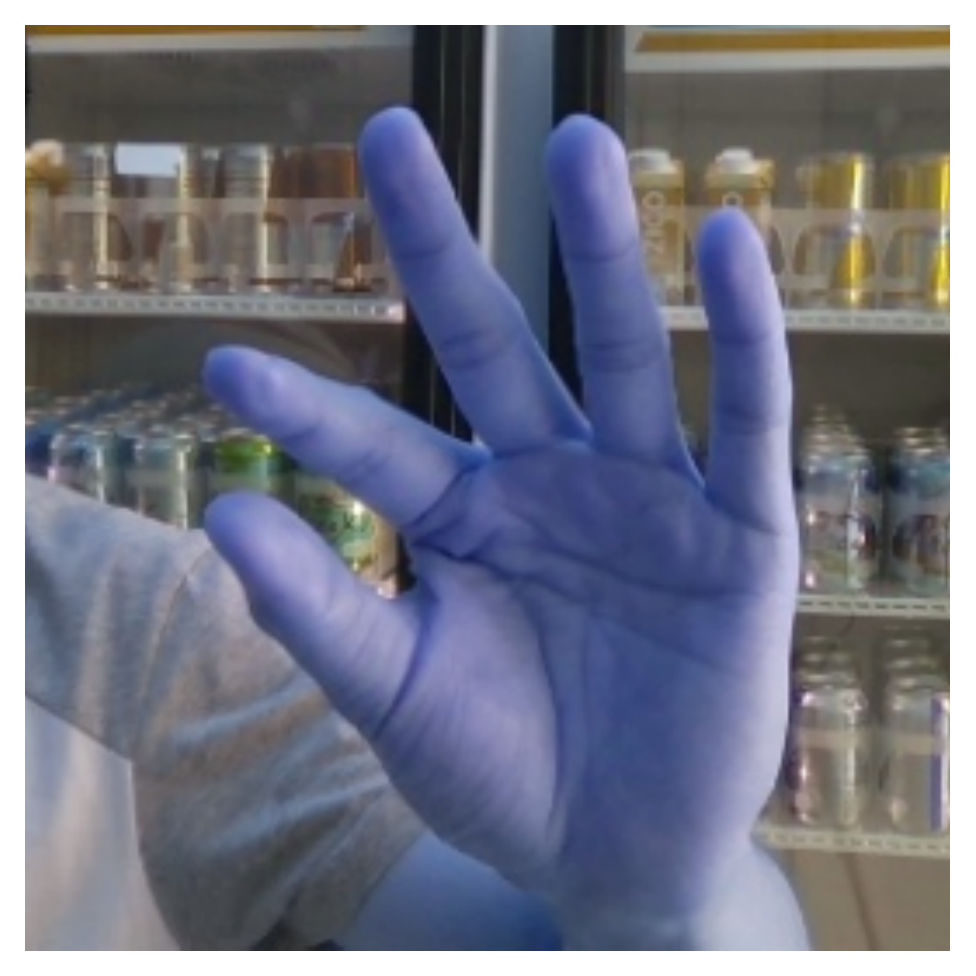}
  \end{subfigure}
  \begin{subfigure}{0.09\linewidth}
    \includegraphics[width=\linewidth]{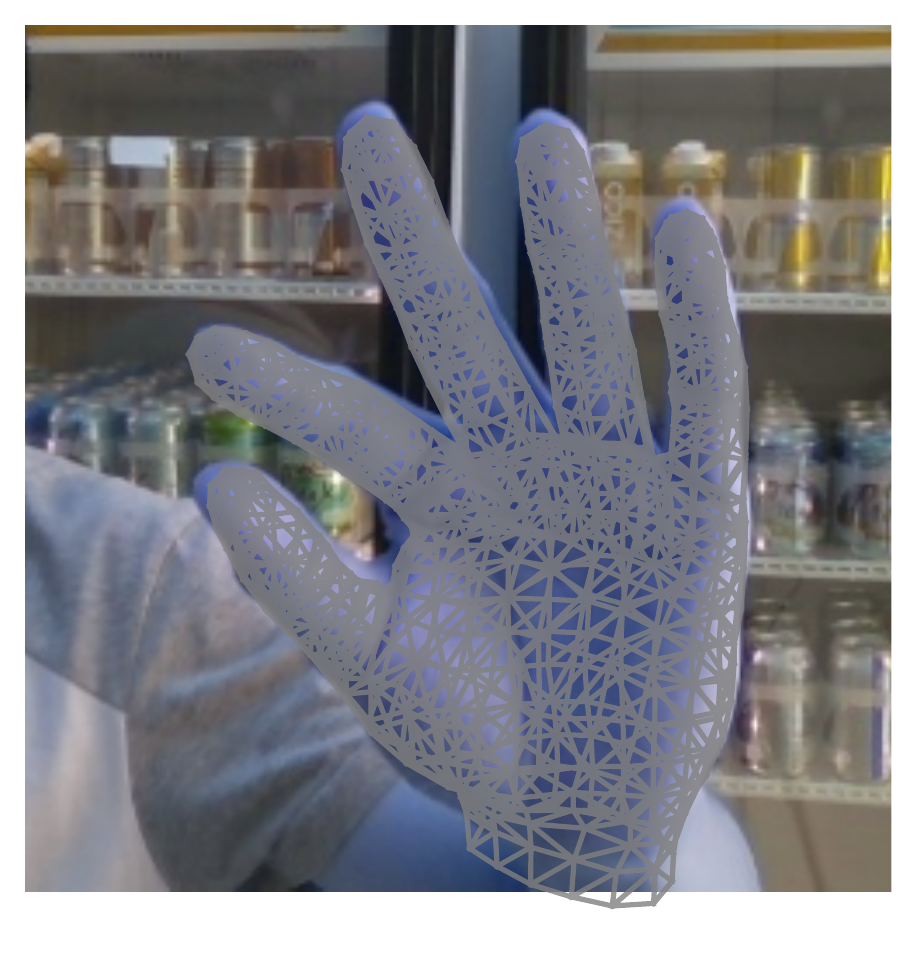}
  \end{subfigure}
  \begin{subfigure}{0.09\linewidth}
    \includegraphics[width=\linewidth]{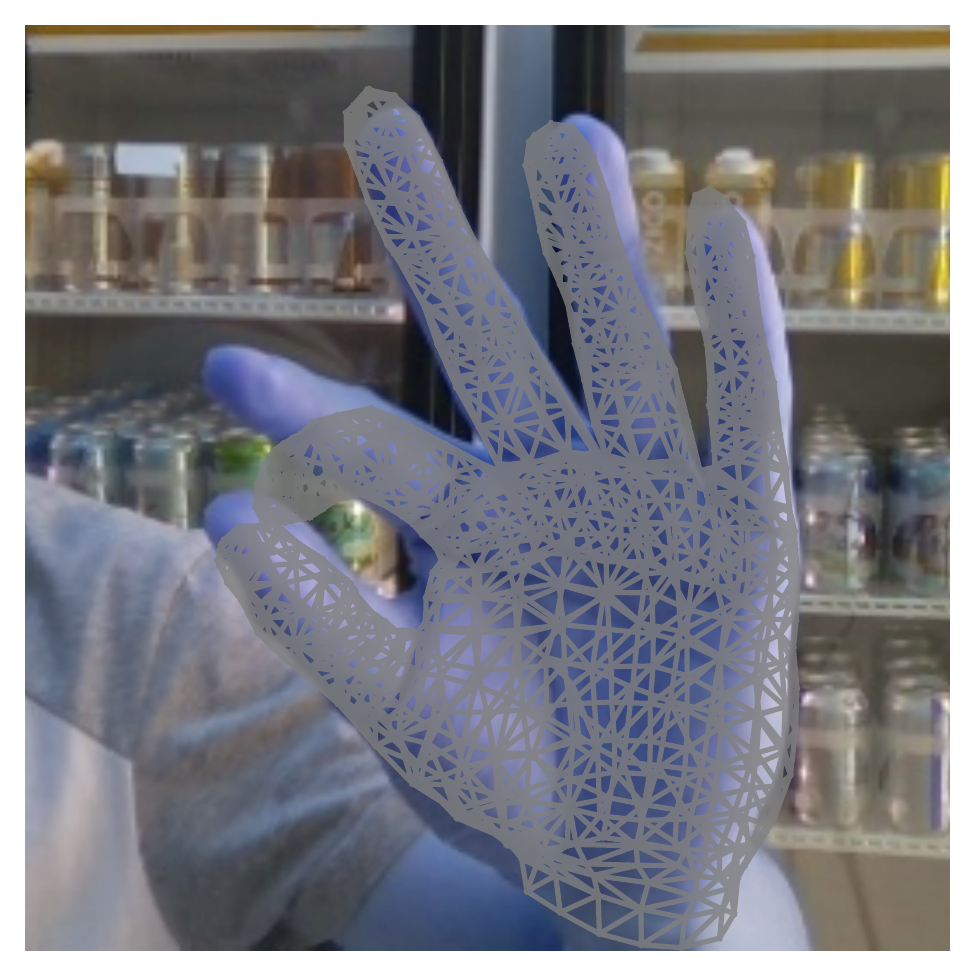}
  \end{subfigure}
  \begin{subfigure}{0.09\linewidth}
    \includegraphics[width=\linewidth]{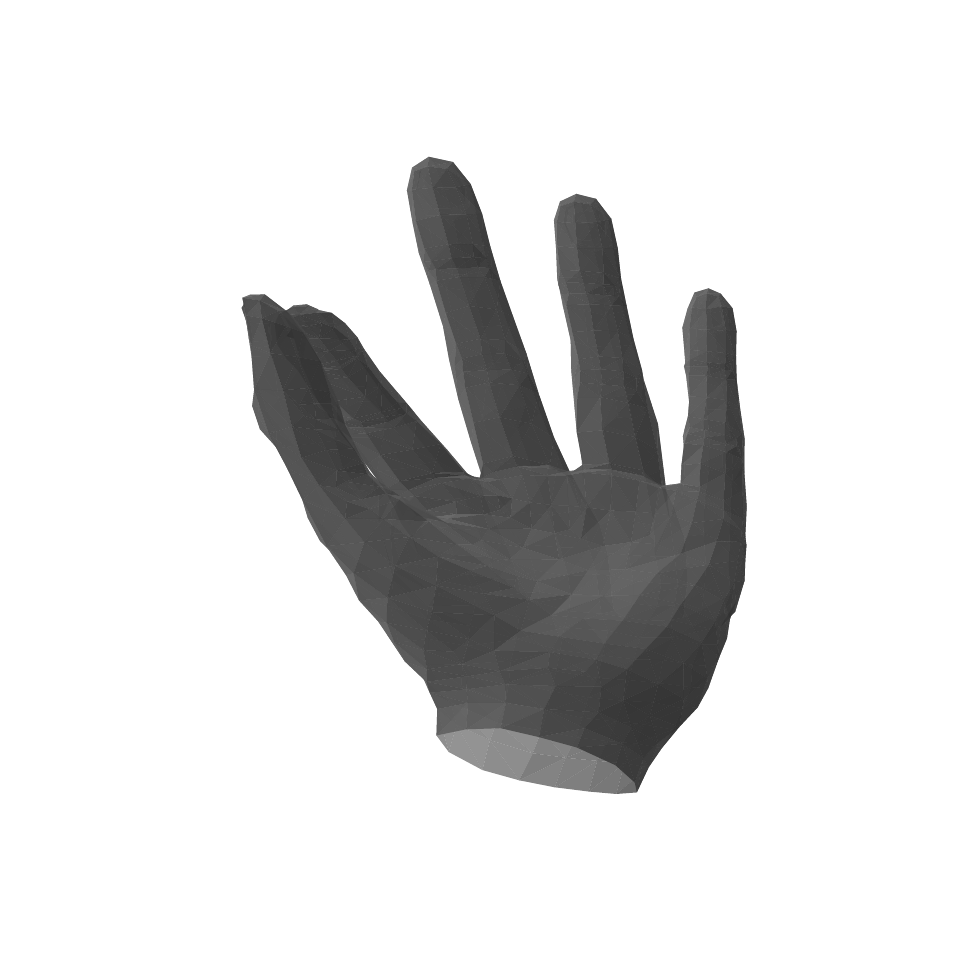}
  \end{subfigure}
  \begin{subfigure}{0.09\linewidth}
    \includegraphics[width=\linewidth]{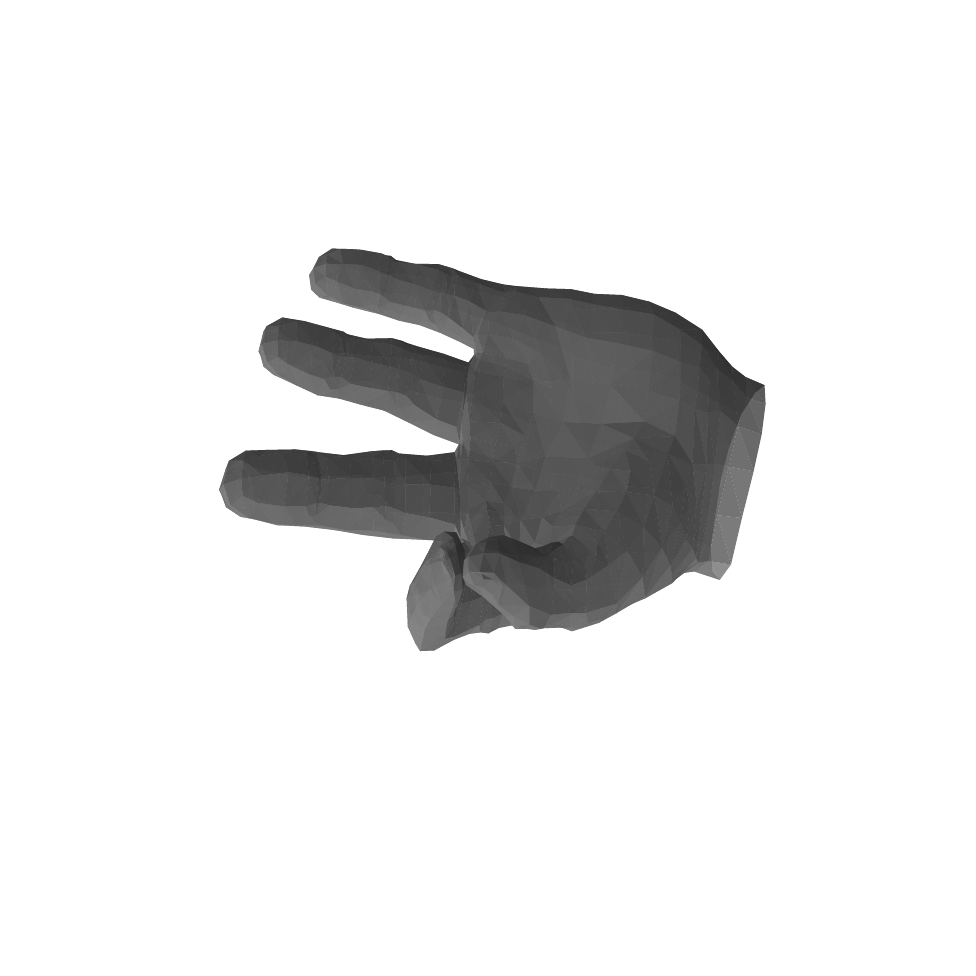}
  \end{subfigure}
  \begin{subfigure}{0.09\linewidth}
    \includegraphics[width=\linewidth]{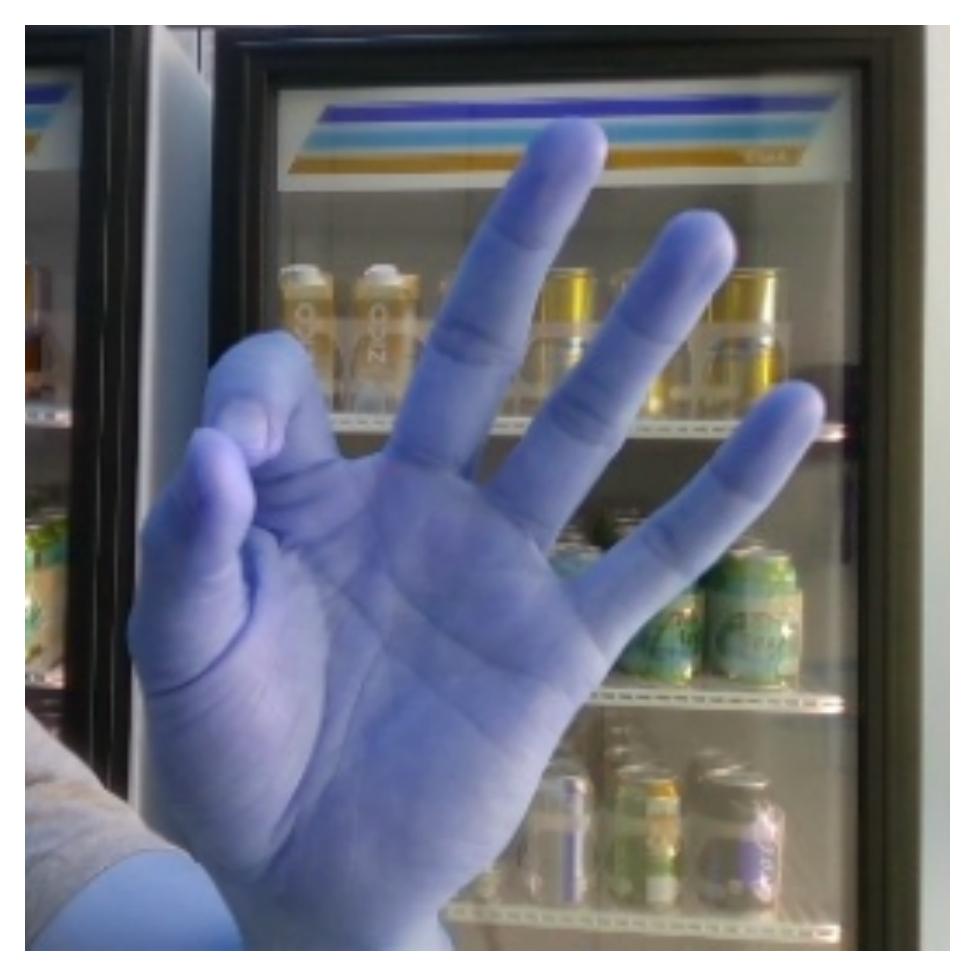}
  \end{subfigure}
  \begin{subfigure}{0.09\linewidth}
    \includegraphics[width=\linewidth]{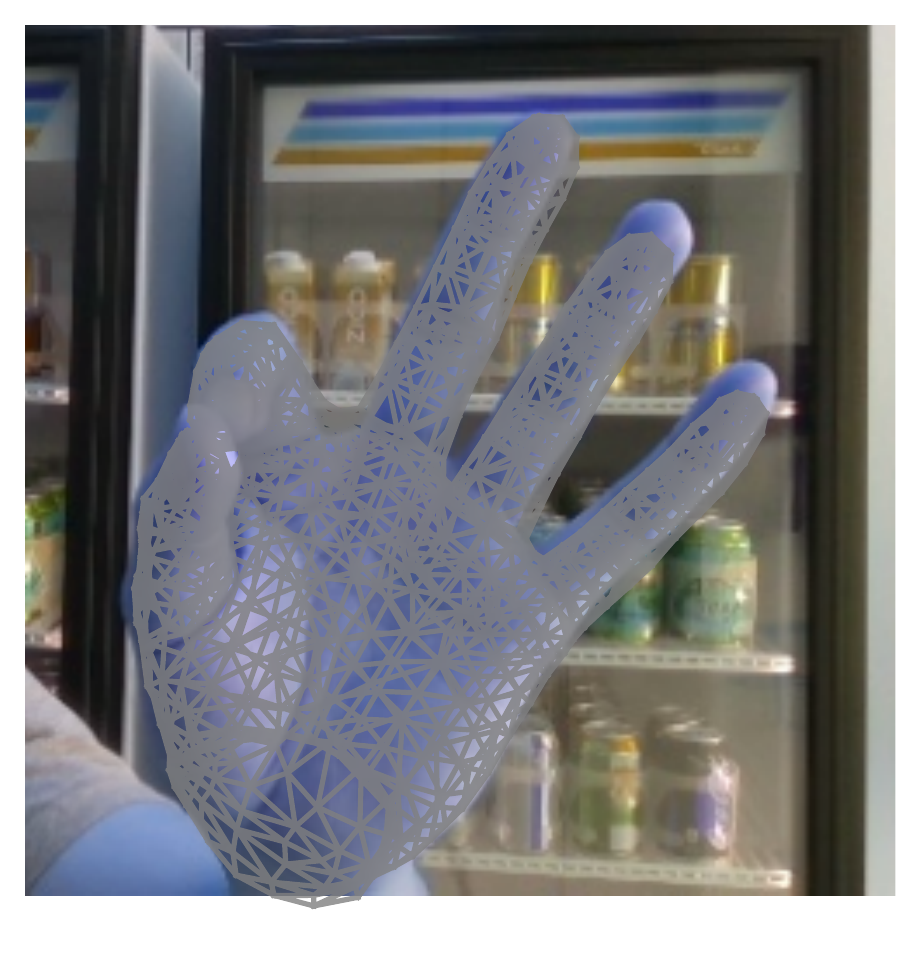}
  \end{subfigure}
  \begin{subfigure}{0.09\linewidth}
    \includegraphics[width=\linewidth]{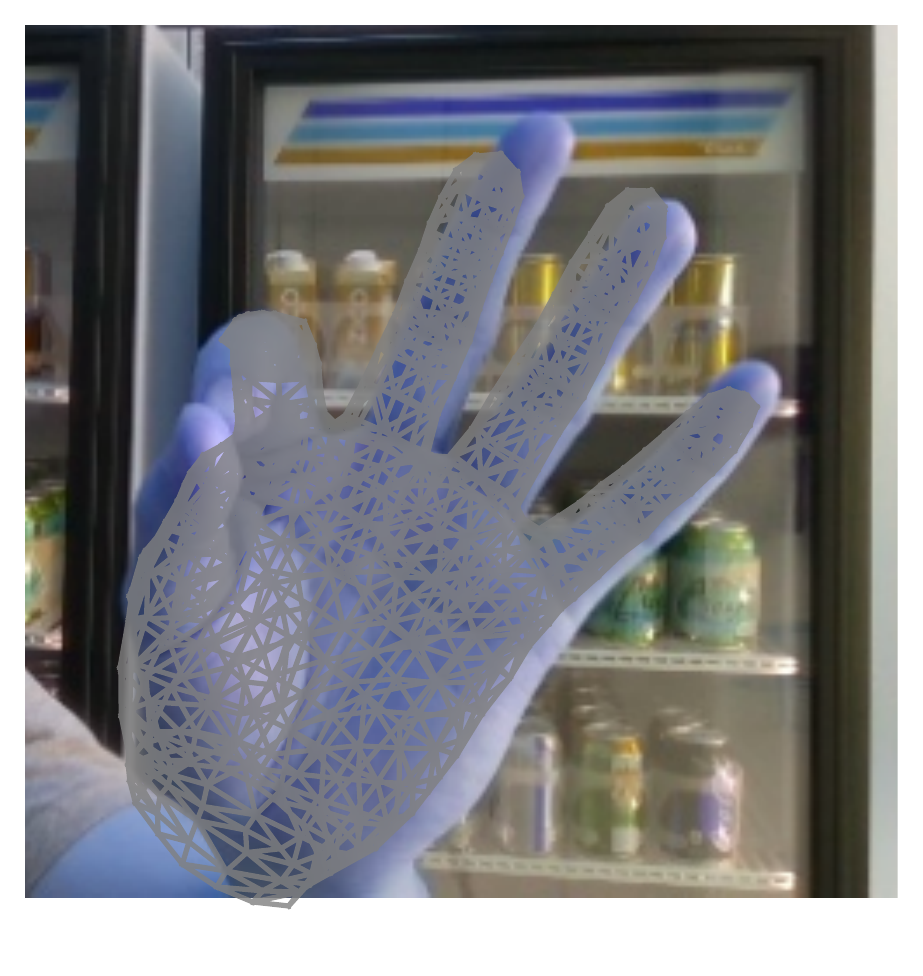}
  \end{subfigure}
  \begin{subfigure}{0.09\linewidth}
    \includegraphics[width=\linewidth]{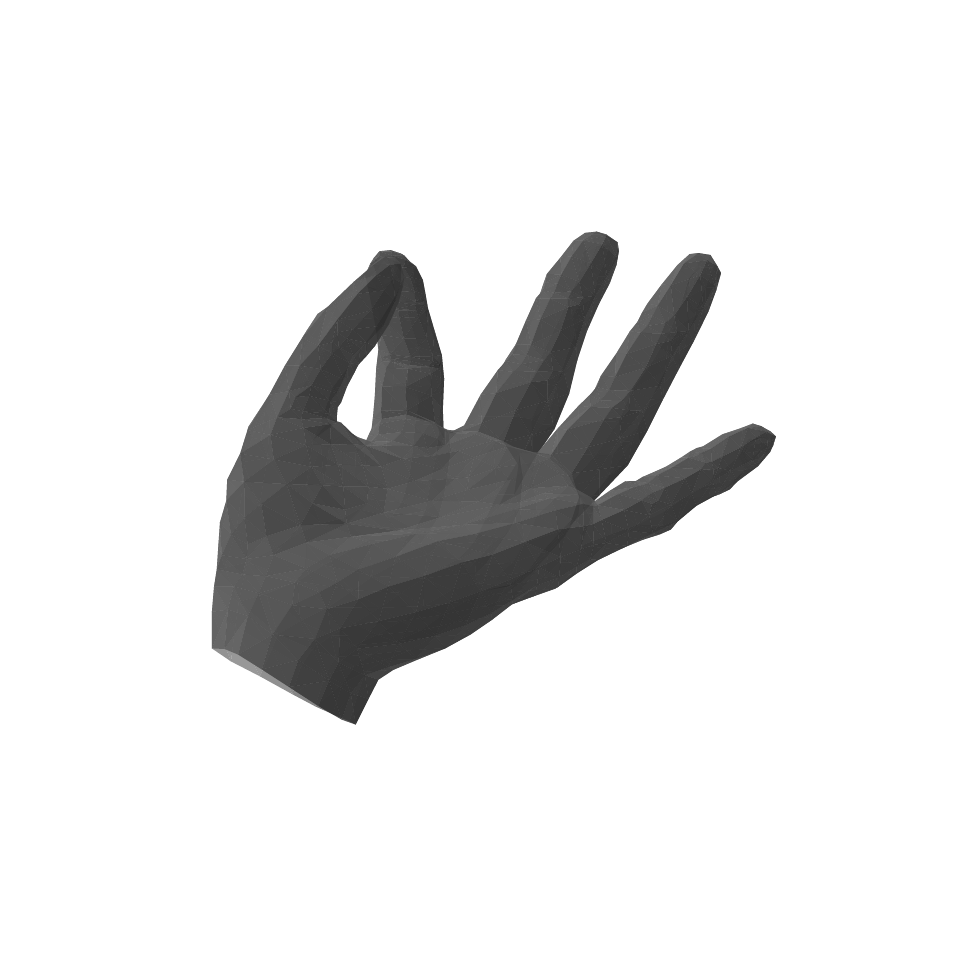}
  \end{subfigure}
  \begin{subfigure}{0.09\linewidth}
    \includegraphics[width=\linewidth]{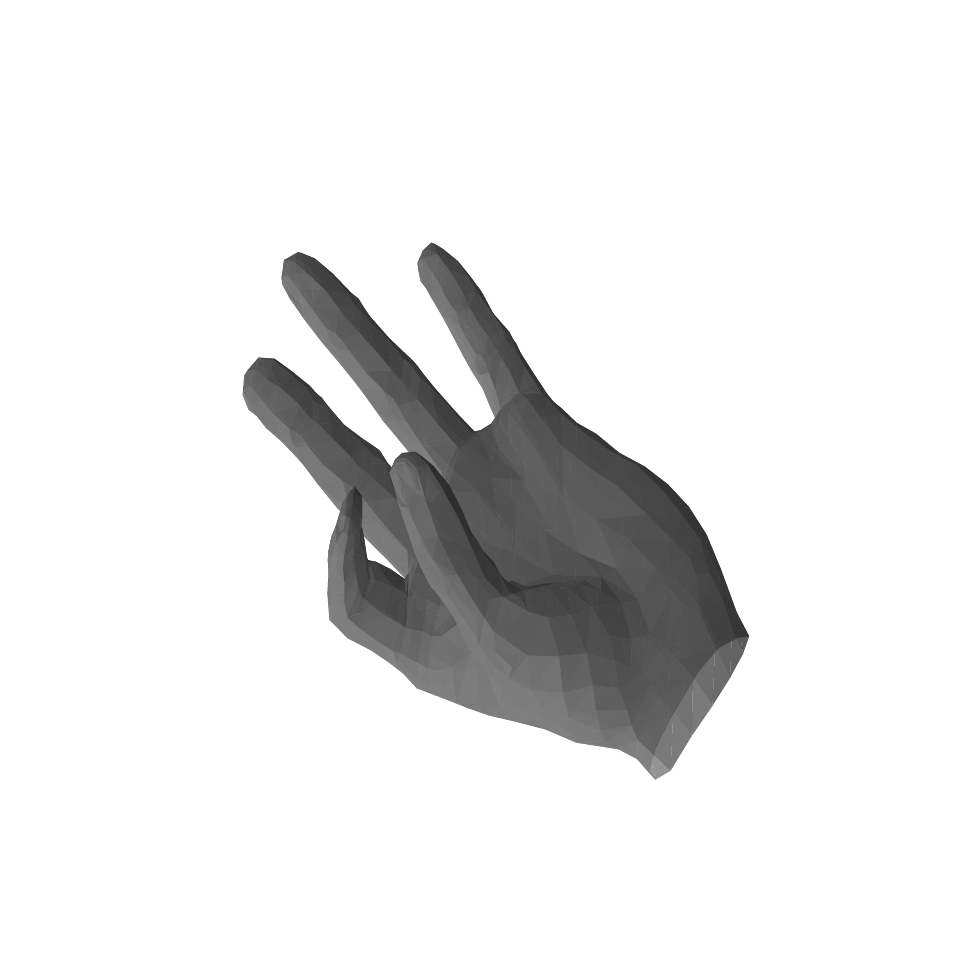}
  \end{subfigure}
  
  \begin{subfigure}{0.09\linewidth}
    \includegraphics[width=\linewidth]{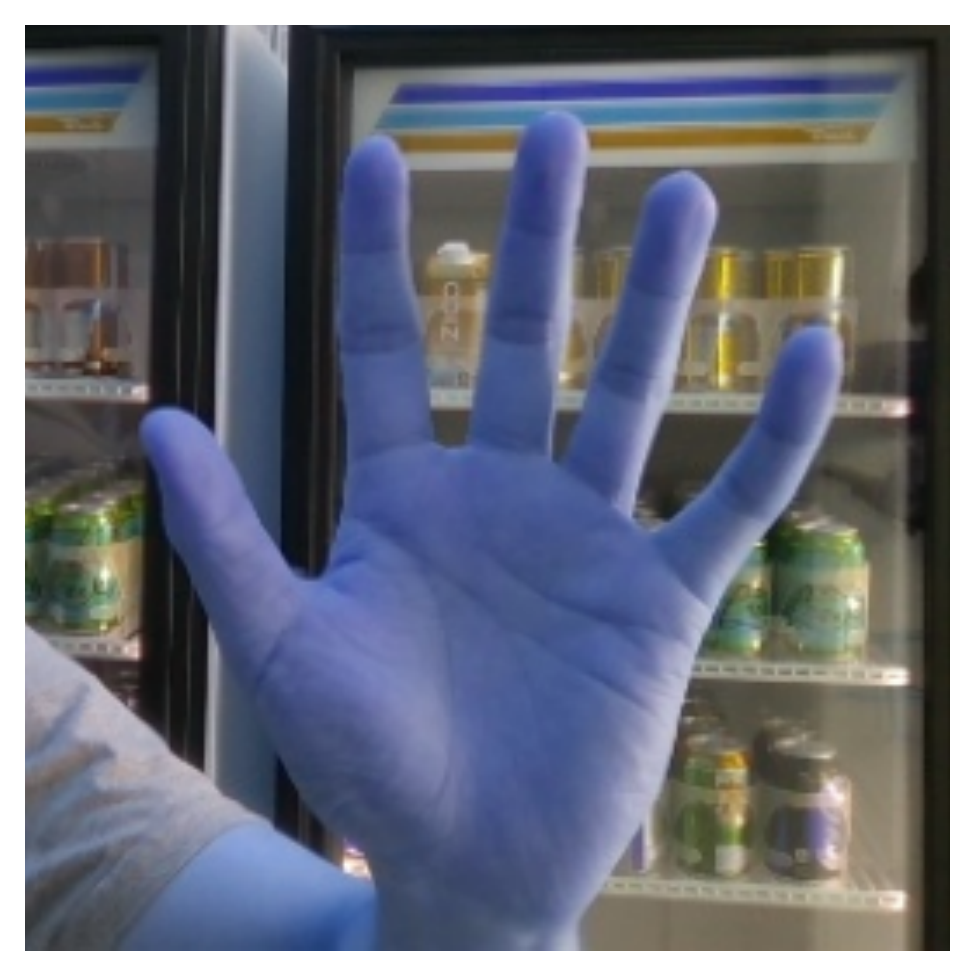}
  \end{subfigure}
  \begin{subfigure}{0.09\linewidth}
    \includegraphics[width=\linewidth]{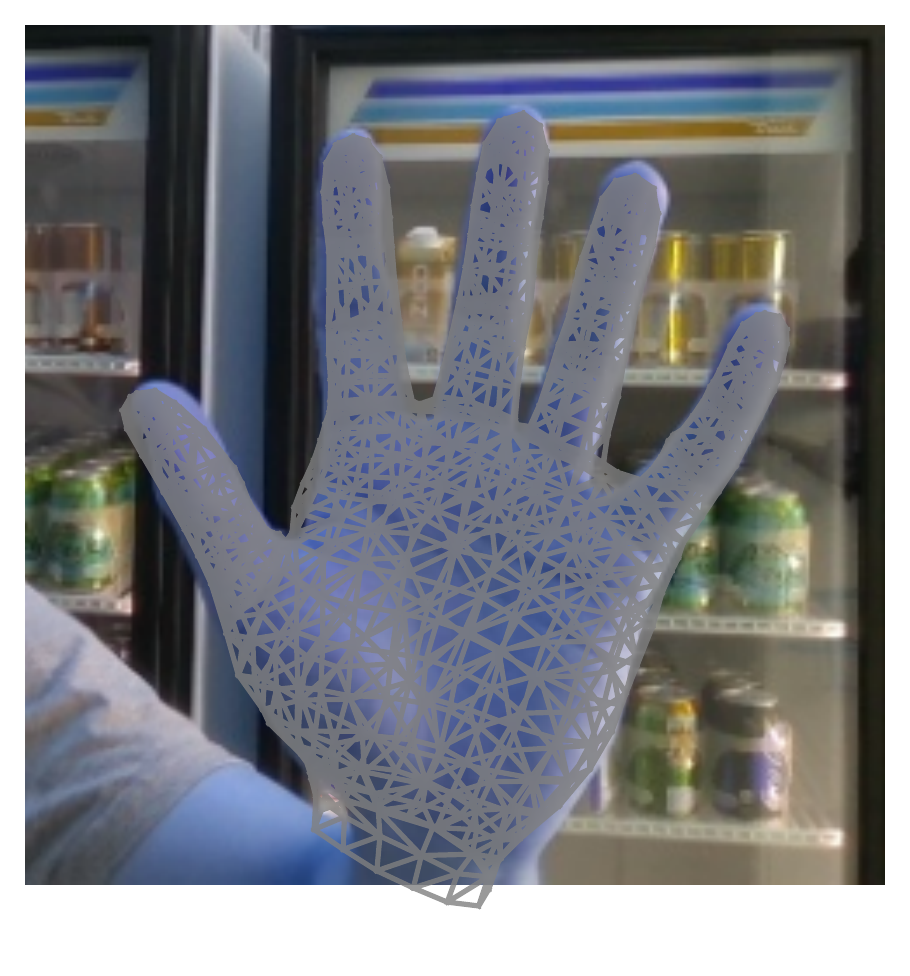}
  \end{subfigure}
  \begin{subfigure}{0.09\linewidth}
    \includegraphics[width=\linewidth]{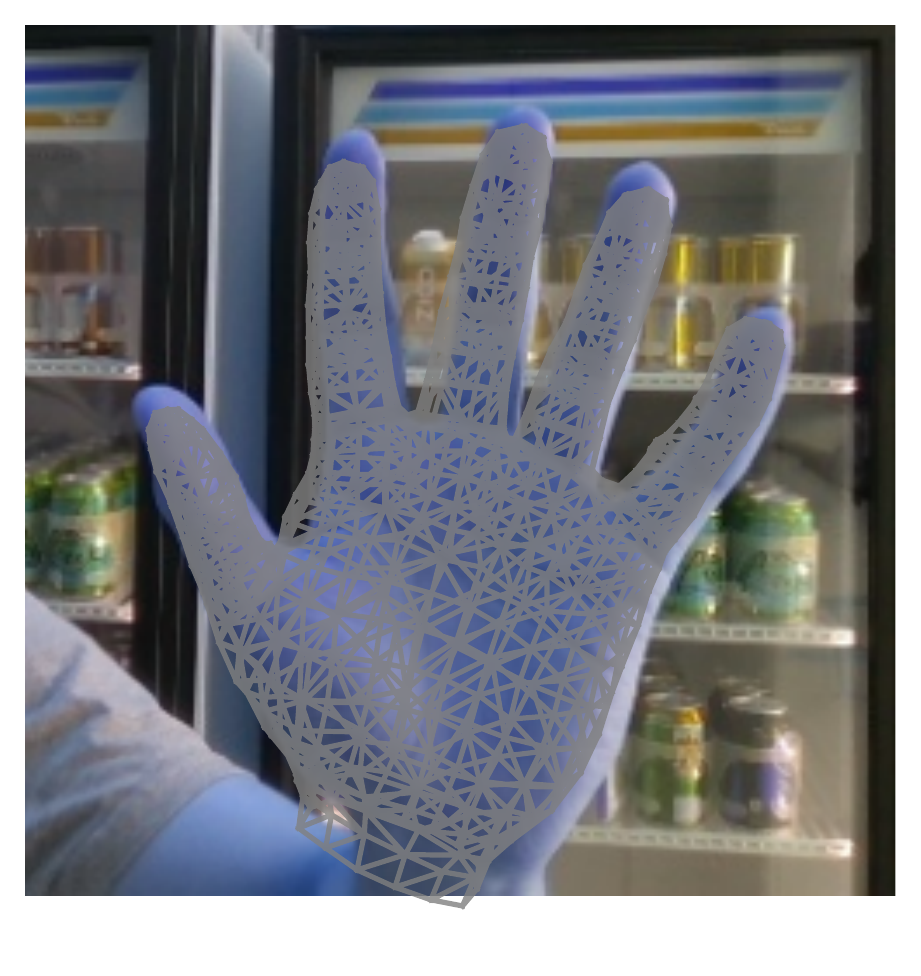}
  \end{subfigure}
  \begin{subfigure}{0.09\linewidth}
    \includegraphics[width=\linewidth]{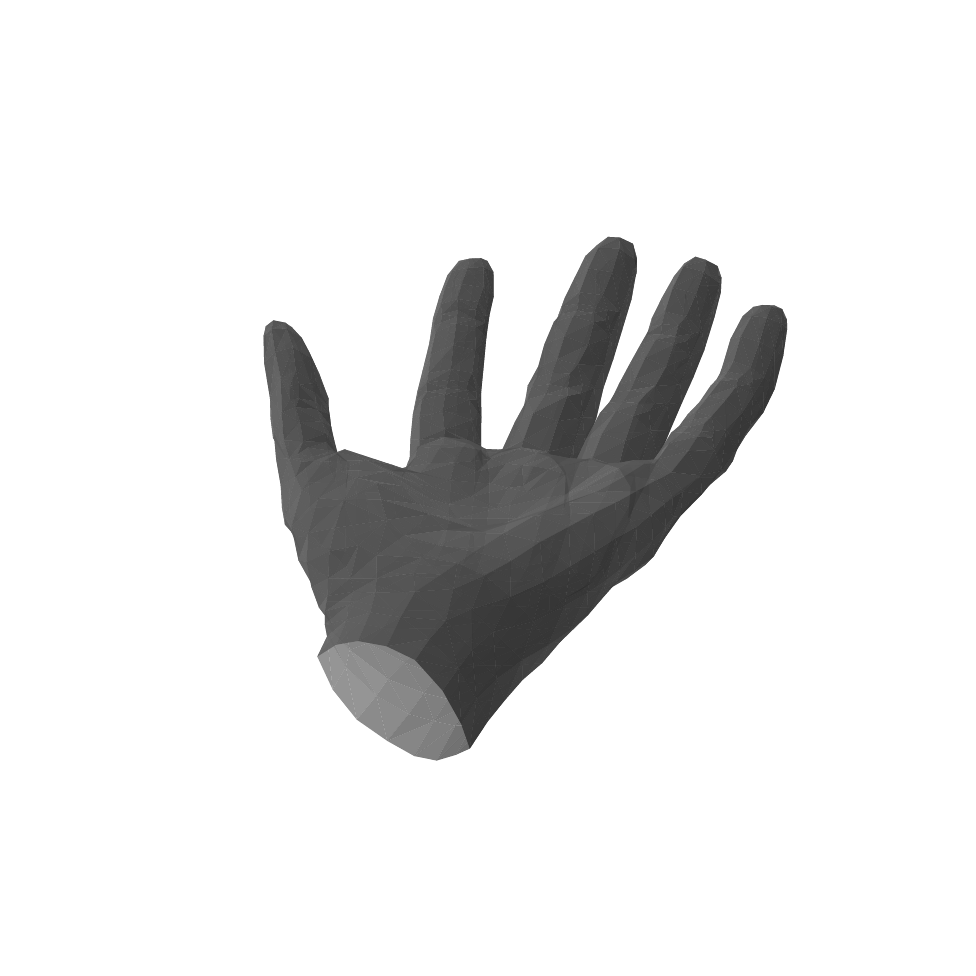}
  \end{subfigure}
  \begin{subfigure}{0.09\linewidth}
    \includegraphics[width=\linewidth]{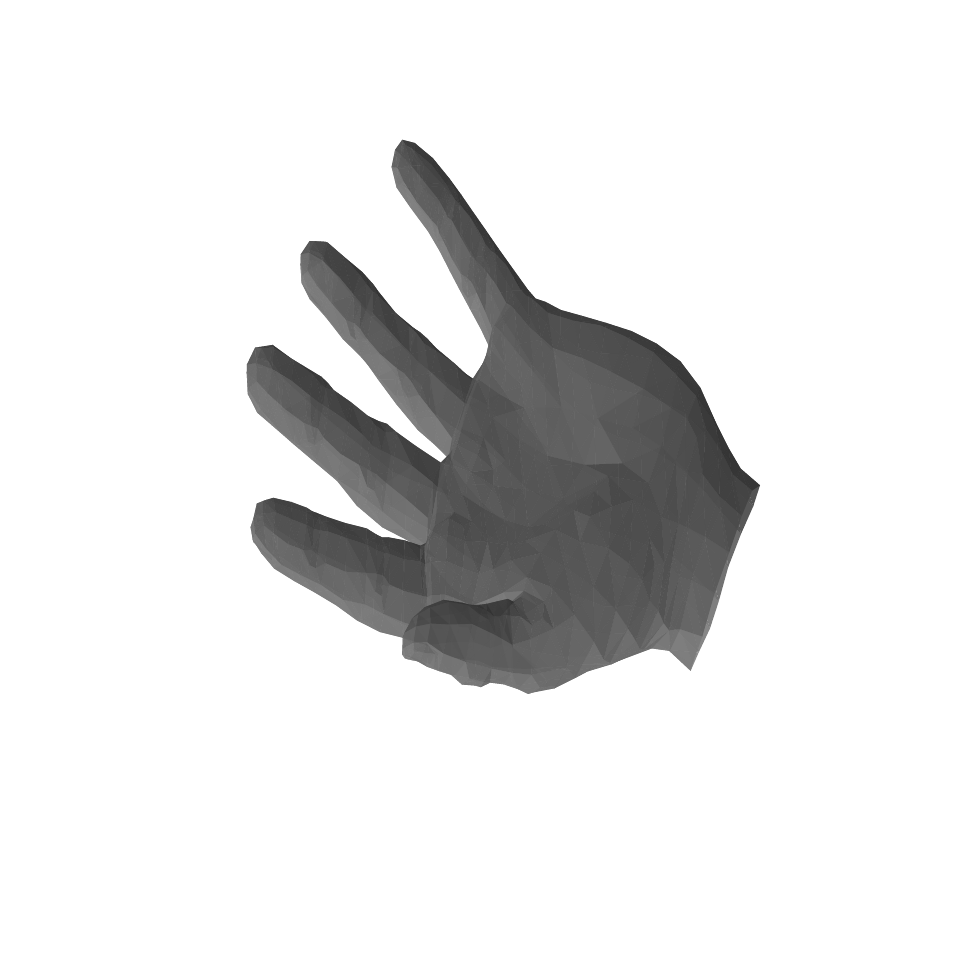}
  \end{subfigure}
  \begin{subfigure}{0.09\linewidth}
    \includegraphics[width=\linewidth]{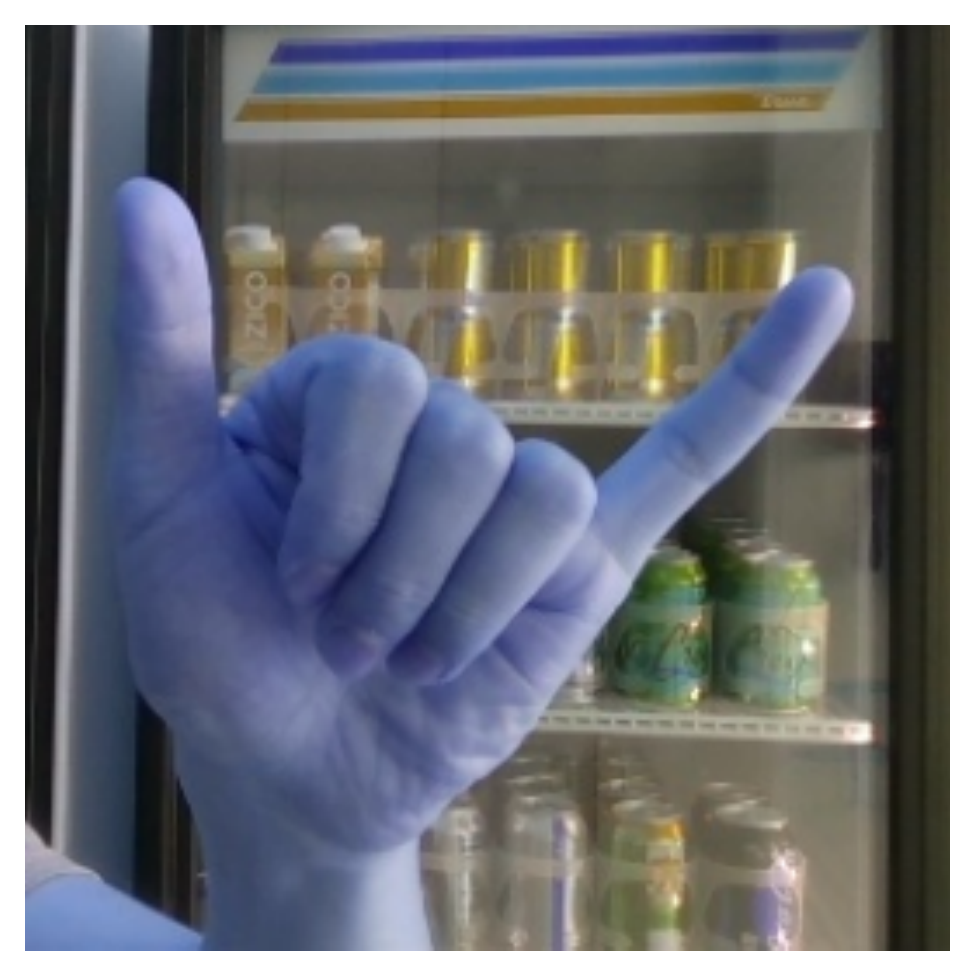}
  \end{subfigure}
  \begin{subfigure}{0.09\linewidth}
    \includegraphics[width=\linewidth]{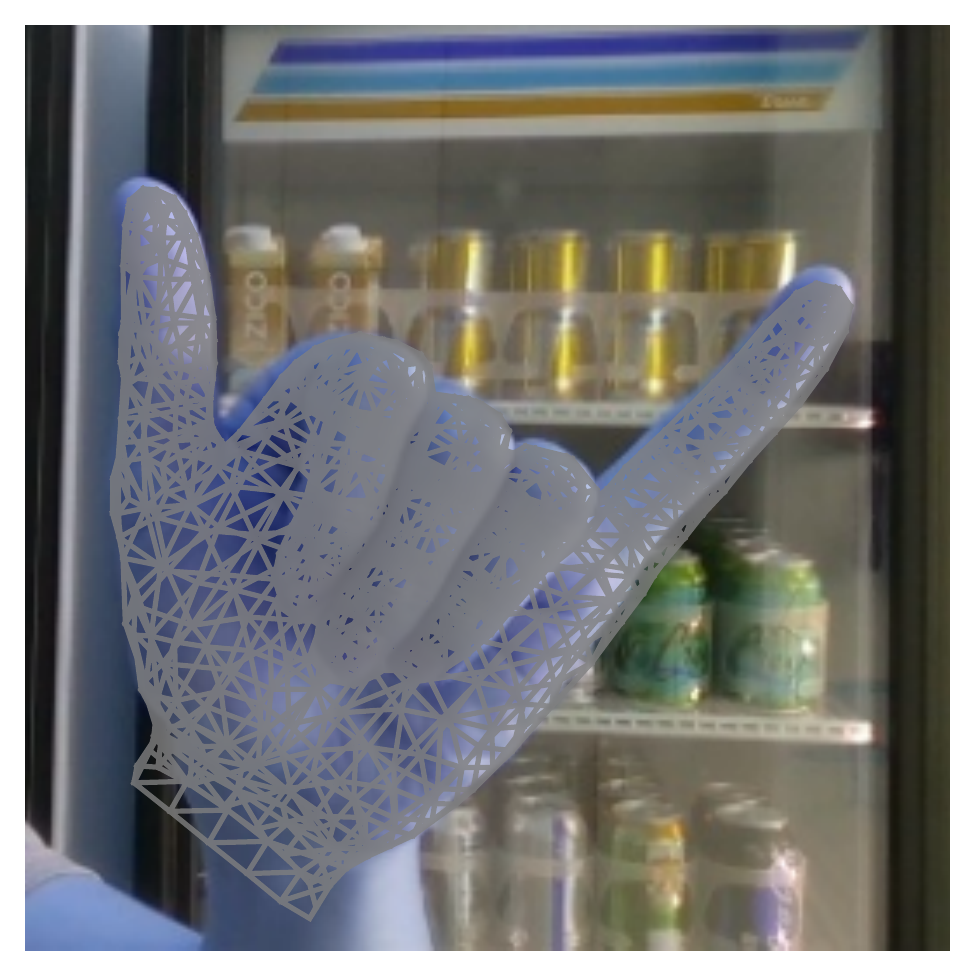}
  \end{subfigure}
  \begin{subfigure}{0.09\linewidth}
    \includegraphics[width=\linewidth]{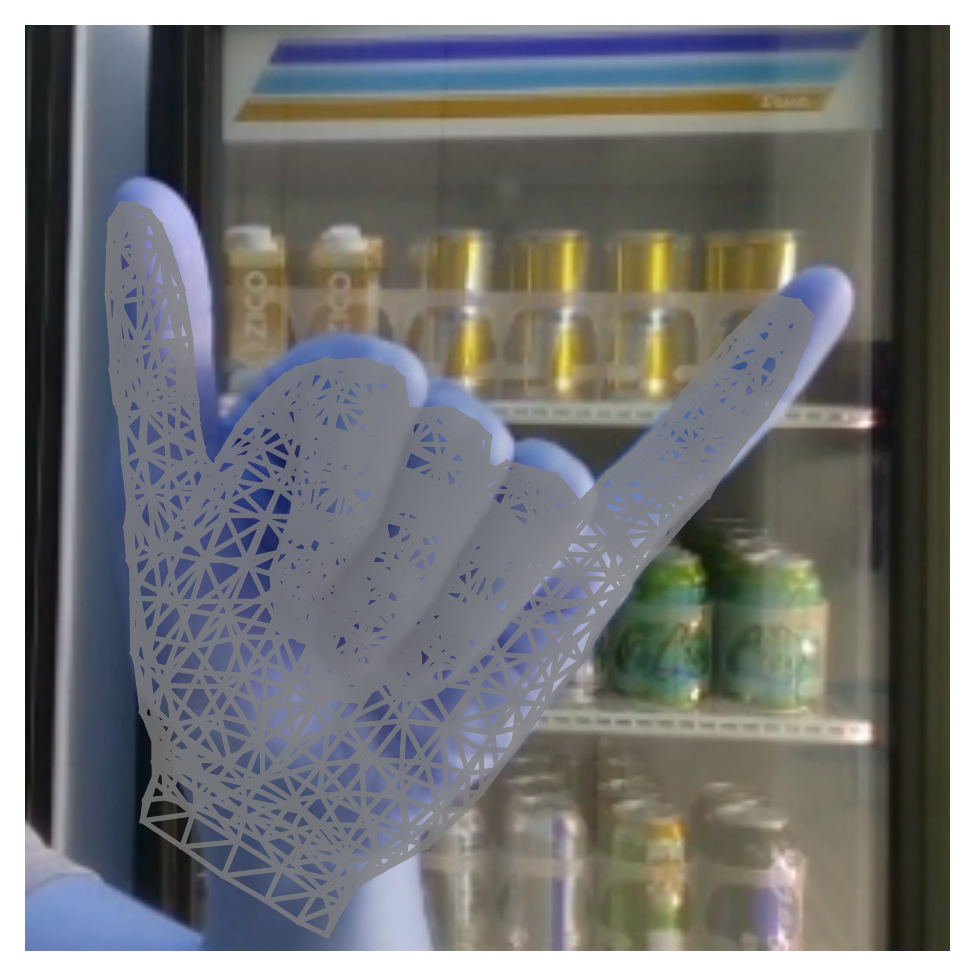}
  \end{subfigure}
  \begin{subfigure}{0.09\linewidth}
    \includegraphics[width=\linewidth]{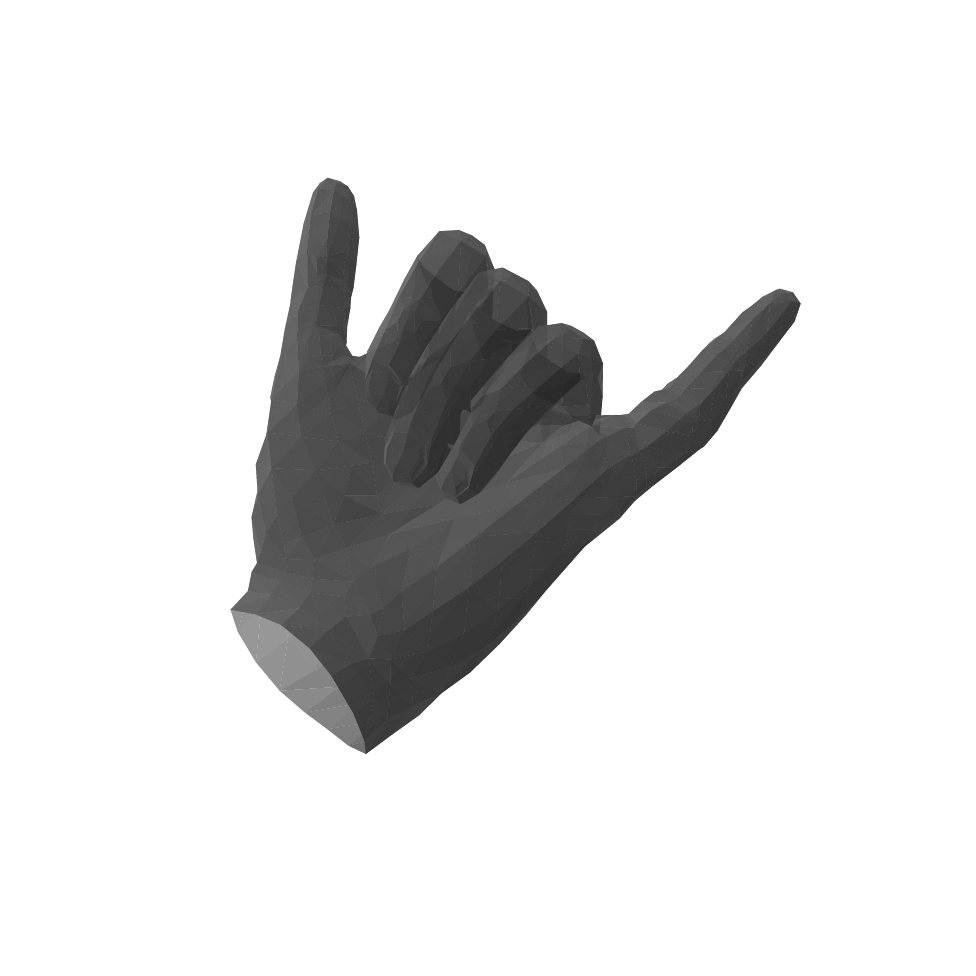}
  \end{subfigure}
  \begin{subfigure}{0.09\linewidth}
    \includegraphics[width=\linewidth]{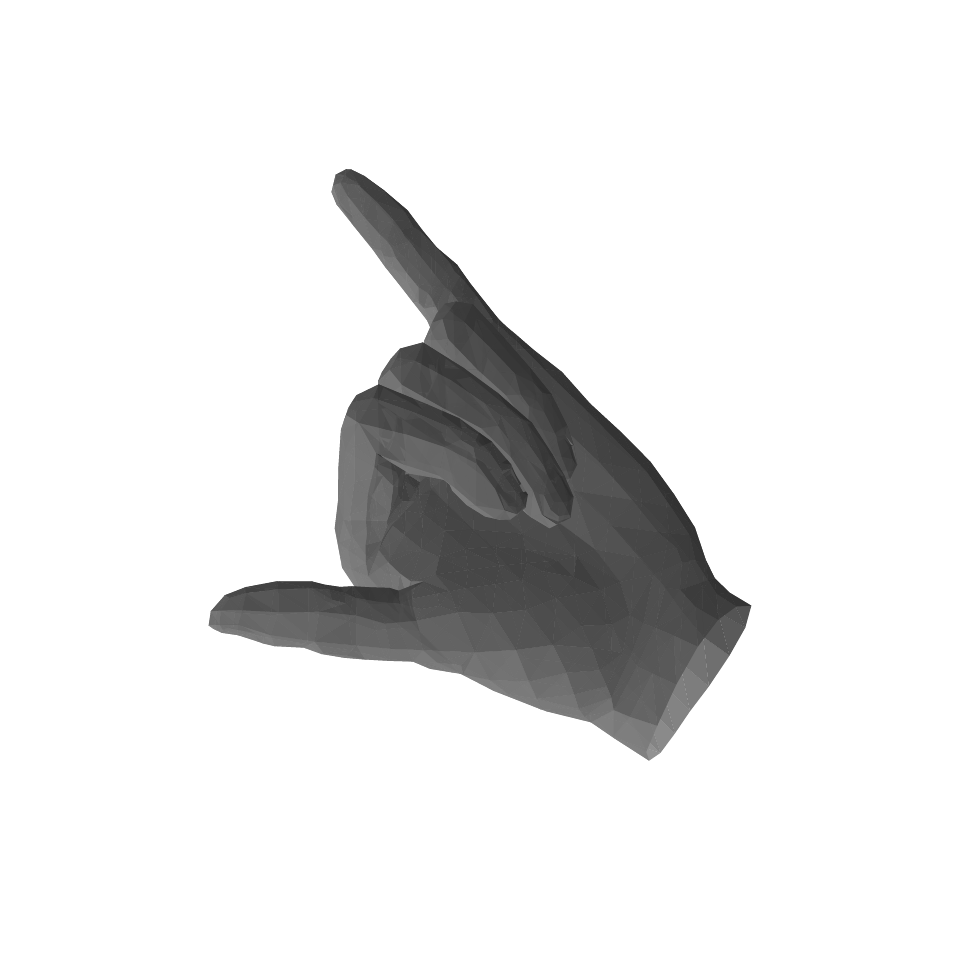}
  \end{subfigure}
  
  \begin{subfigure}{0.09\linewidth}
    \includegraphics[width=\linewidth]{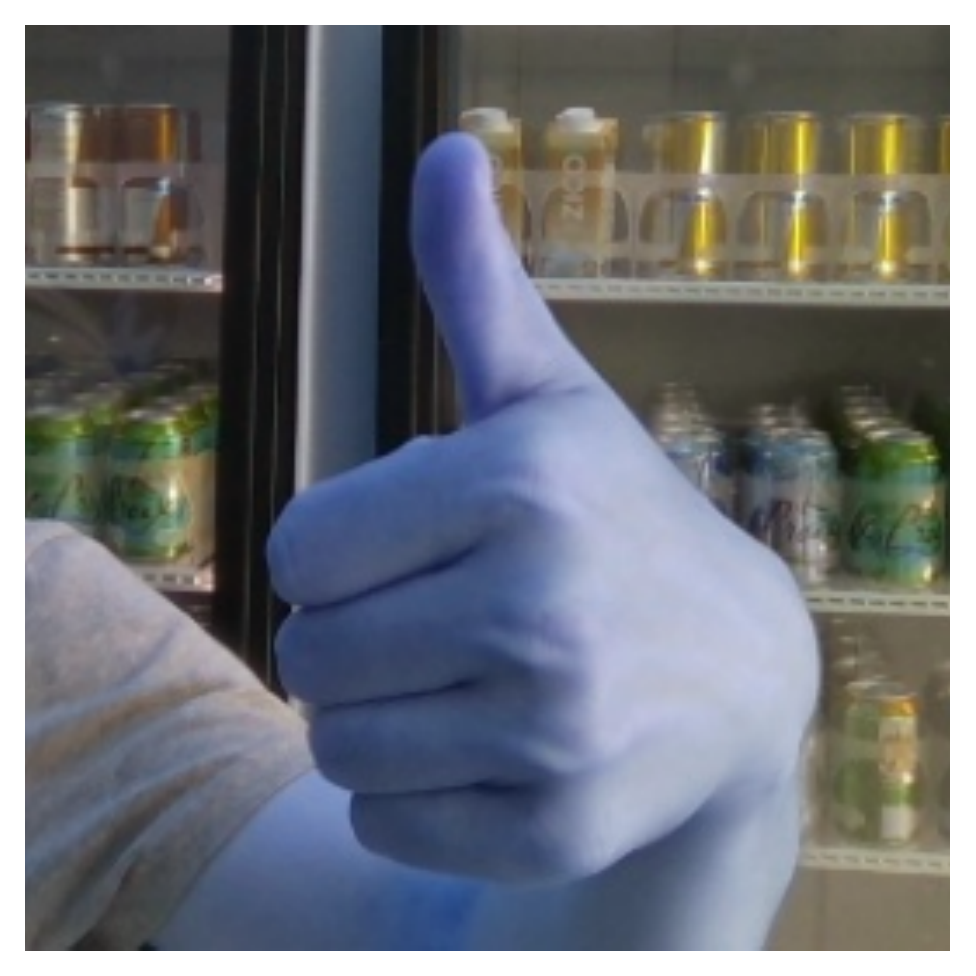}
  \end{subfigure}
  \begin{subfigure}{0.09\linewidth}
    \includegraphics[width=\linewidth]{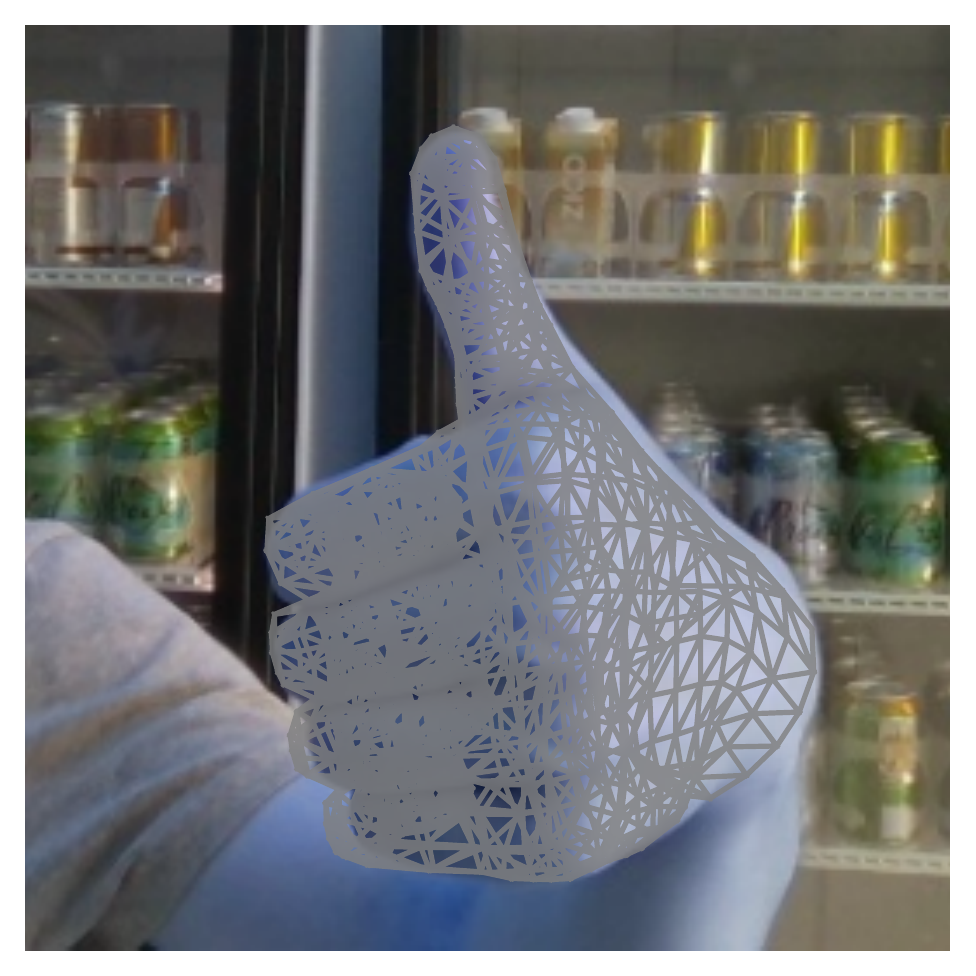}
  \end{subfigure}
  \begin{subfigure}{0.09\linewidth}
    \includegraphics[width=\linewidth]{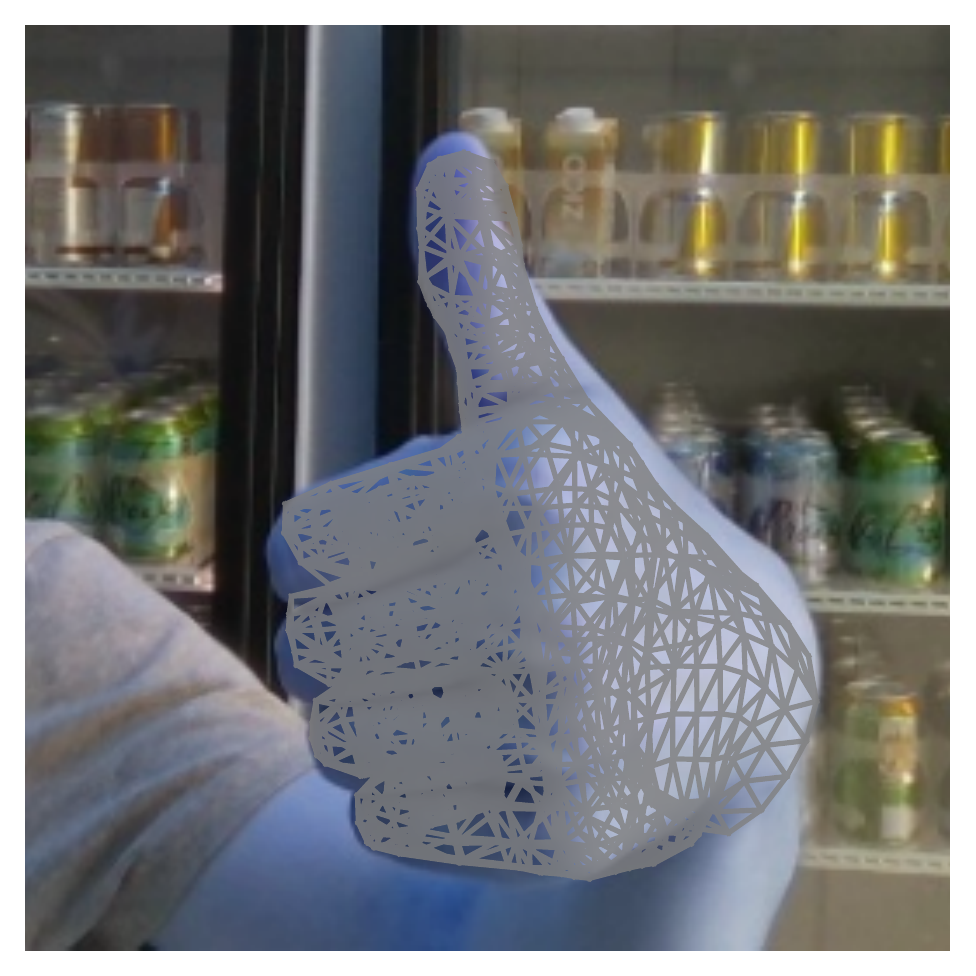}
  \end{subfigure}
  \begin{subfigure}{0.09\linewidth}
    \includegraphics[width=\linewidth]{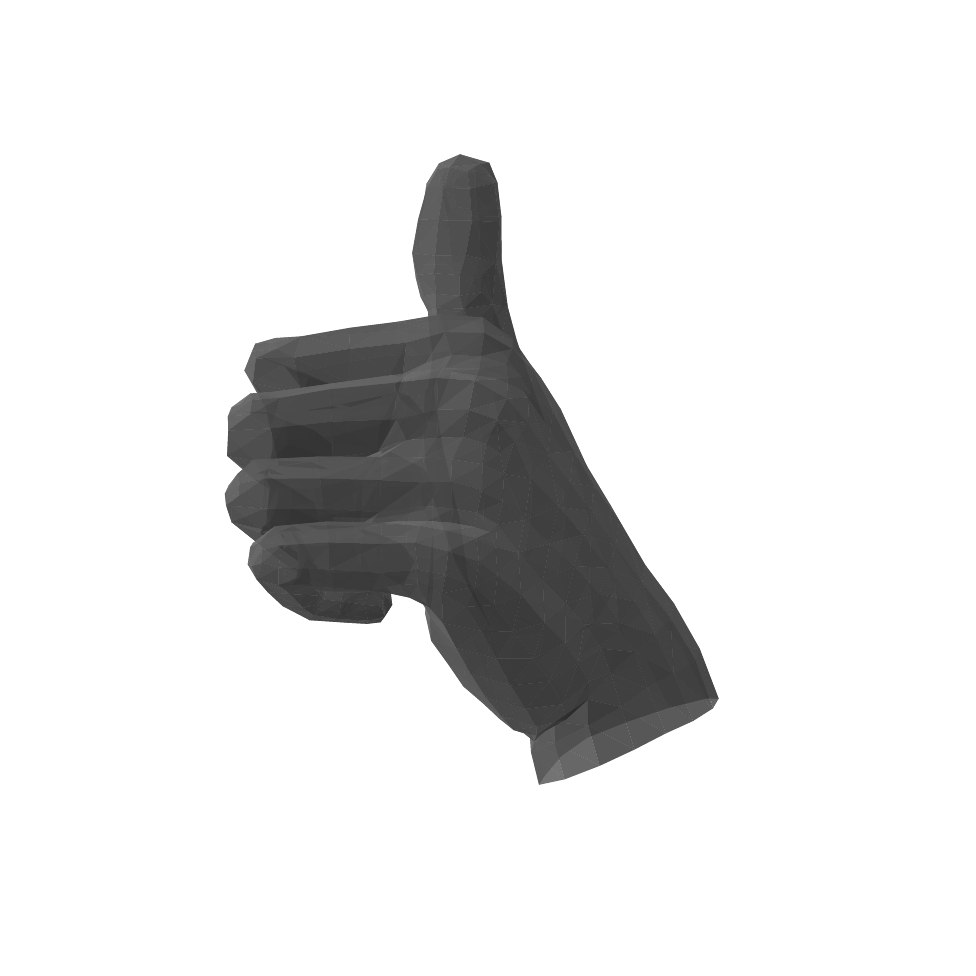}
  \end{subfigure}
  \begin{subfigure}{0.09\linewidth}
    \includegraphics[width=\linewidth]{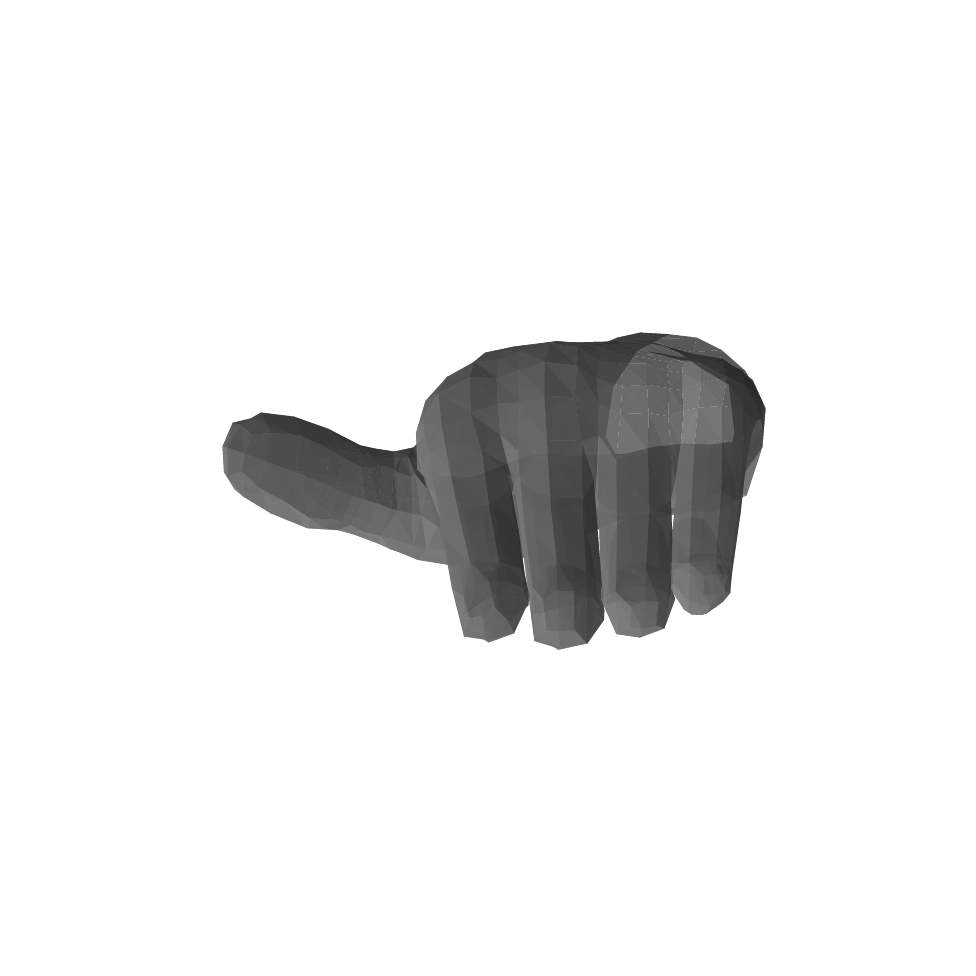}
  \end{subfigure}
  \begin{subfigure}{0.09\linewidth}
    \includegraphics[width=\linewidth]{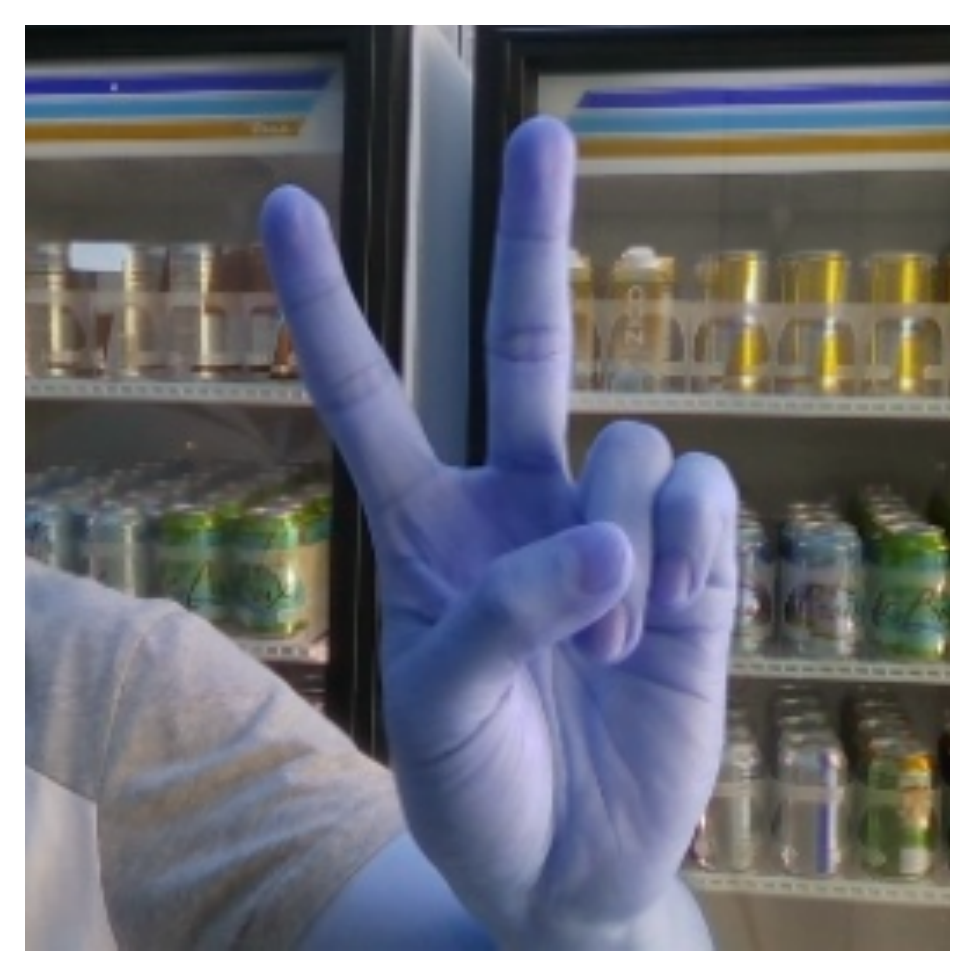}
  \end{subfigure}
  \begin{subfigure}{0.09\linewidth}
    \includegraphics[width=\linewidth]{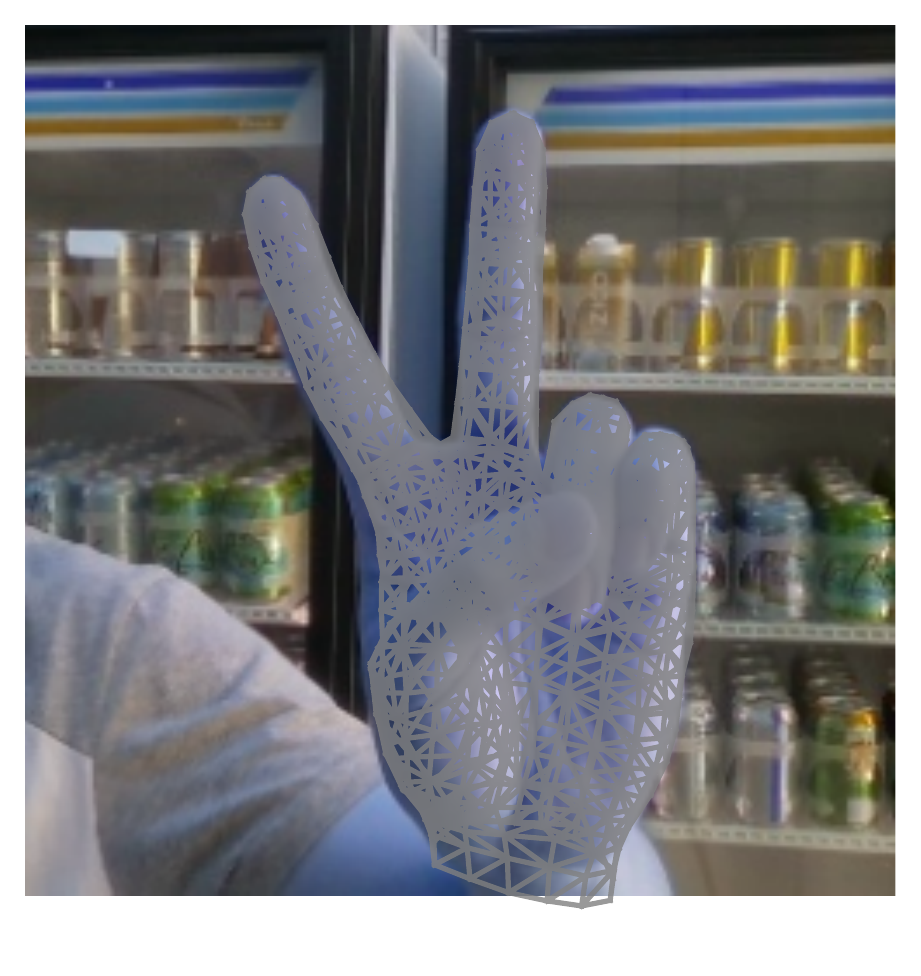}
  \end{subfigure}
  \begin{subfigure}{0.09\linewidth}
    \includegraphics[width=\linewidth]{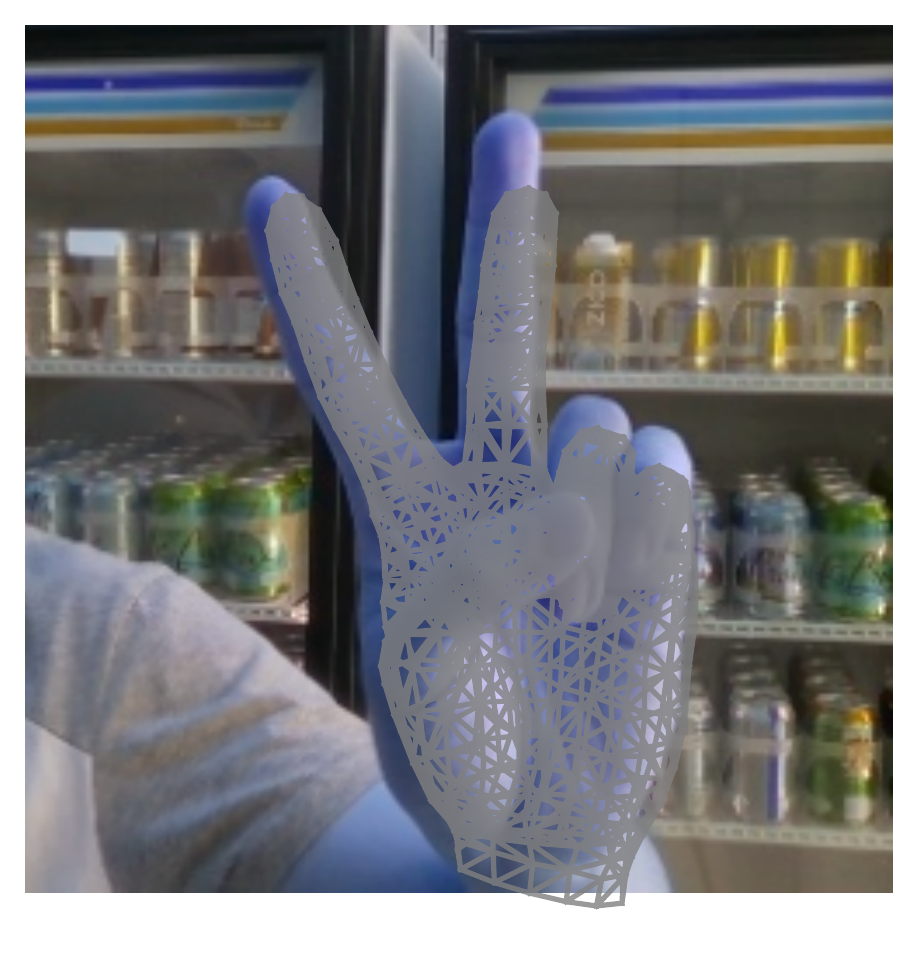}
  \end{subfigure}
  \begin{subfigure}{0.09\linewidth}
    \includegraphics[width=\linewidth]{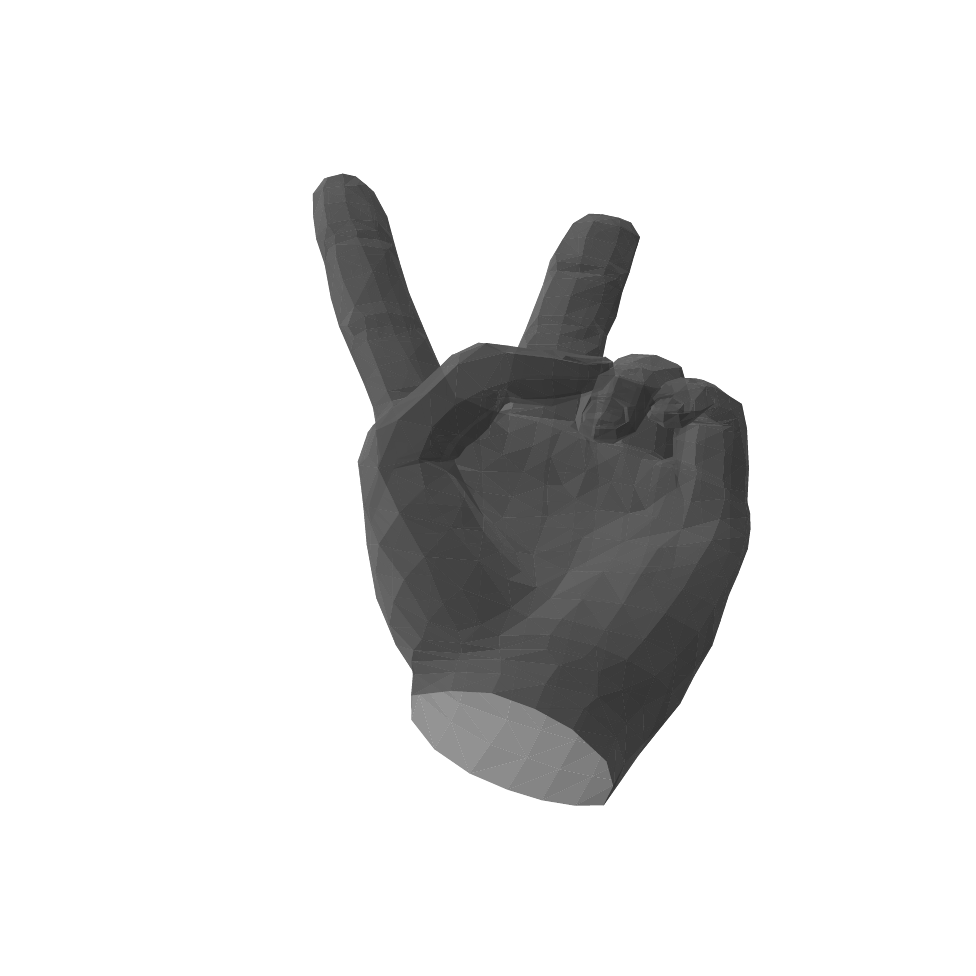}
  \end{subfigure}
  \begin{subfigure}{0.09\linewidth}
    \includegraphics[width=\linewidth]{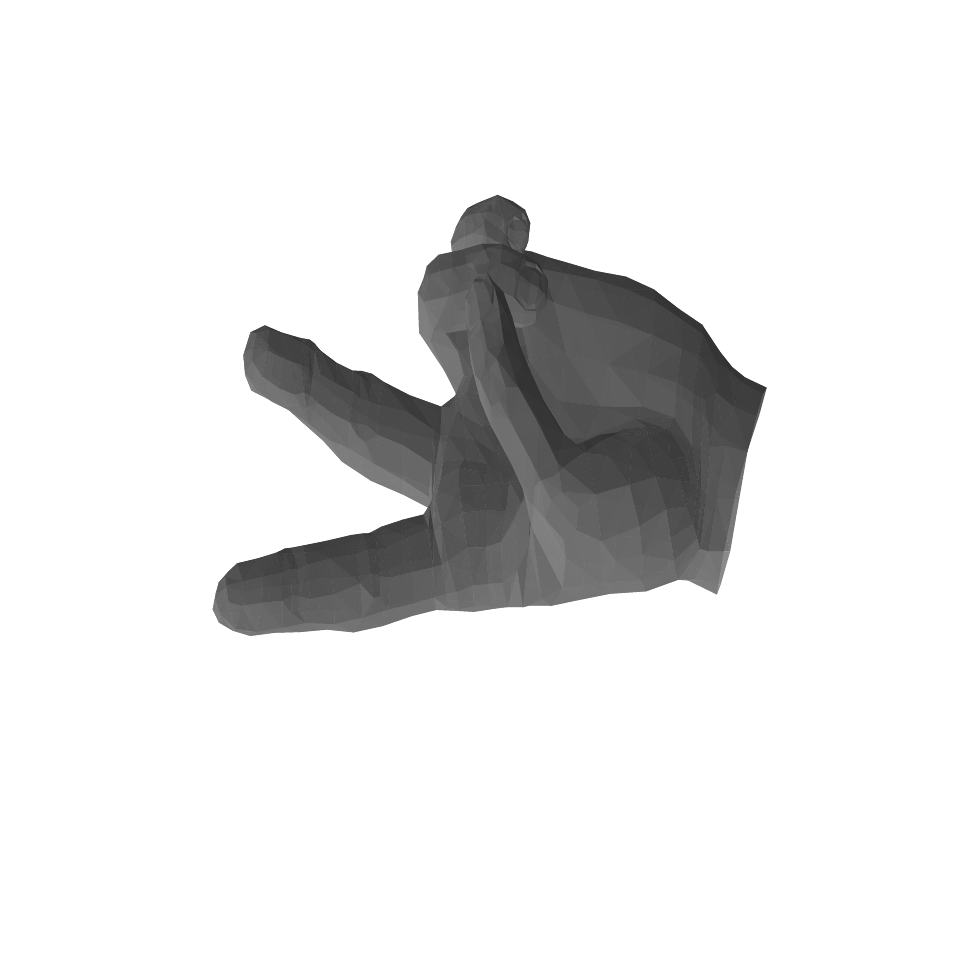}
  \end{subfigure}
  
  \begin{subfigure}{0.09\linewidth}
    \includegraphics[width=\linewidth]{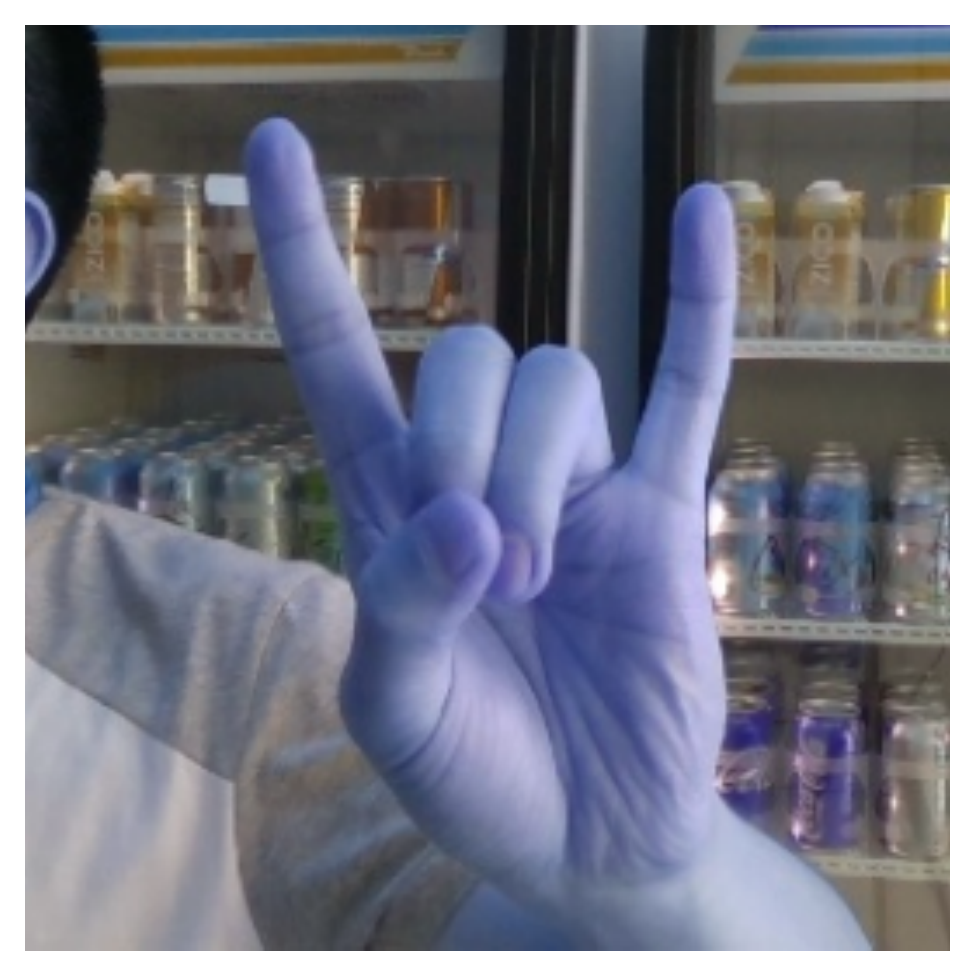}
  \end{subfigure}
  \begin{subfigure}{0.09\linewidth}
    \includegraphics[width=\linewidth]{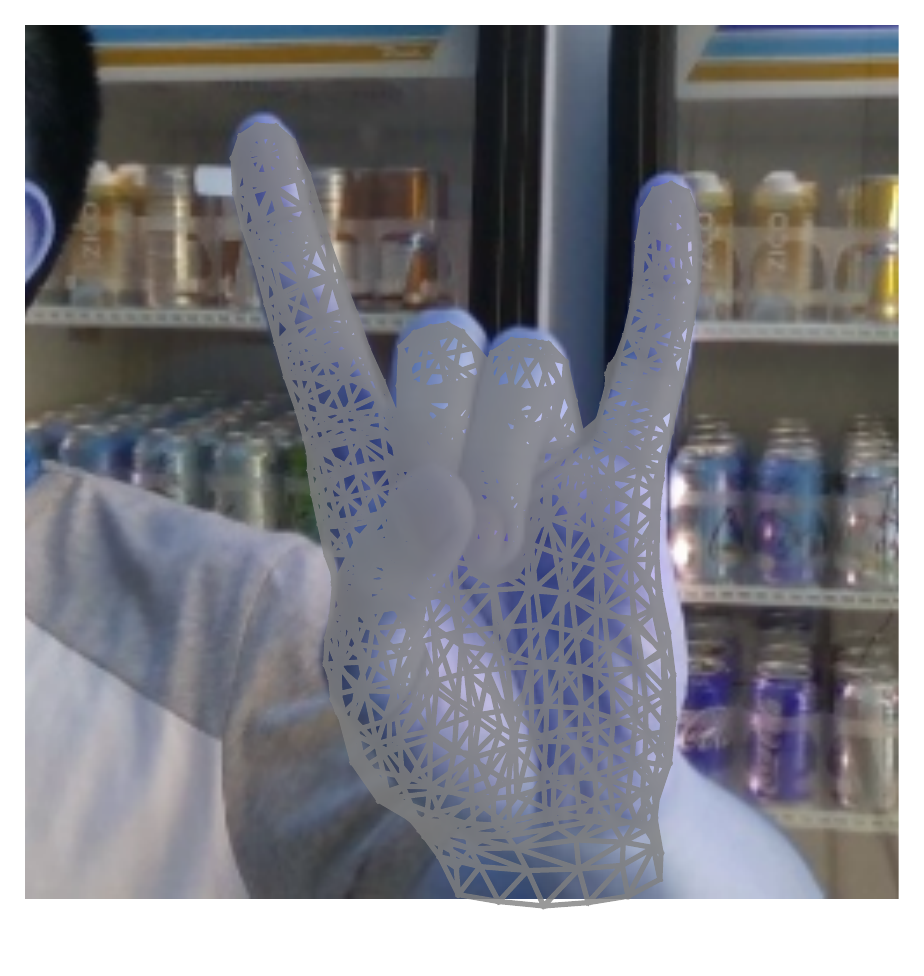}
  \end{subfigure}
  \begin{subfigure}{0.09\linewidth}
    \includegraphics[width=\linewidth]{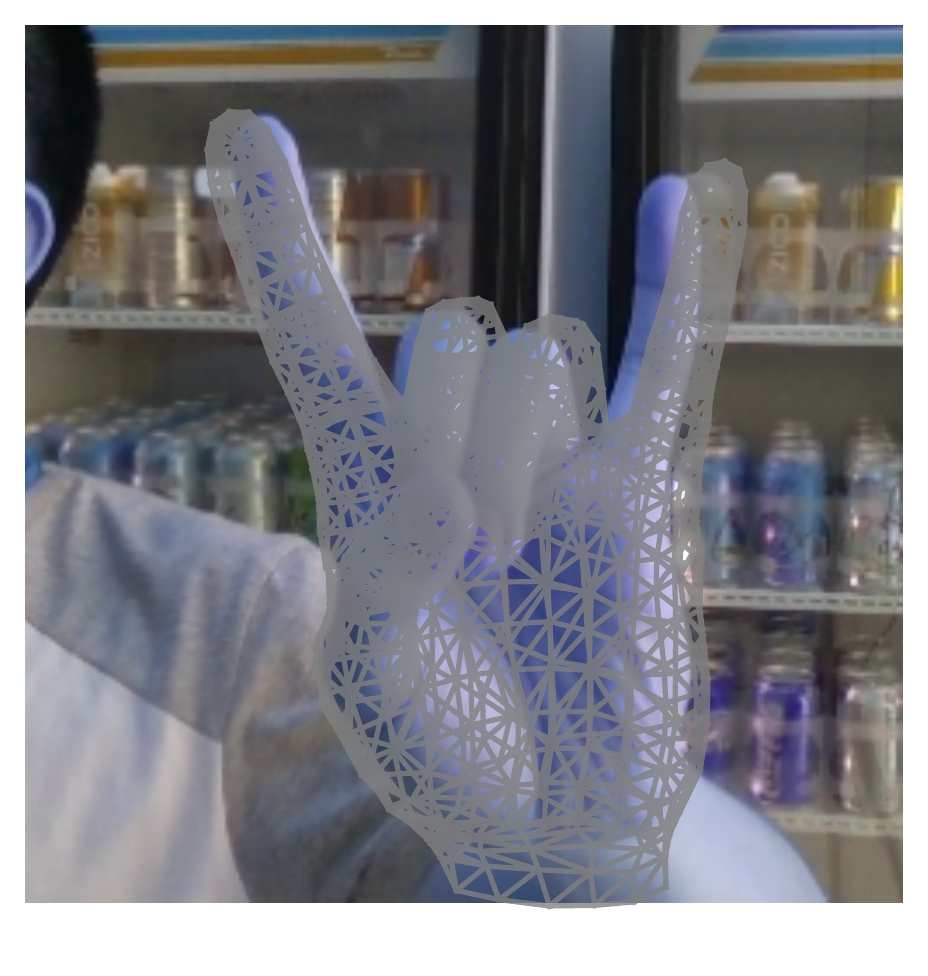}
  \end{subfigure}
  \begin{subfigure}{0.09\linewidth}
    \includegraphics[width=\linewidth]{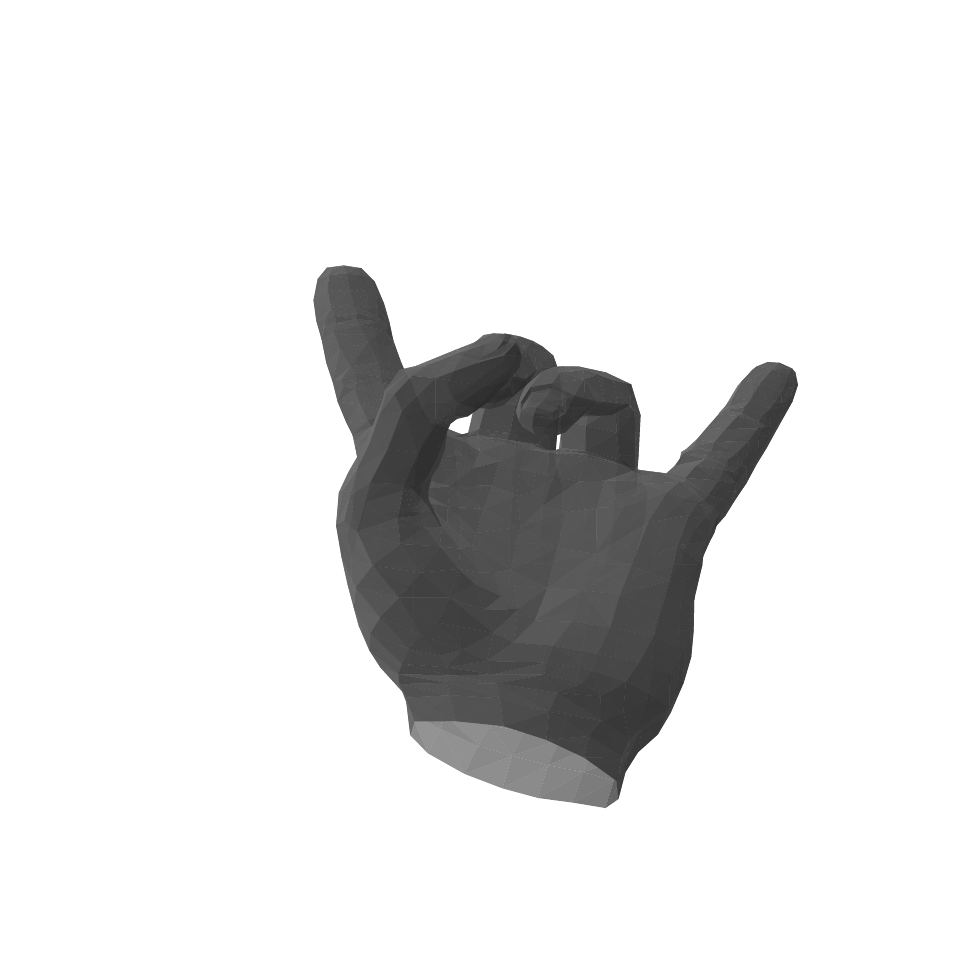}
  \end{subfigure}
  \begin{subfigure}{0.09\linewidth}
    \includegraphics[width=\linewidth]{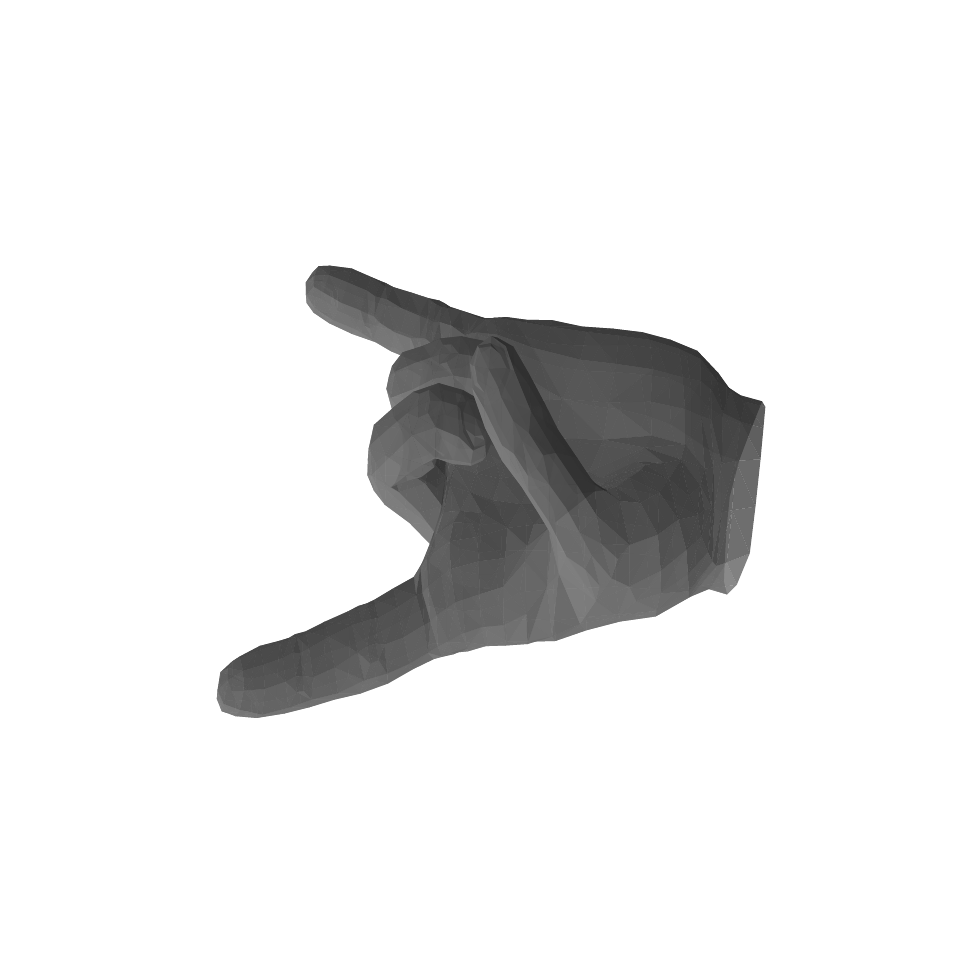}
  \end{subfigure}
  \begin{subfigure}{0.09\linewidth}
    \includegraphics[width=\linewidth]{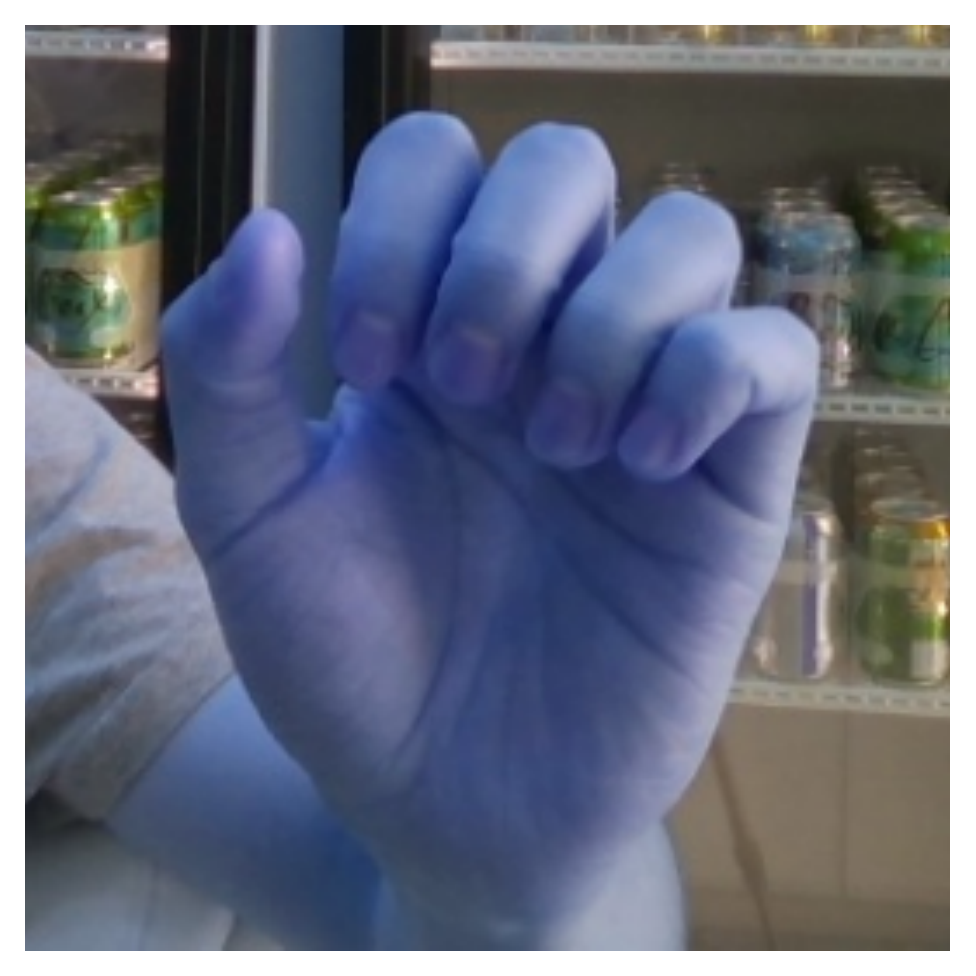}
  \end{subfigure}
  \begin{subfigure}{0.09\linewidth}
    \includegraphics[width=\linewidth]{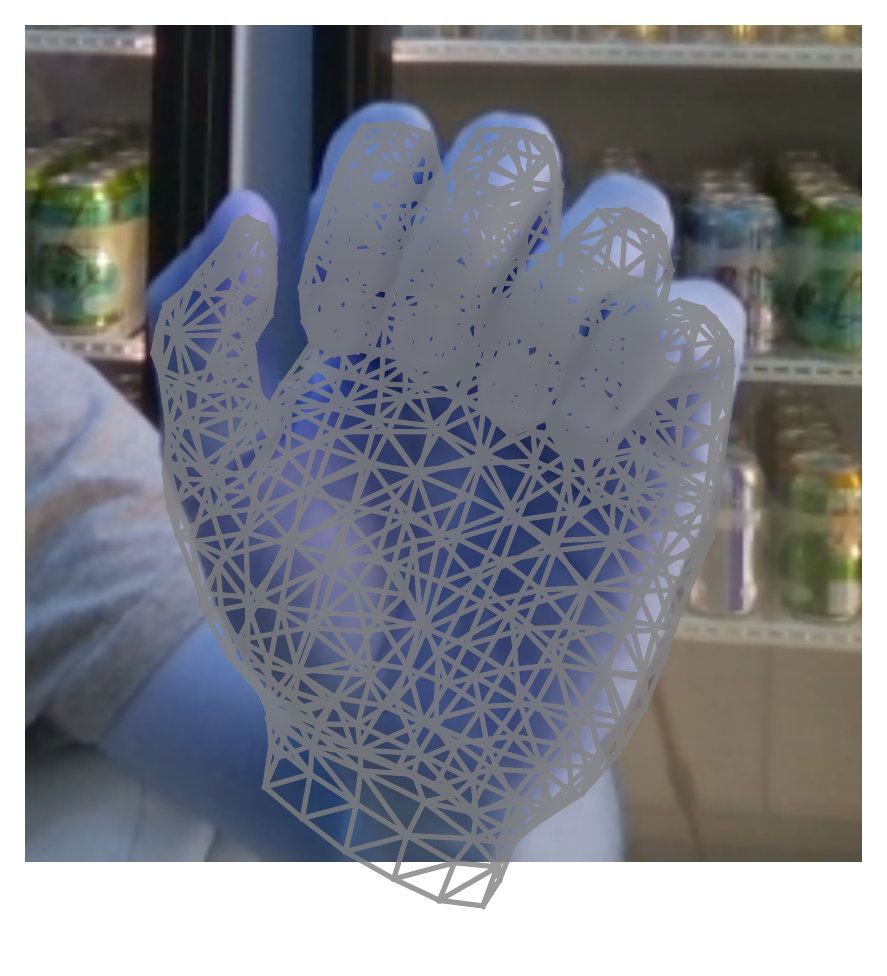}
  \end{subfigure}
  \begin{subfigure}{0.09\linewidth}
    \includegraphics[width=\linewidth]{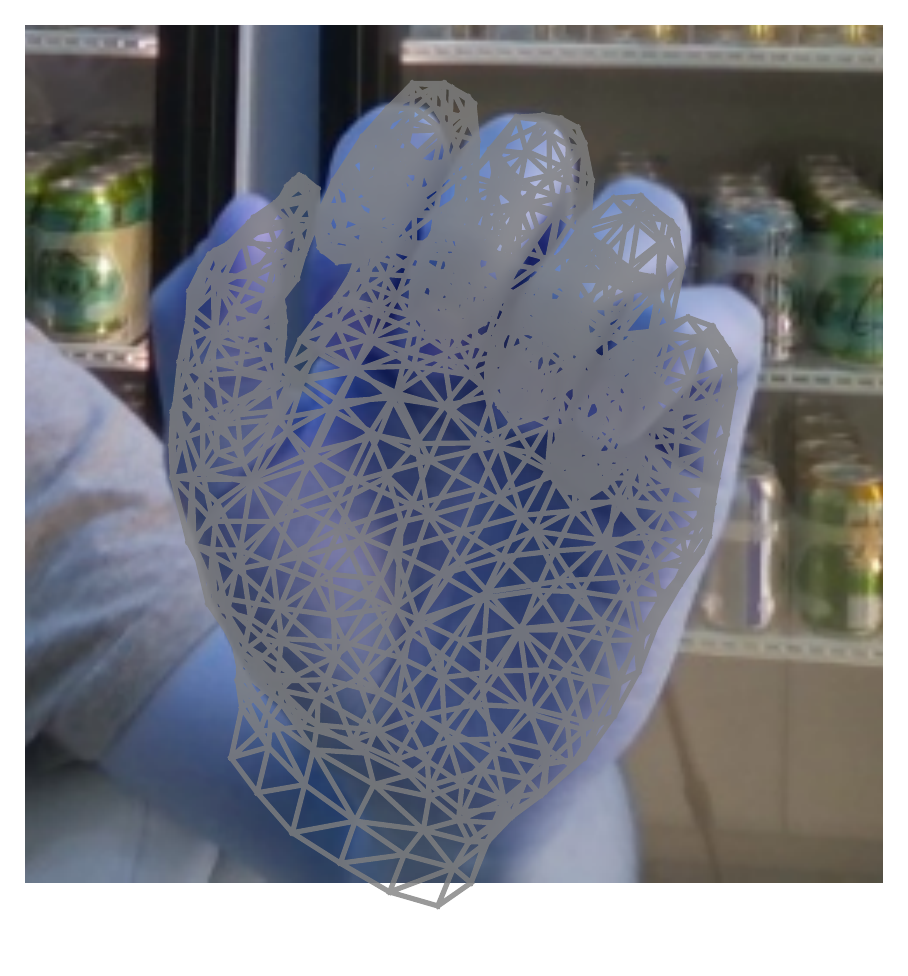}
  \end{subfigure}
  \begin{subfigure}{0.09\linewidth}
    \includegraphics[width=\linewidth]{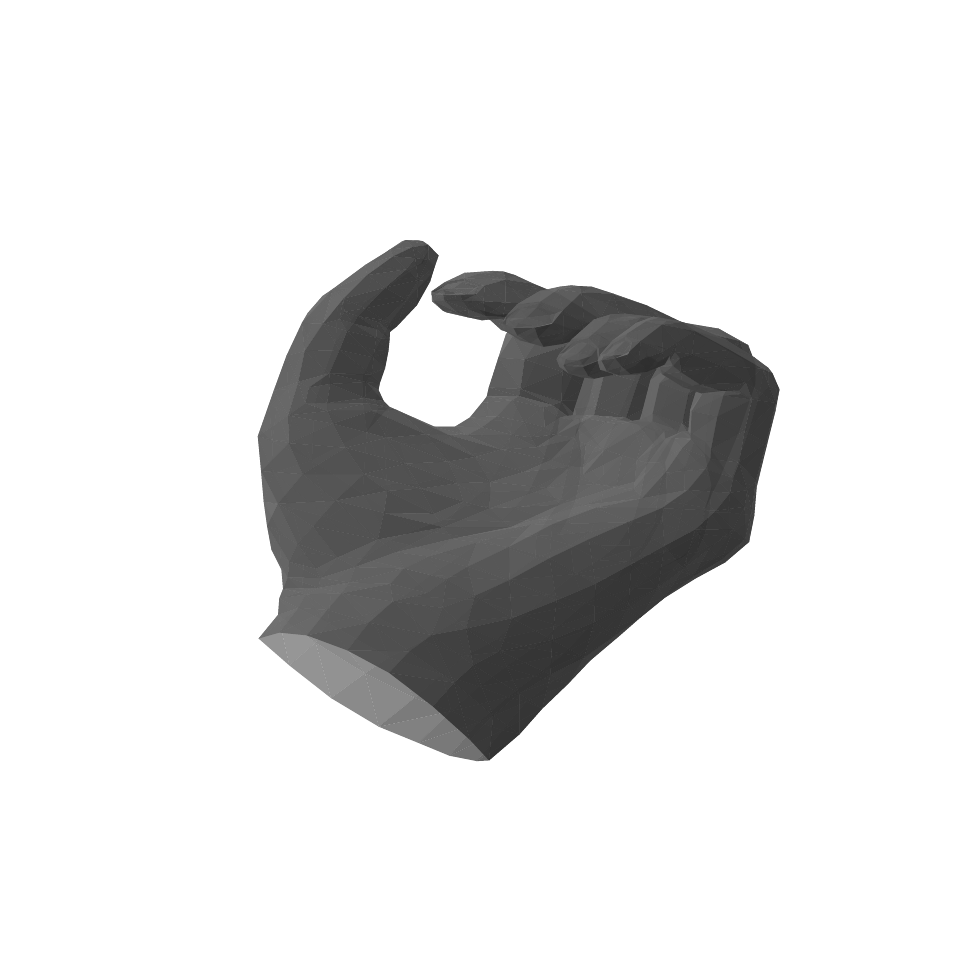}
  \end{subfigure}
  \begin{subfigure}{0.09\linewidth}
    \includegraphics[width=\linewidth]{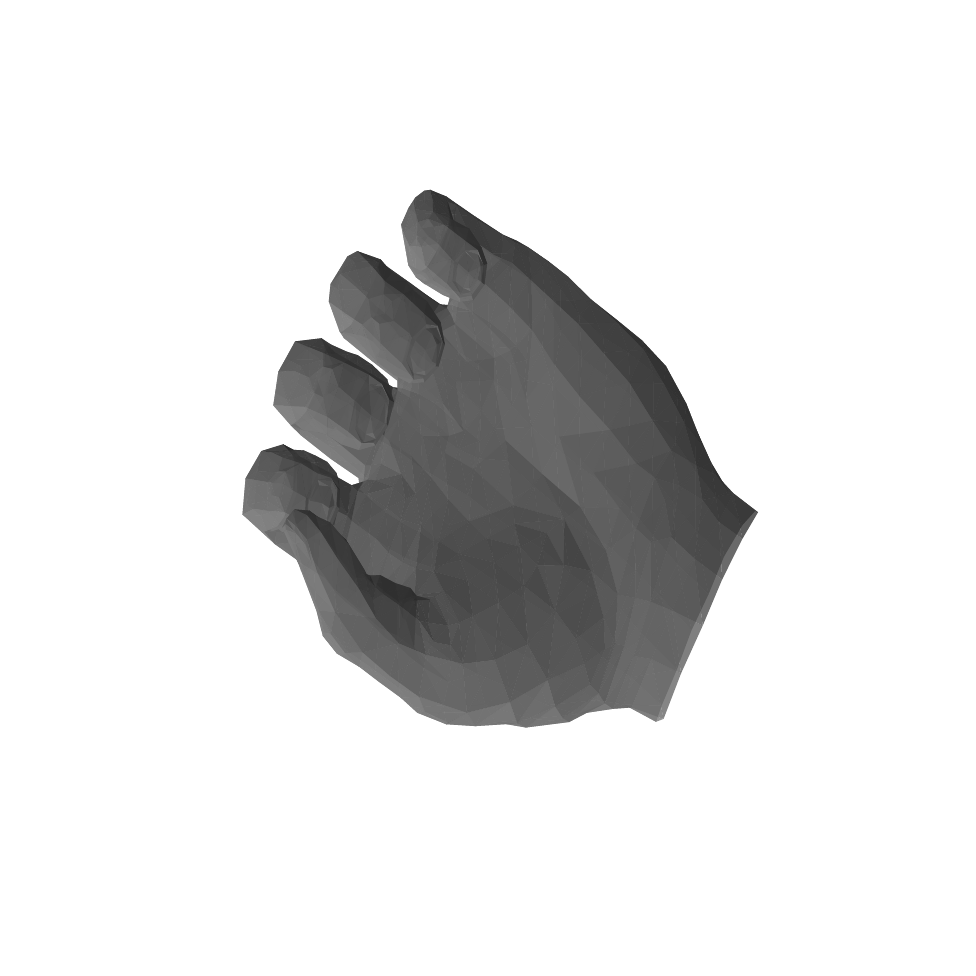}
  \end{subfigure}
  
  \begin{subfigure}{0.09\linewidth}
    \includegraphics[width=\linewidth]{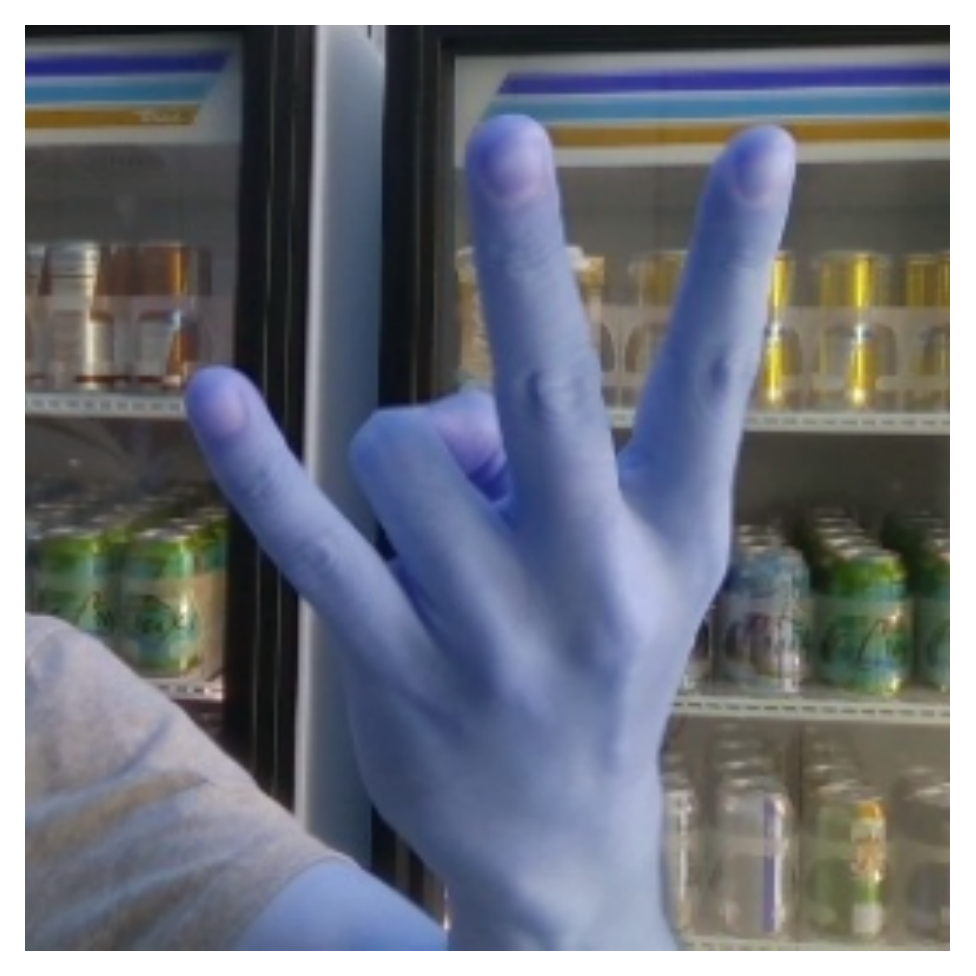}
  \end{subfigure}
  \begin{subfigure}{0.09\linewidth}
    \includegraphics[width=\linewidth]{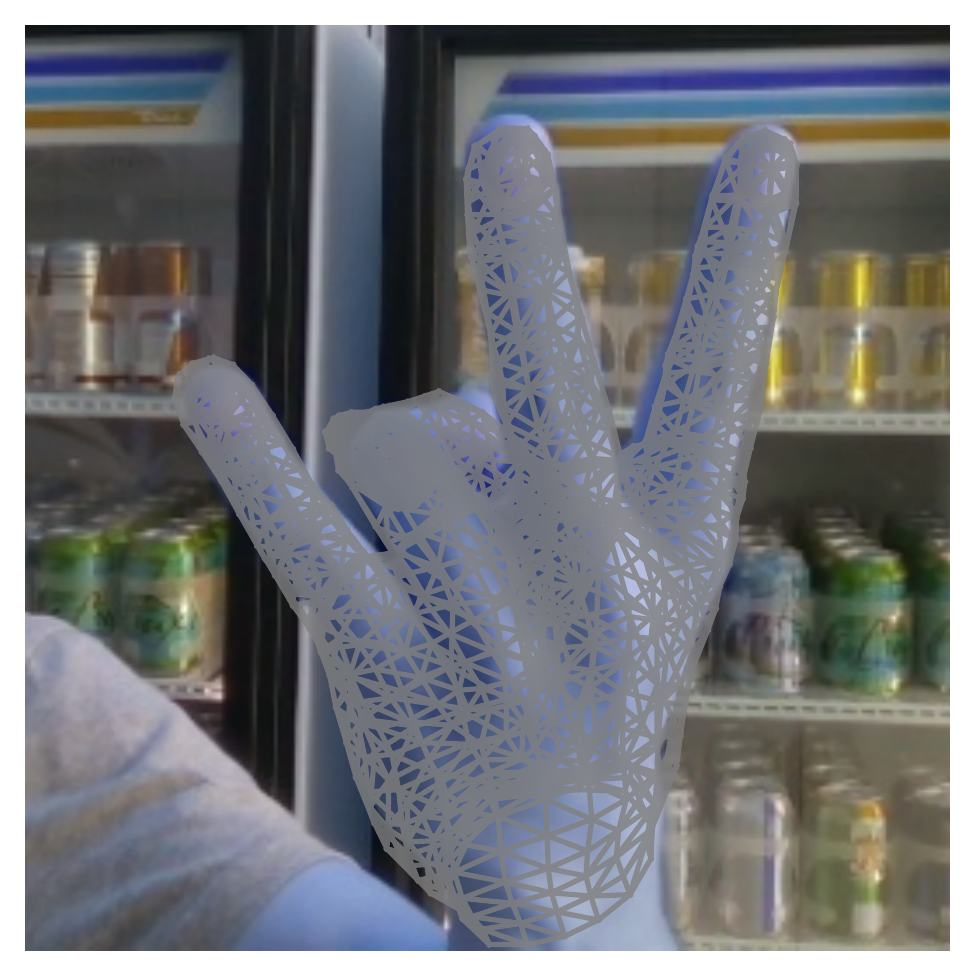}
  \end{subfigure}
  \begin{subfigure}{0.09\linewidth}
    \includegraphics[width=\linewidth]{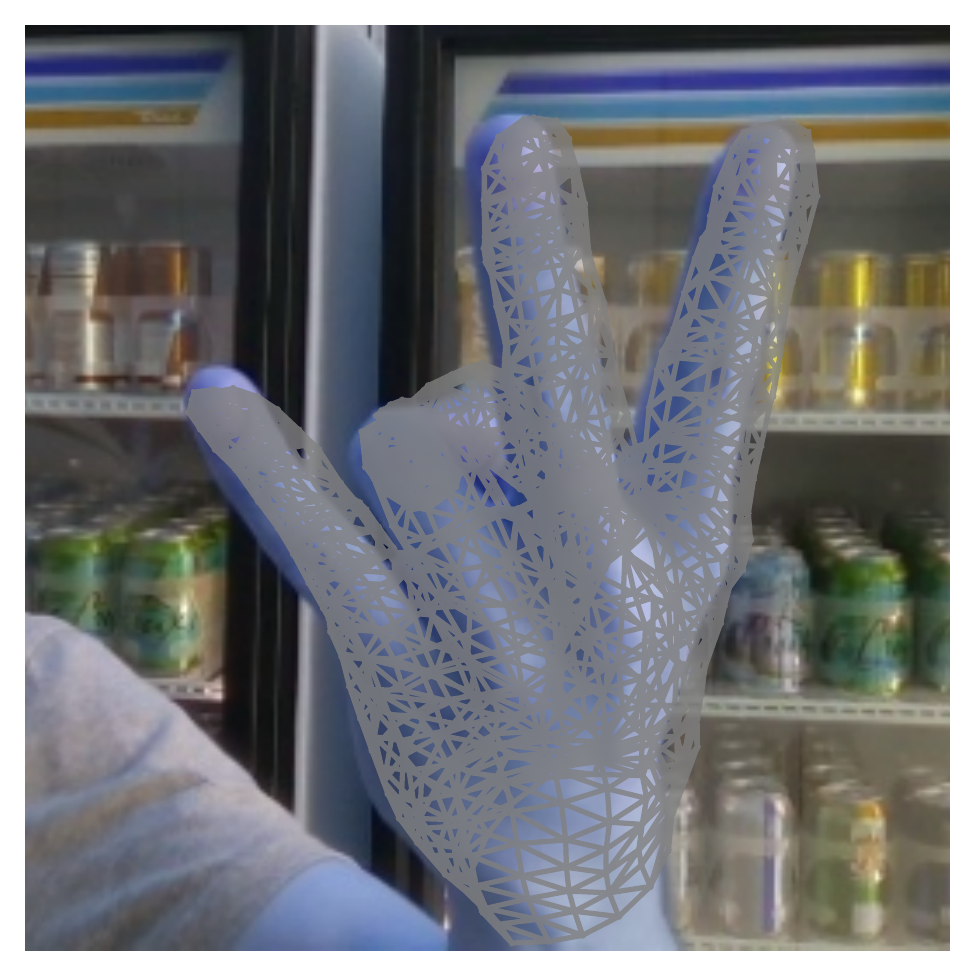}
  \end{subfigure}
  \begin{subfigure}{0.09\linewidth}
    \includegraphics[width=\linewidth]{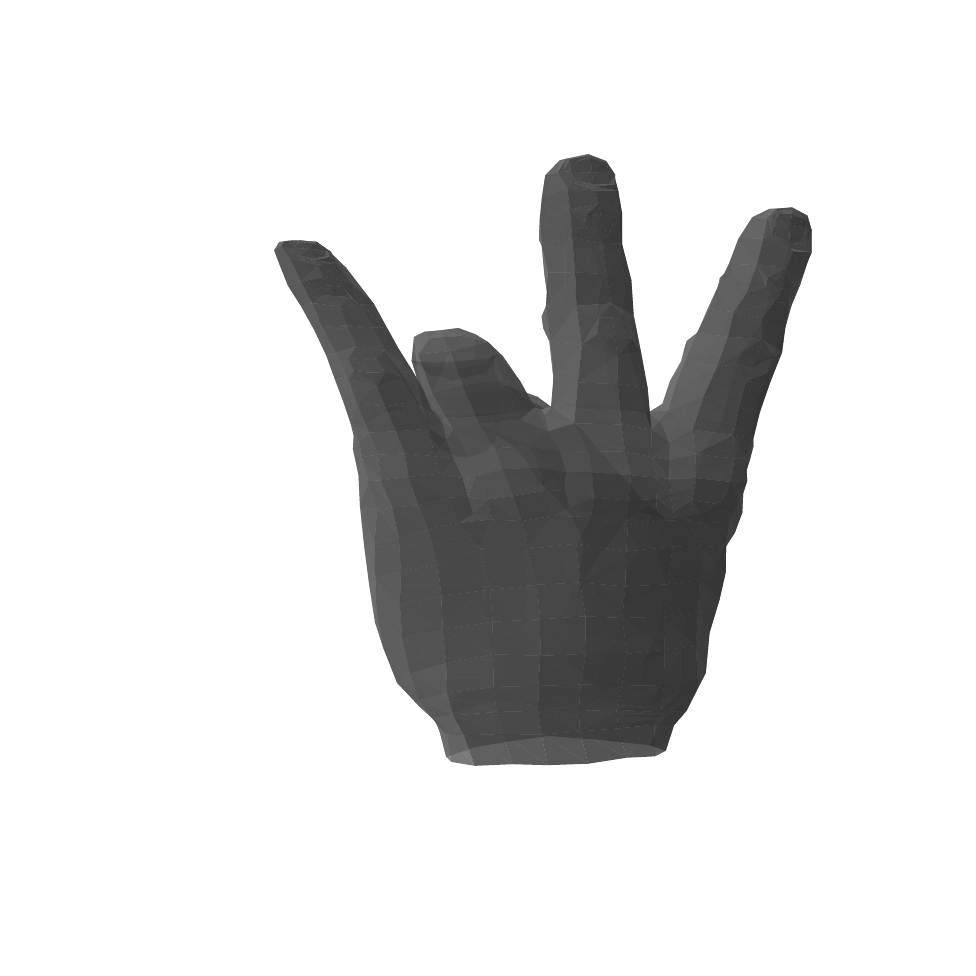}
  \end{subfigure}
  \begin{subfigure}{0.09\linewidth}
    \includegraphics[width=\linewidth]{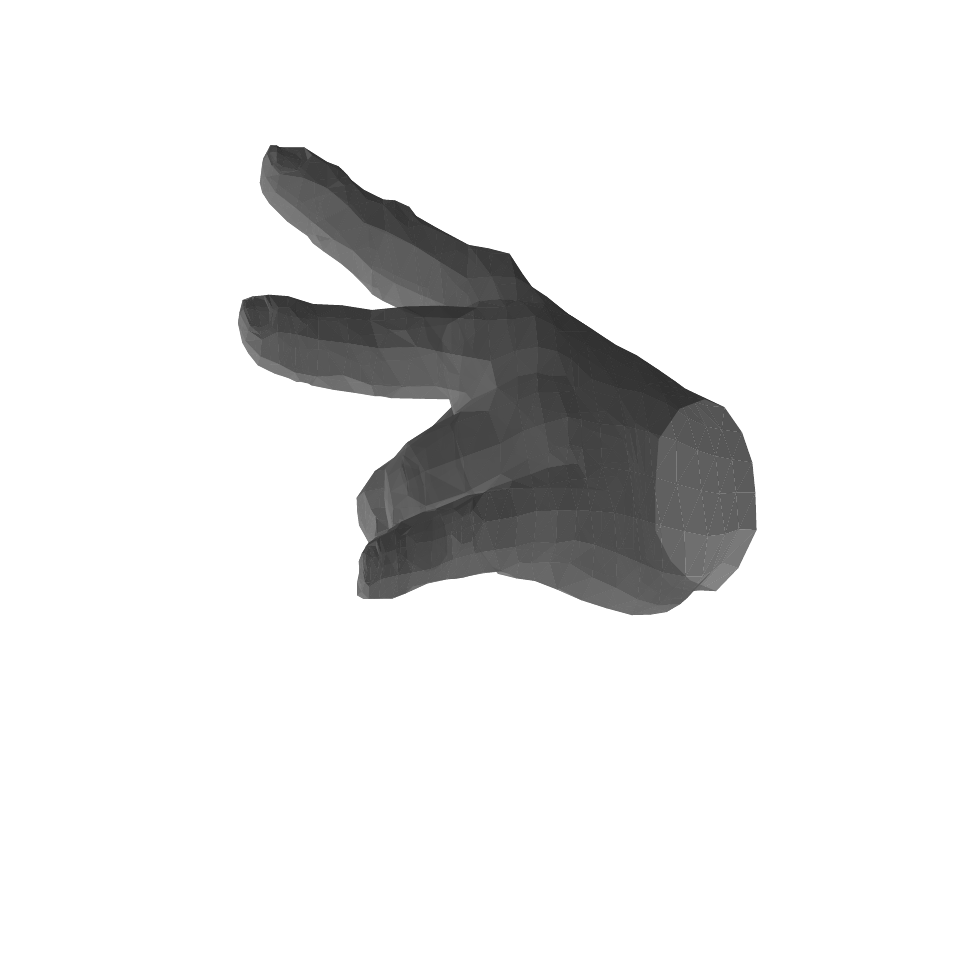}
  \end{subfigure}
  \begin{subfigure}{0.09\linewidth}
    \includegraphics[width=\linewidth]{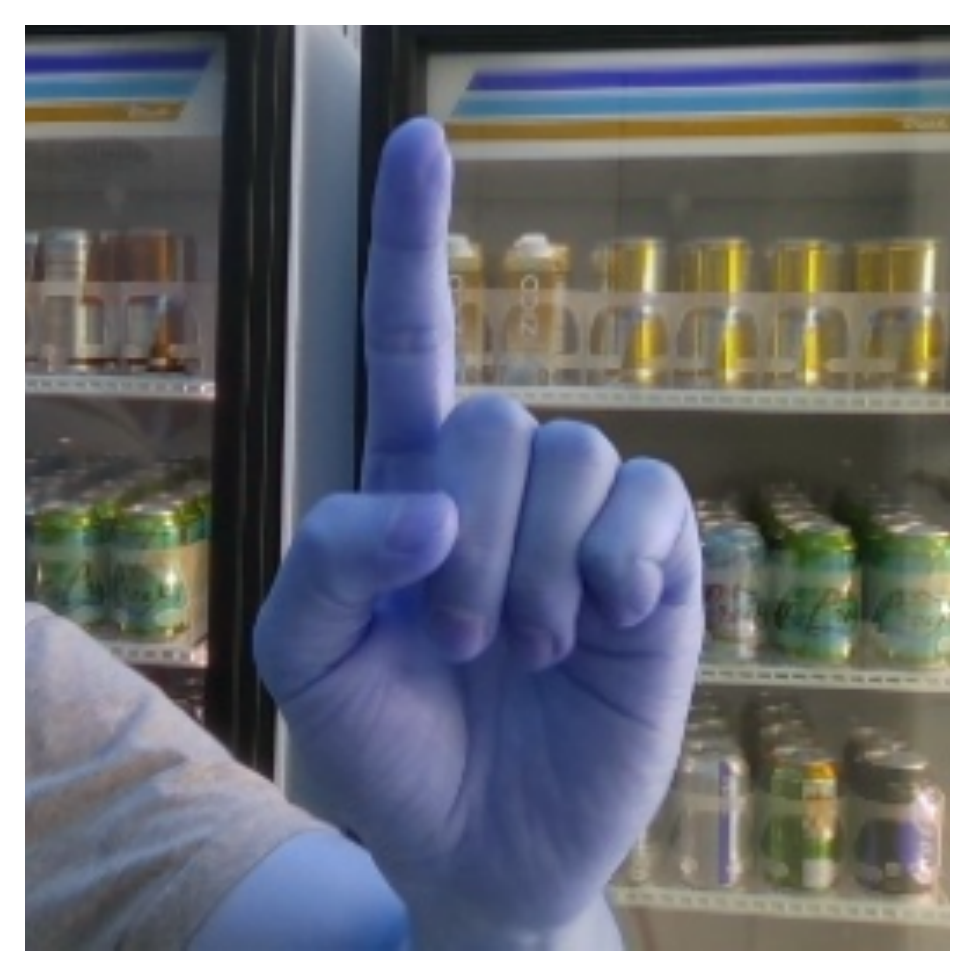}
  \end{subfigure}
  \begin{subfigure}{0.09\linewidth}
    \includegraphics[width=\linewidth]{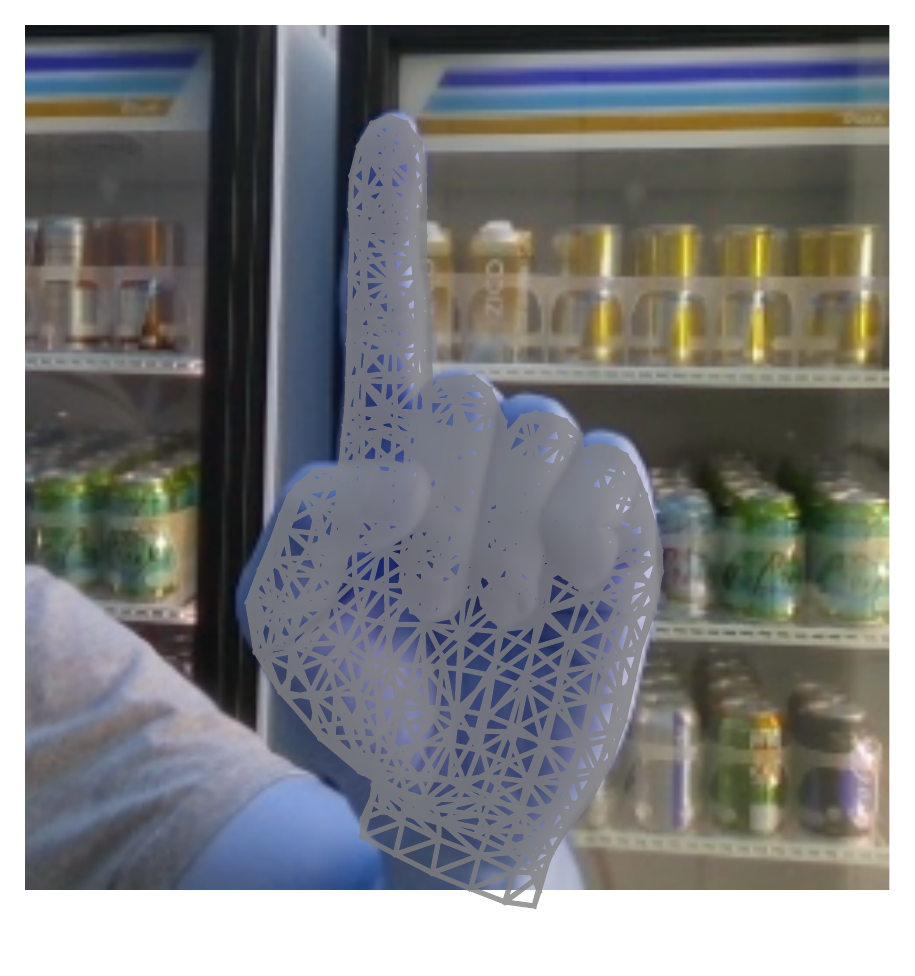}
  \end{subfigure}
  \begin{subfigure}{0.09\linewidth}
    \includegraphics[width=\linewidth]{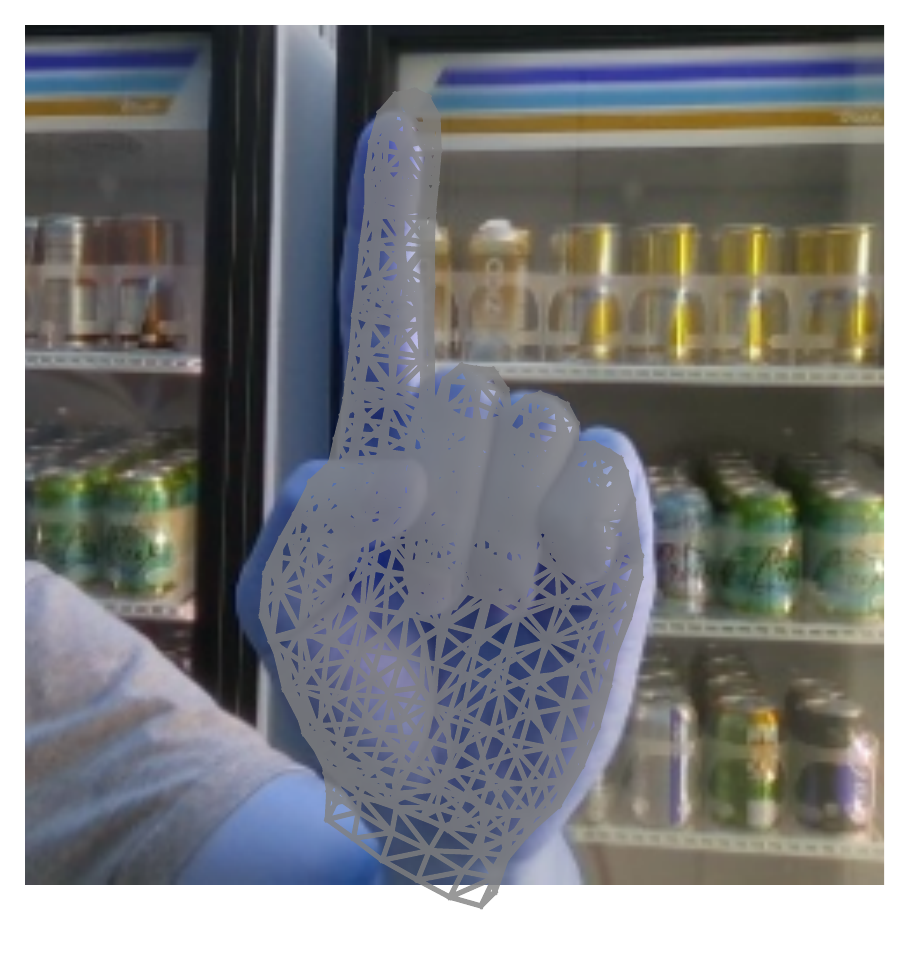}
  \end{subfigure}
  \begin{subfigure}{0.09\linewidth}
    \includegraphics[width=\linewidth]{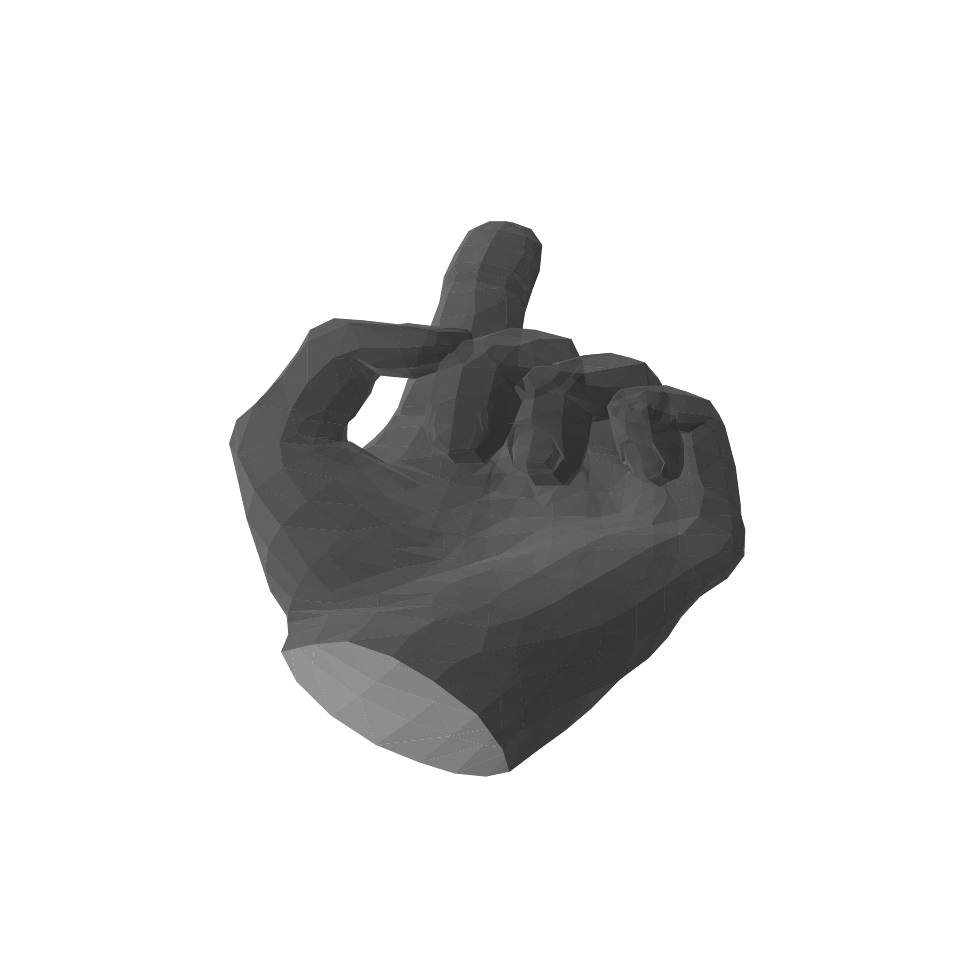}
  \end{subfigure}
  \begin{subfigure}{0.09\linewidth}
    \includegraphics[width=\linewidth]{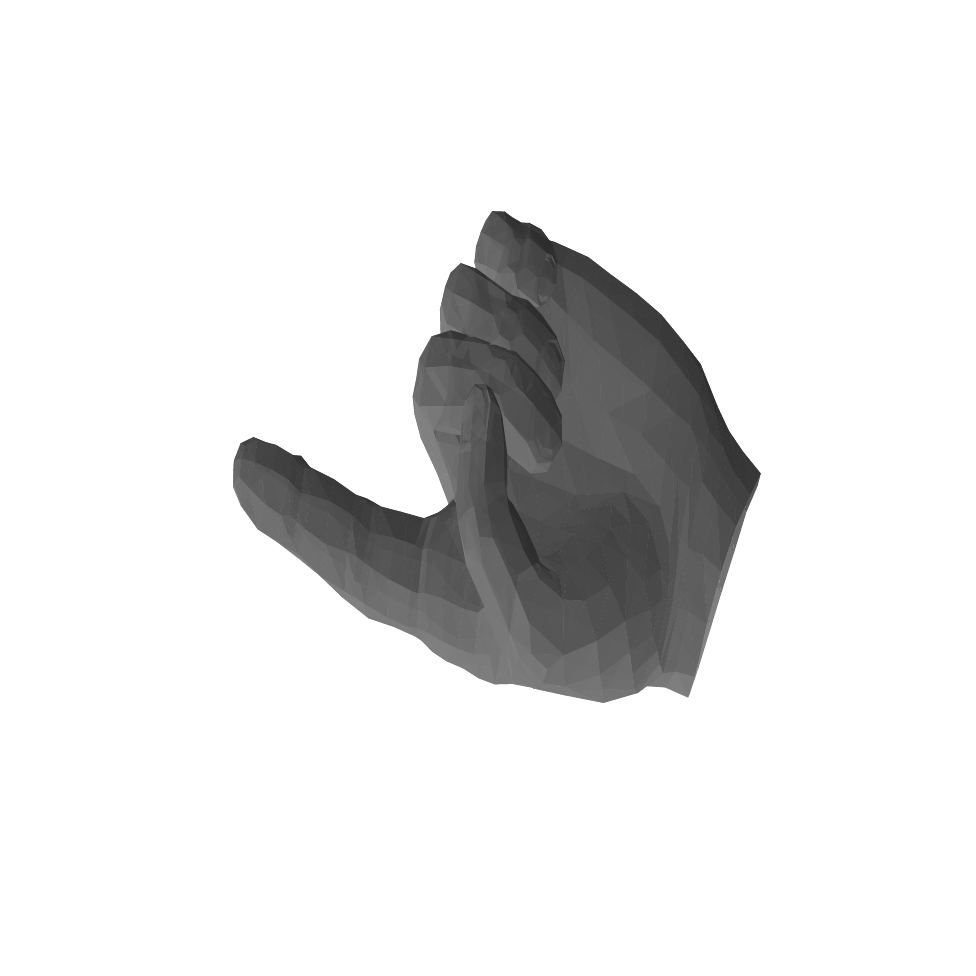}
  \end{subfigure}
  
  \begin{subfigure}{0.09\linewidth}
    \includegraphics[width=\linewidth]{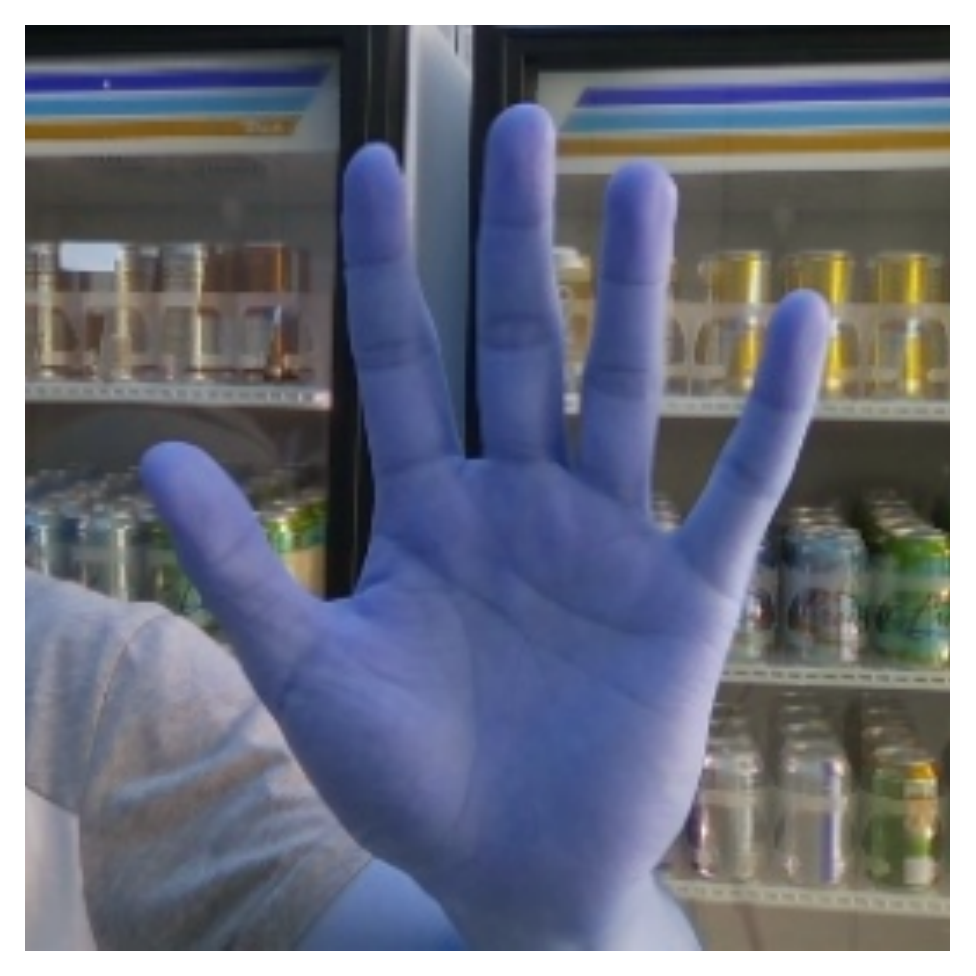}
    \centering
    \caption*{Input}
  \end{subfigure}
  \begin{subfigure}{0.09\linewidth}
    \includegraphics[width=\linewidth]{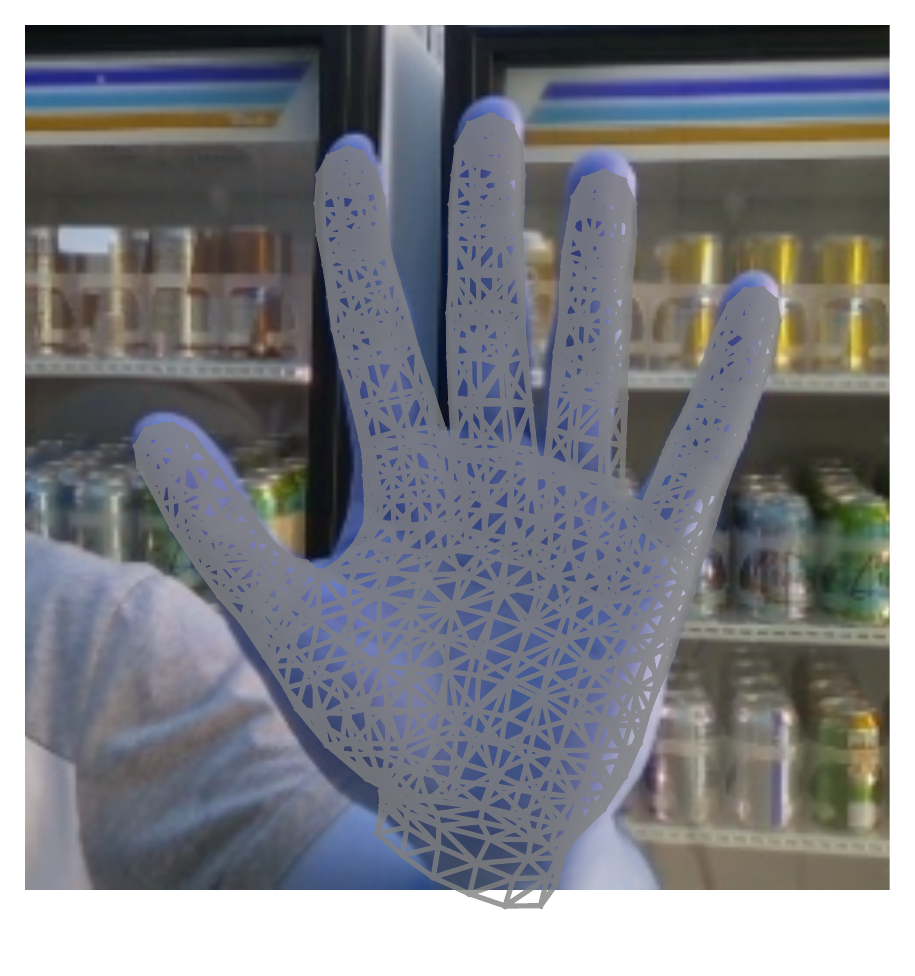}
    \centering
    \caption*{Target}
  \end{subfigure}
  \begin{subfigure}{0.09\linewidth}
    \includegraphics[width=\linewidth]{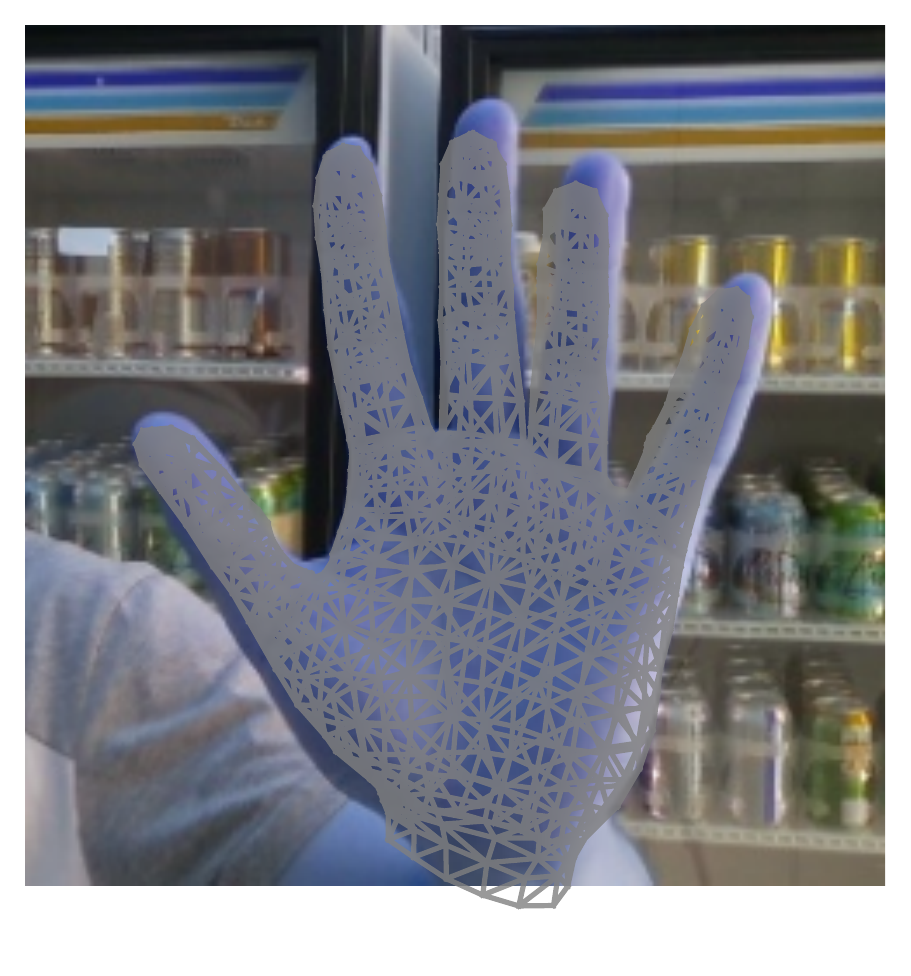}
    \centering
    \caption*{Prediction}
  \end{subfigure}
  \begin{subfigure}{0.09\linewidth}
    \includegraphics[width=\linewidth]{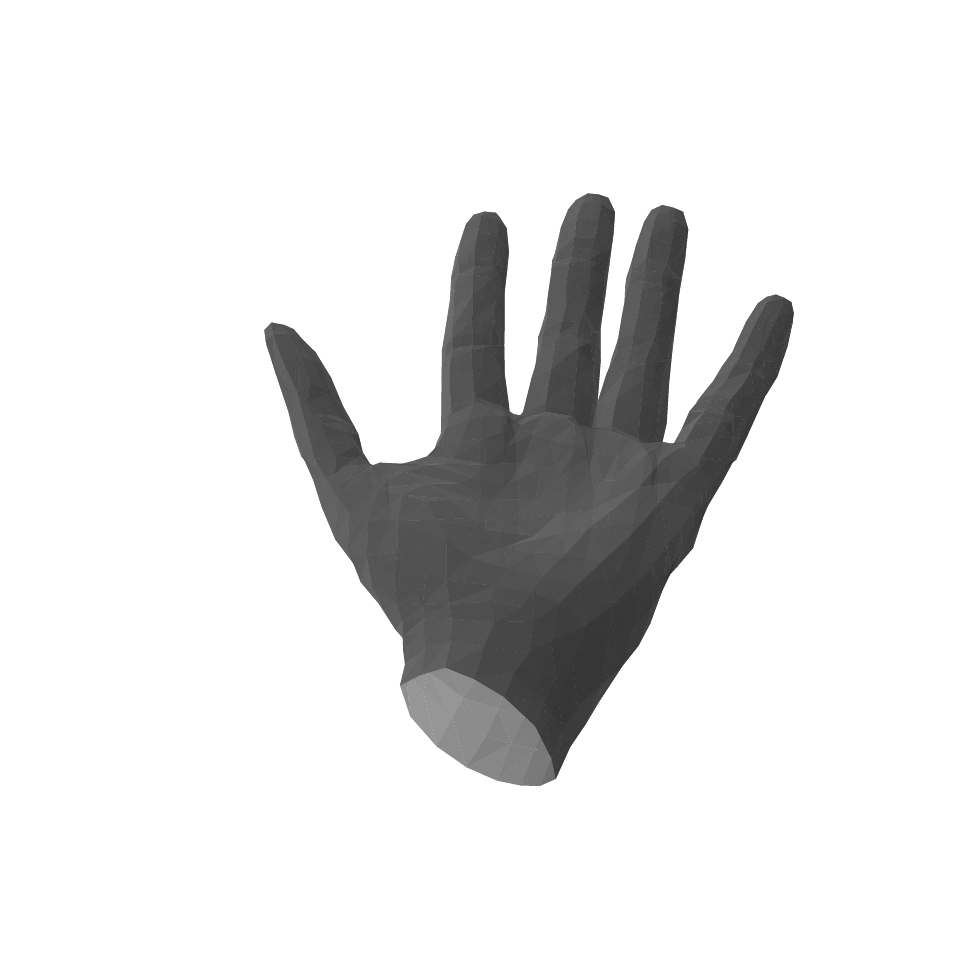}
    \centering
    \caption*{View A}
  \end{subfigure}
  \begin{subfigure}{0.09\linewidth}
    \includegraphics[width=\linewidth]{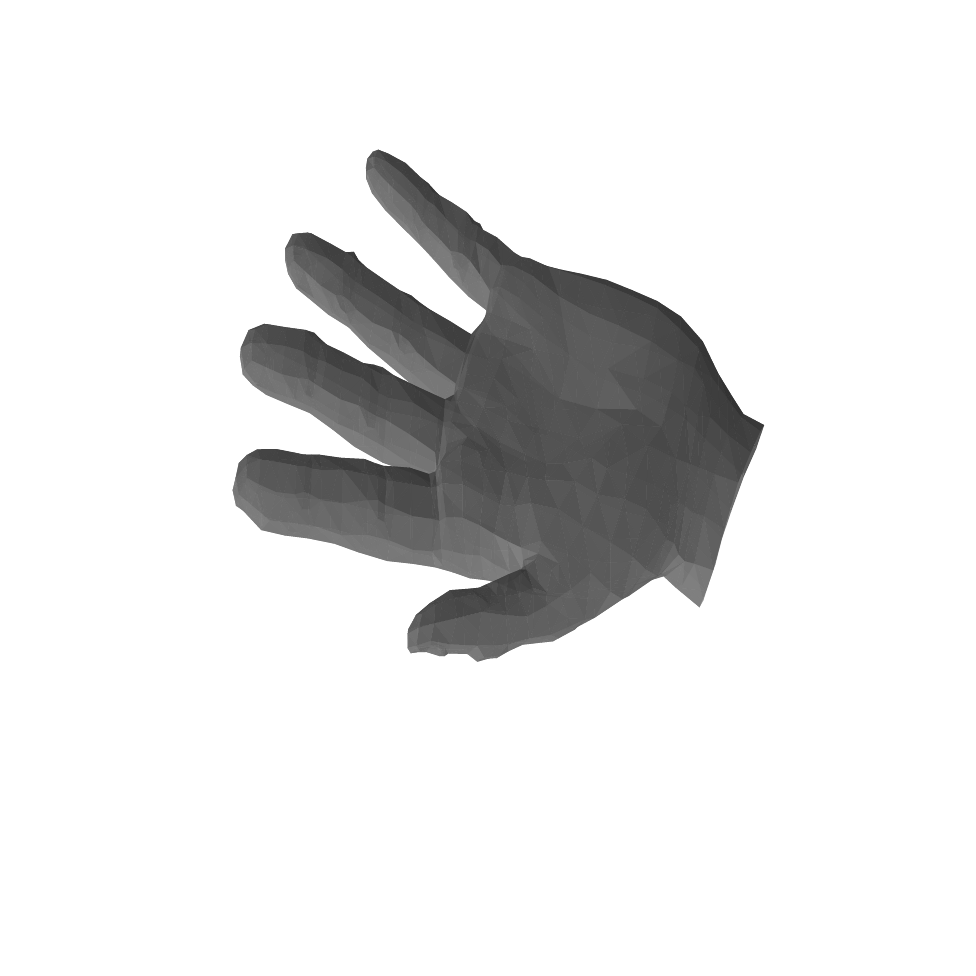}
    \centering
    \caption*{View B}
  \end{subfigure}
  \begin{subfigure}{0.09\linewidth}
    \includegraphics[width=\linewidth]{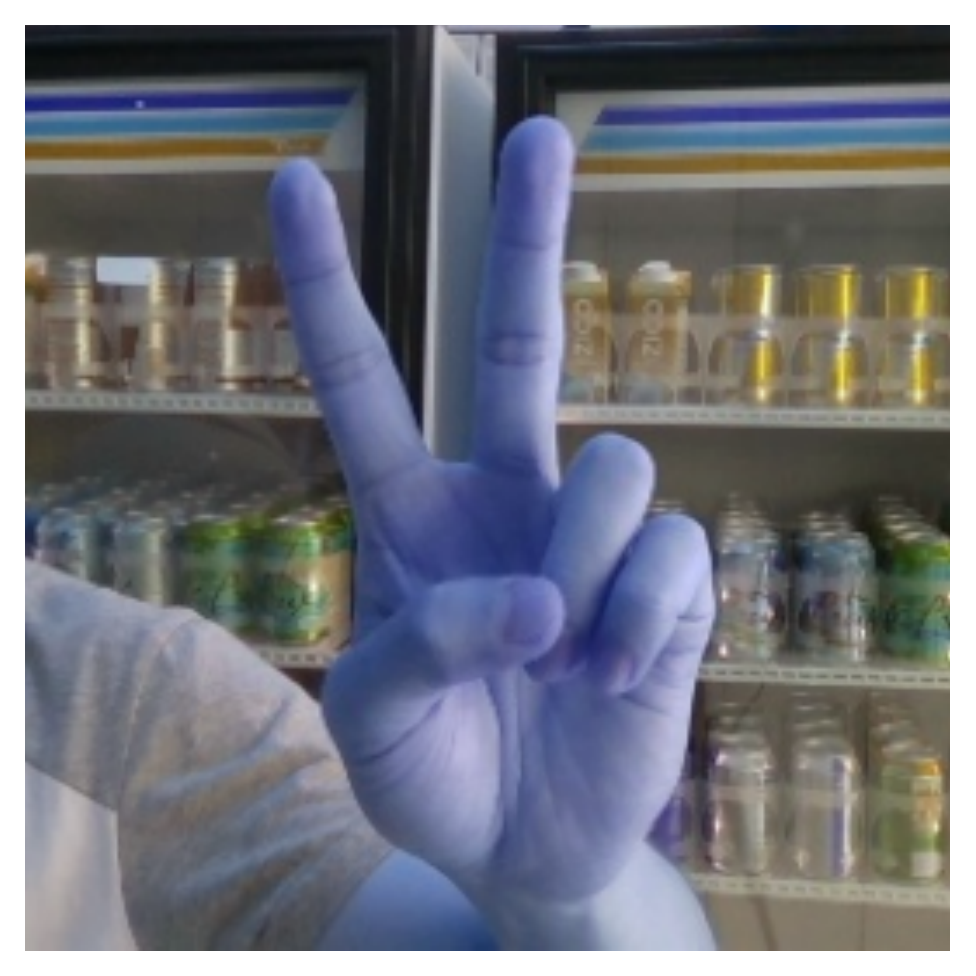}
    \centering
    \caption*{Input}
  \end{subfigure}
  \begin{subfigure}{0.09\linewidth}
    \includegraphics[width=\linewidth]{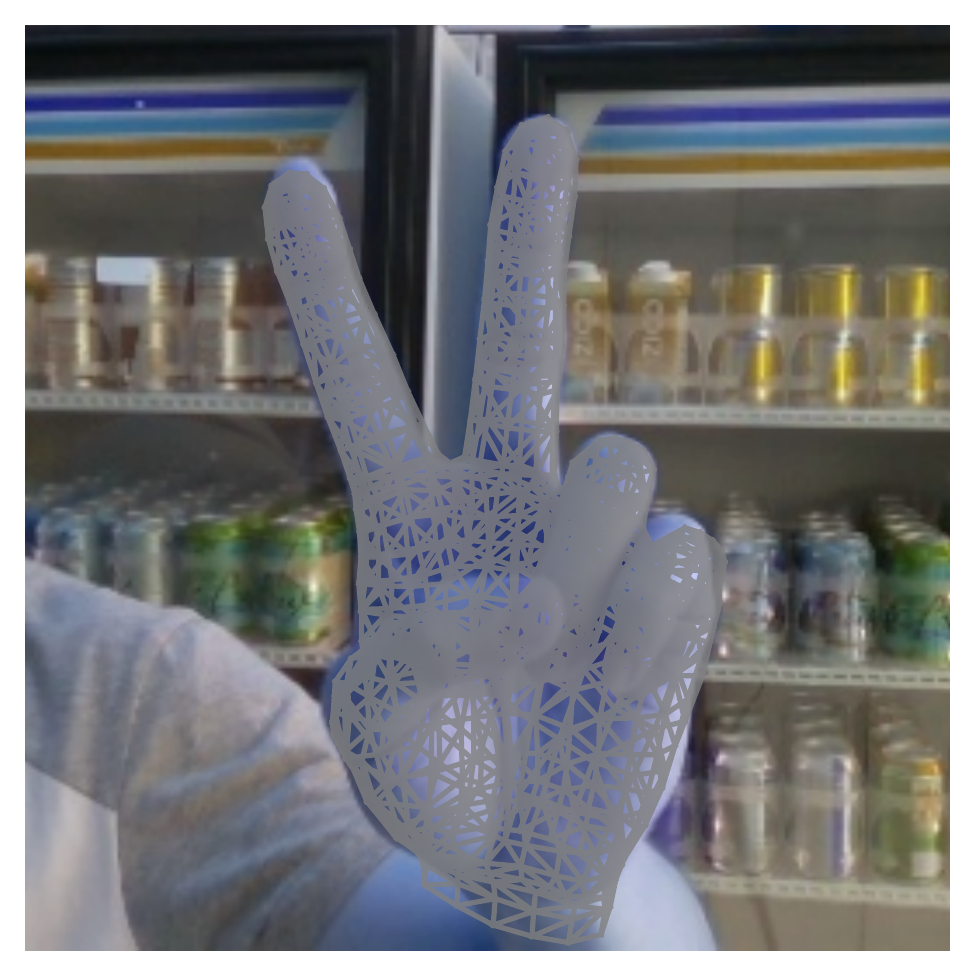}
    \centering
    \caption*{Target}
  \end{subfigure}
  \begin{subfigure}{0.09\linewidth}
    \includegraphics[width=\linewidth]{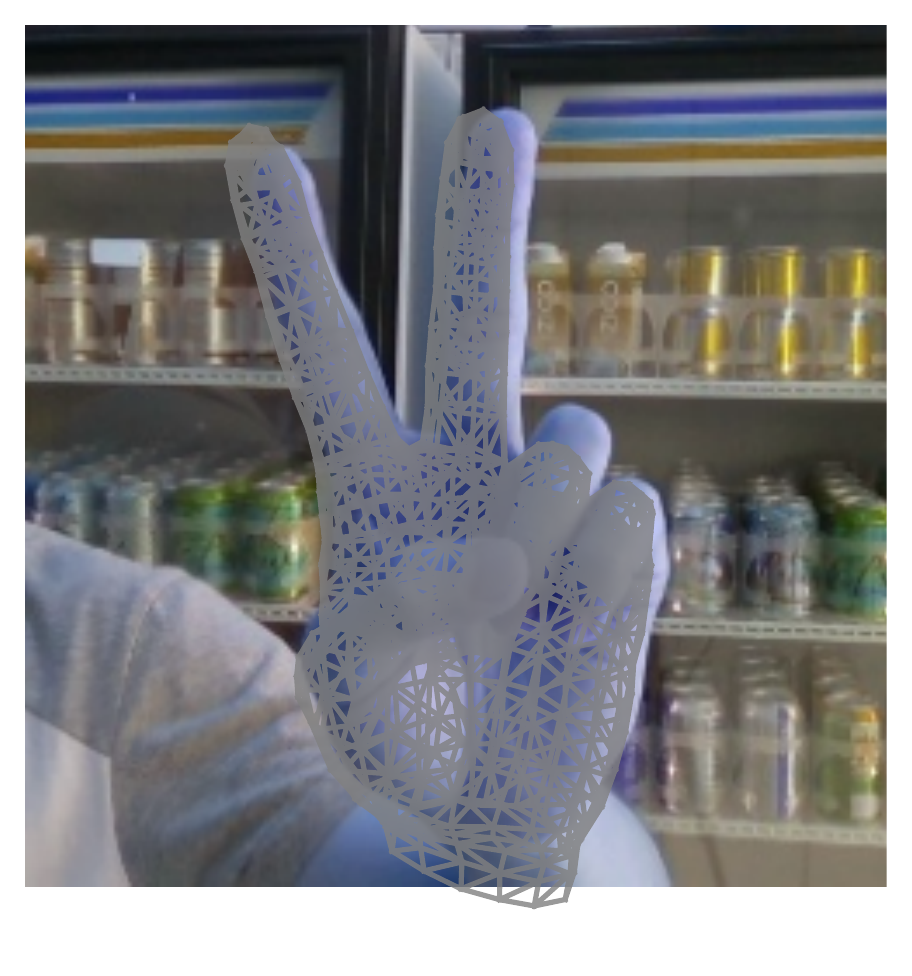}
    \centering
    \caption*{Prediction}
  \end{subfigure}
  \begin{subfigure}{0.09\linewidth}
    \includegraphics[width=\linewidth]{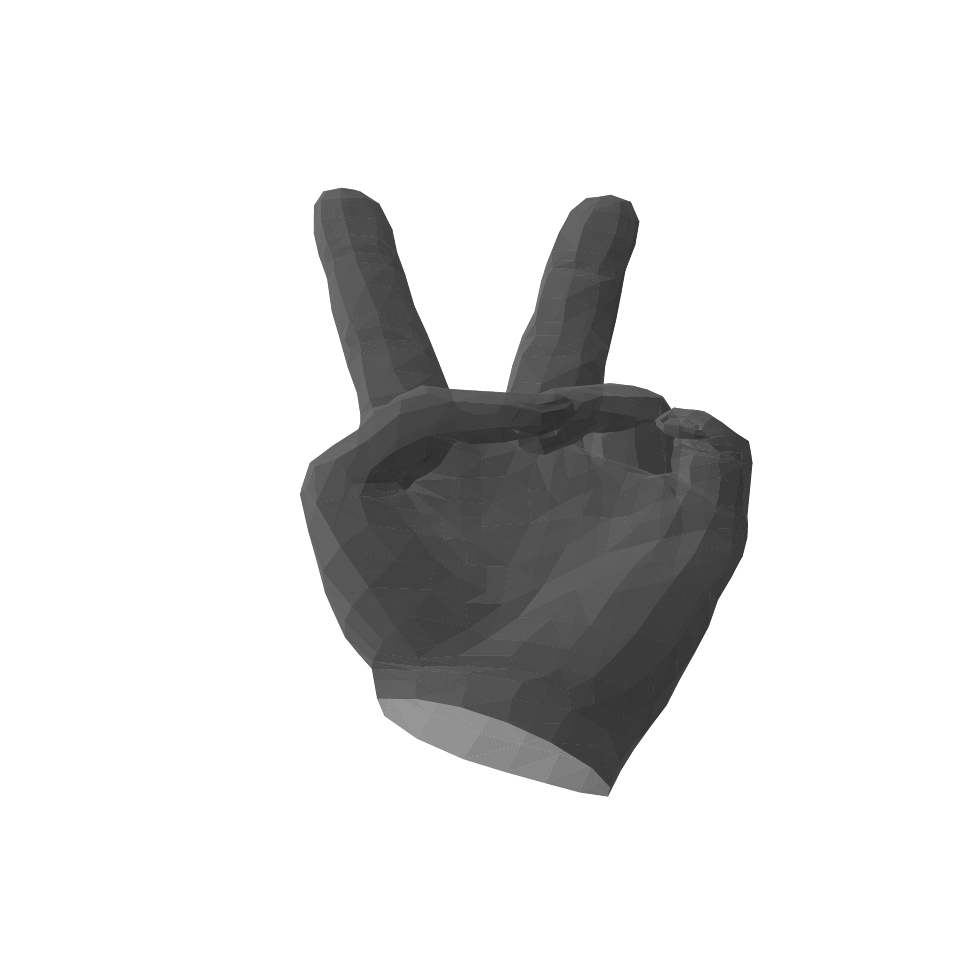}
    \centering
    \caption*{View A}
  \end{subfigure}
  \begin{subfigure}{0.09\linewidth}
    \includegraphics[width=\linewidth]{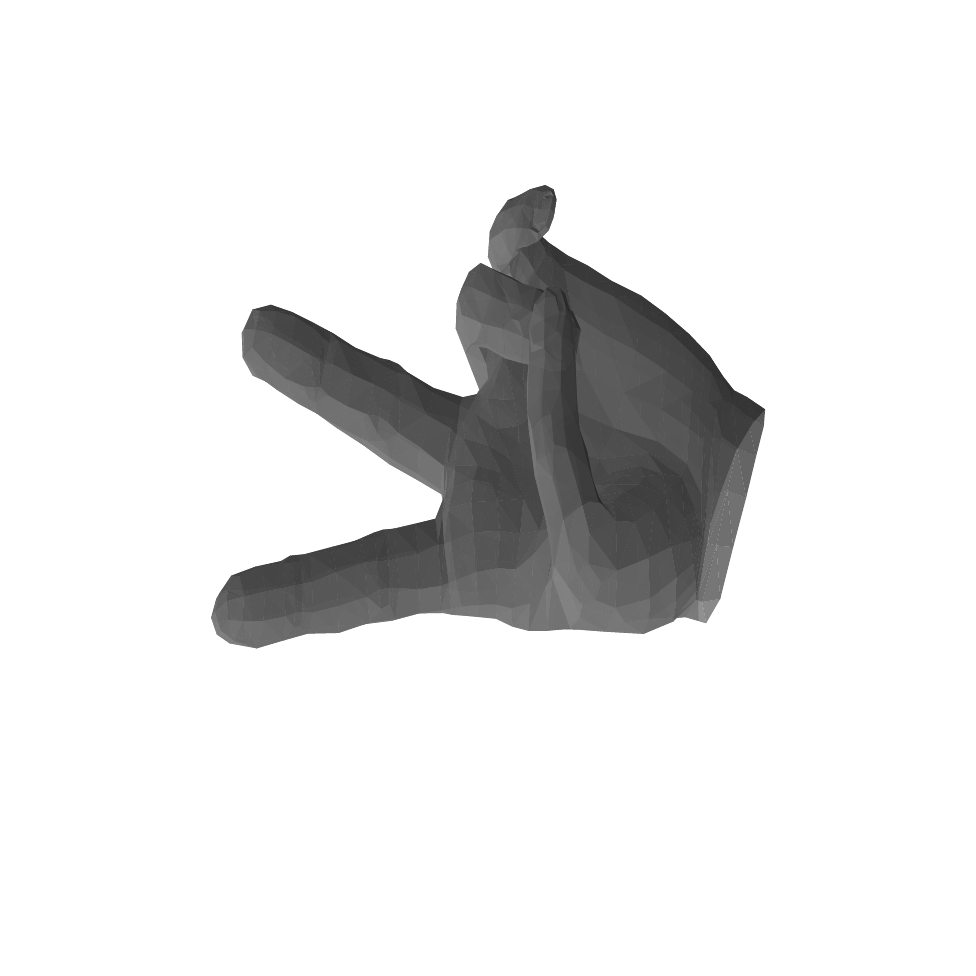}
    \centering
    \caption*{View B}
  \end{subfigure}
  
  \caption{Qualitative mesh reconstruction results on the validation dataset from the real-world dataset of \cite{ge20193d} comparing our model's predictions to the target meshes.}
  \label{fig:meshs}
\end{figure*}

In addition to the performance of the three networks, we present the number of trainable parameters. Table~\ref{tab:params} Shows that overall our EMLP model has two orders of magnitude less trainable parameters than both the MLP and GNN model. Our model has significantly less parameters in the encoder due to both the improved weight sharing and containing only convolutional layers, where as the other two models have MLP layers at the end of the encoder to maintain similarity with previous work. Our model also has less trainable parameters in the reshaper. This is also due to the improved weight sharing through the use of $\mathrm{SO}(2)$-equivariance. In the decoder our model reduces the parameter count over the MLP model due to the weight sharing imposed by $\mathrm{SO}(3)$-equivariance. The inductive bias of locality causes the GNN decoder to have a similar number of parameters to the EMLP model.

\begin{table}[h]
  \centering
  \begin{tabular}{lcccc}
    \toprule
    Method & Encoder & Reshaper & Decoder & Total \\
    \midrule
    MLP  & 235.73 & 11.73  & 45.80 & 293.26 \\
    GNN  & 235.73 & 11.73 & 0.56  & 348.02 \\
    EMLP & 0.06   & 1.11   & 0.52  & 1.69  \\
    \bottomrule
  \end{tabular}
  \caption{Number of trainable parameters in each component of the model in millions.}
  \label{tab:params}
\end{table}

Further, we compare our rotation equivariant model to other prior works in Table~\ref{tab:mesherror}. This demonstrates that our method produces superior results to previous methods on a real-world validation dataset.

\begin{table}[h]
  \centering
  \begin{tabular}{@{}lc@{}}
    \toprule
    Method & Mesh error (mm) \\
    \midrule
    MANO-based \cite{romero2017embodied} & 20.86 \\
    Direct LBS \cite{cai2018weakly} & 13.33 \\
    Graph CNN \cite{ge20193d} & 12.72 \\
    Ours & 9.67 \\
    \bottomrule
  \end{tabular}
  \caption{Average mesh error tested on the validation set of the real-world dataset. Results for prior methods are taken from \cite{ge20193d}.}
  \label{tab:mesherror}
\end{table}

\newcommand\rotwidth{0.09}
\begin{figure*}
  \centering

  \begin{subfigure}{\rotwidth\linewidth}
    \includegraphics[width=\linewidth]{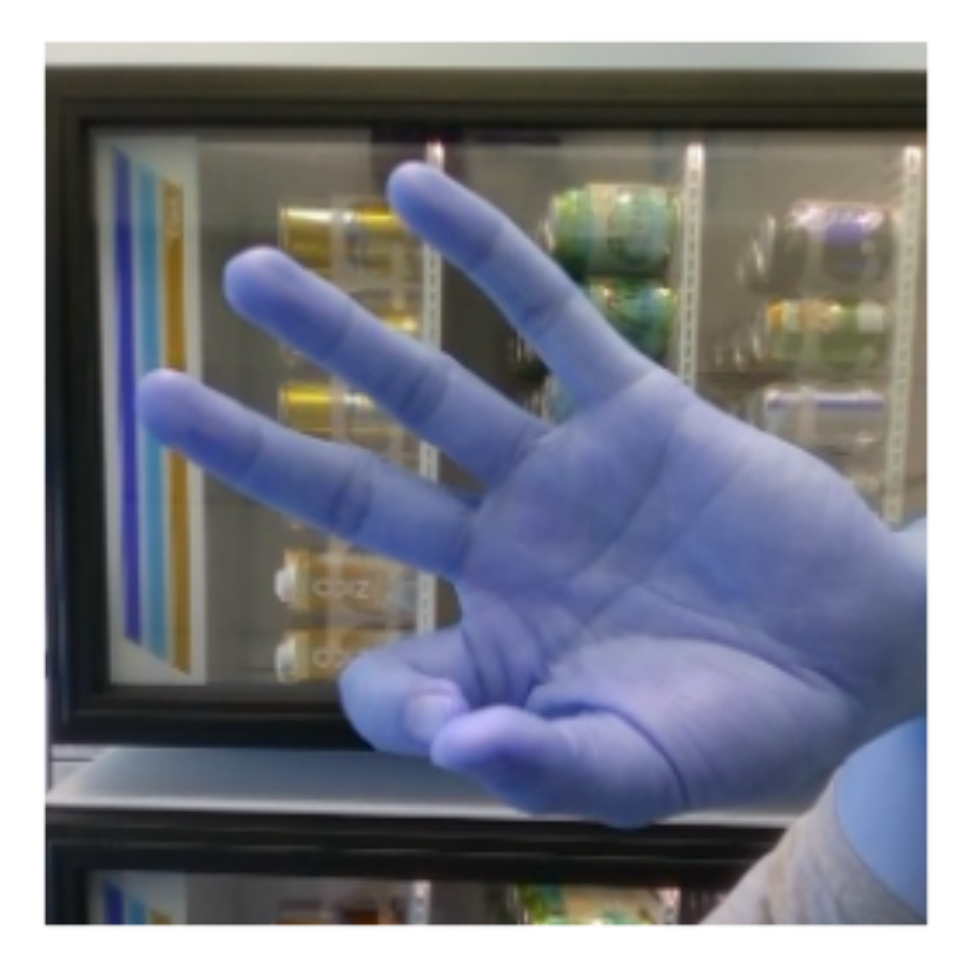}
    \centering
  \end{subfigure}
  \begin{subfigure}{\rotwidth\linewidth}
    \includegraphics[width=\linewidth]{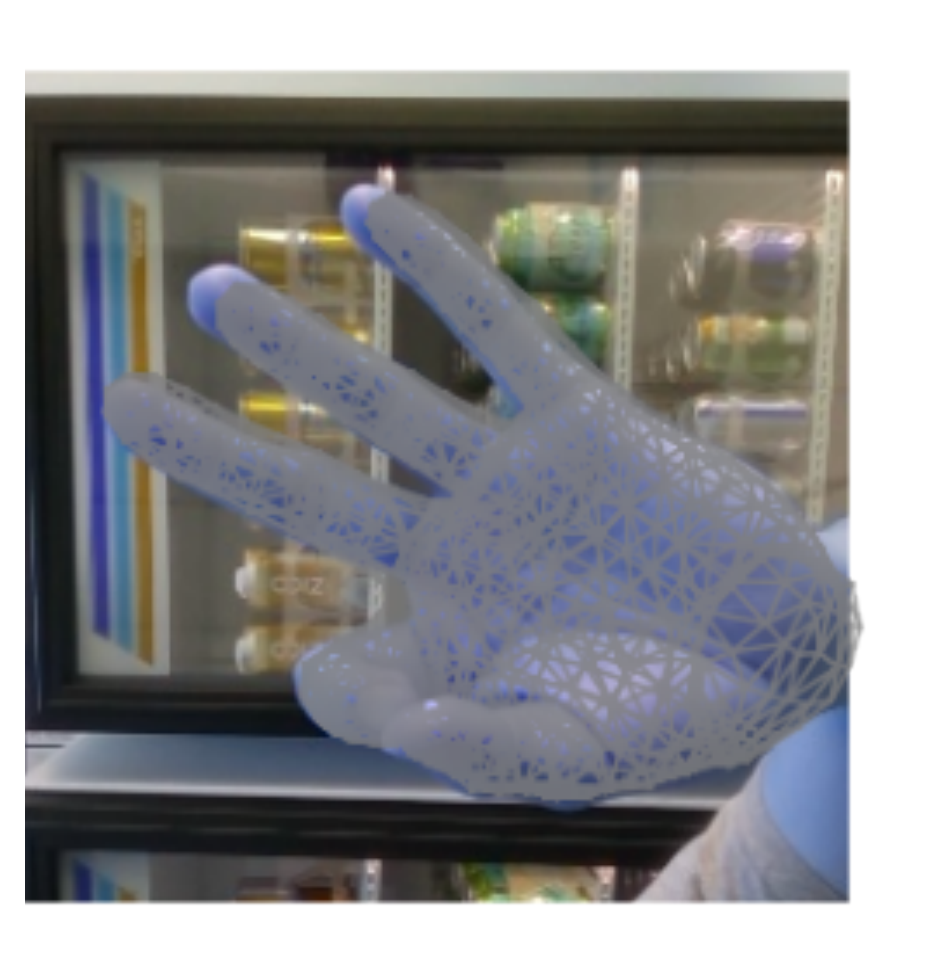}
    \centering
  \end{subfigure}
  \begin{subfigure}{\rotwidth\linewidth}
    \includegraphics[width=\linewidth]{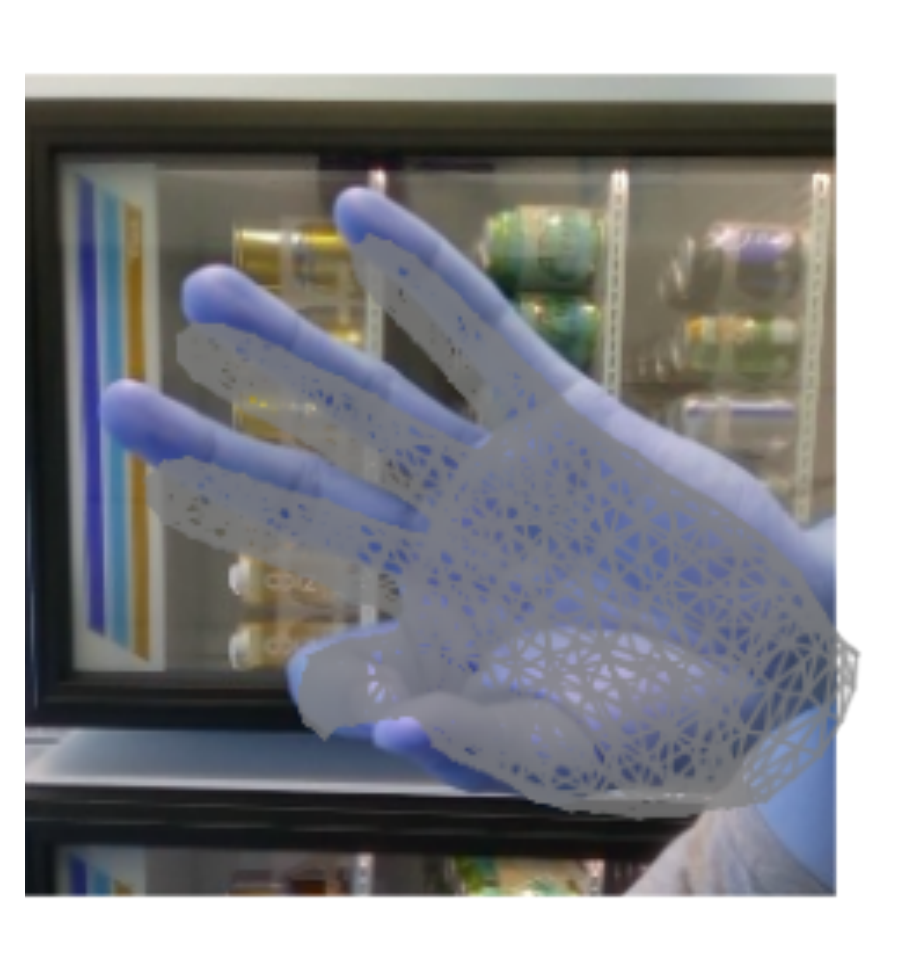}
    \centering
  \end{subfigure}
  \begin{subfigure}{\rotwidth\linewidth}
    \includegraphics[width=\linewidth]{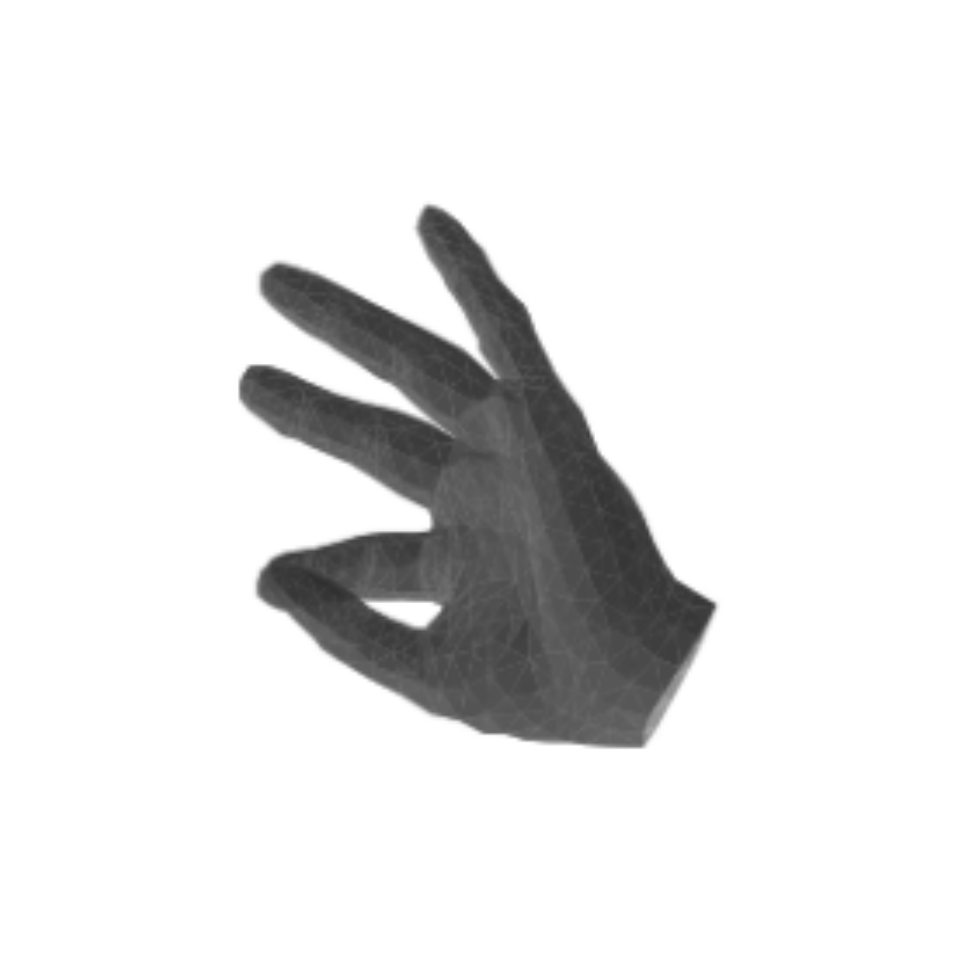}
    \centering
  \end{subfigure}
  \begin{subfigure}{\rotwidth\linewidth}
    \includegraphics[width=\linewidth]{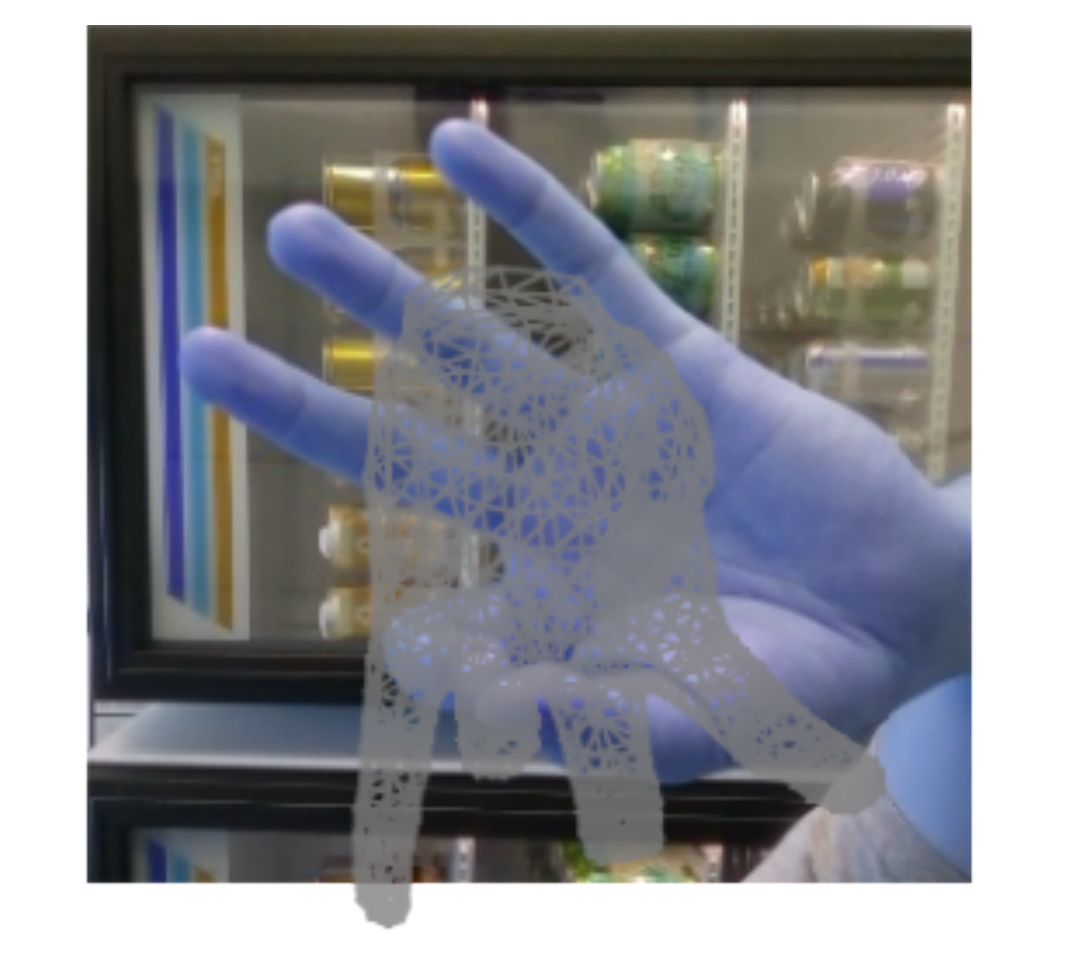}
    \centering
  \end{subfigure}
  \begin{subfigure}{\rotwidth\linewidth}
    \includegraphics[width=\linewidth]{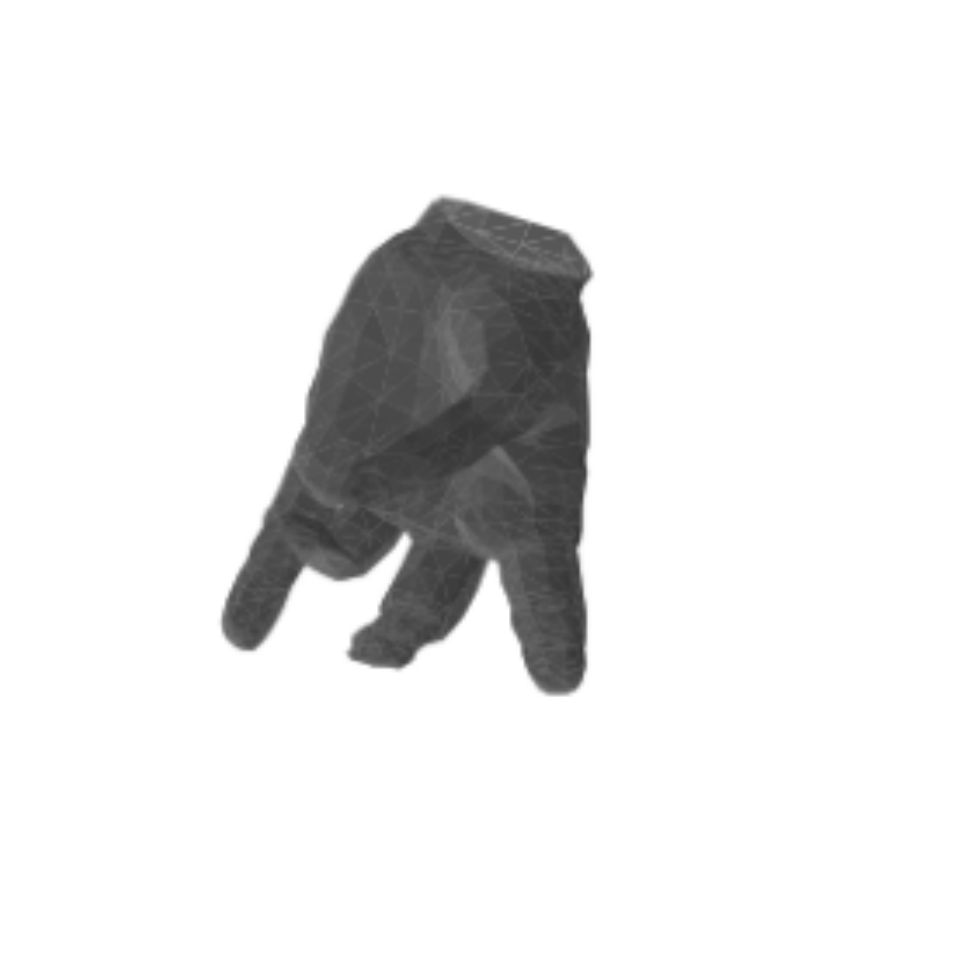}
    \centering
  \end{subfigure}
  \begin{subfigure}{\rotwidth\linewidth}
    \includegraphics[width=\linewidth]{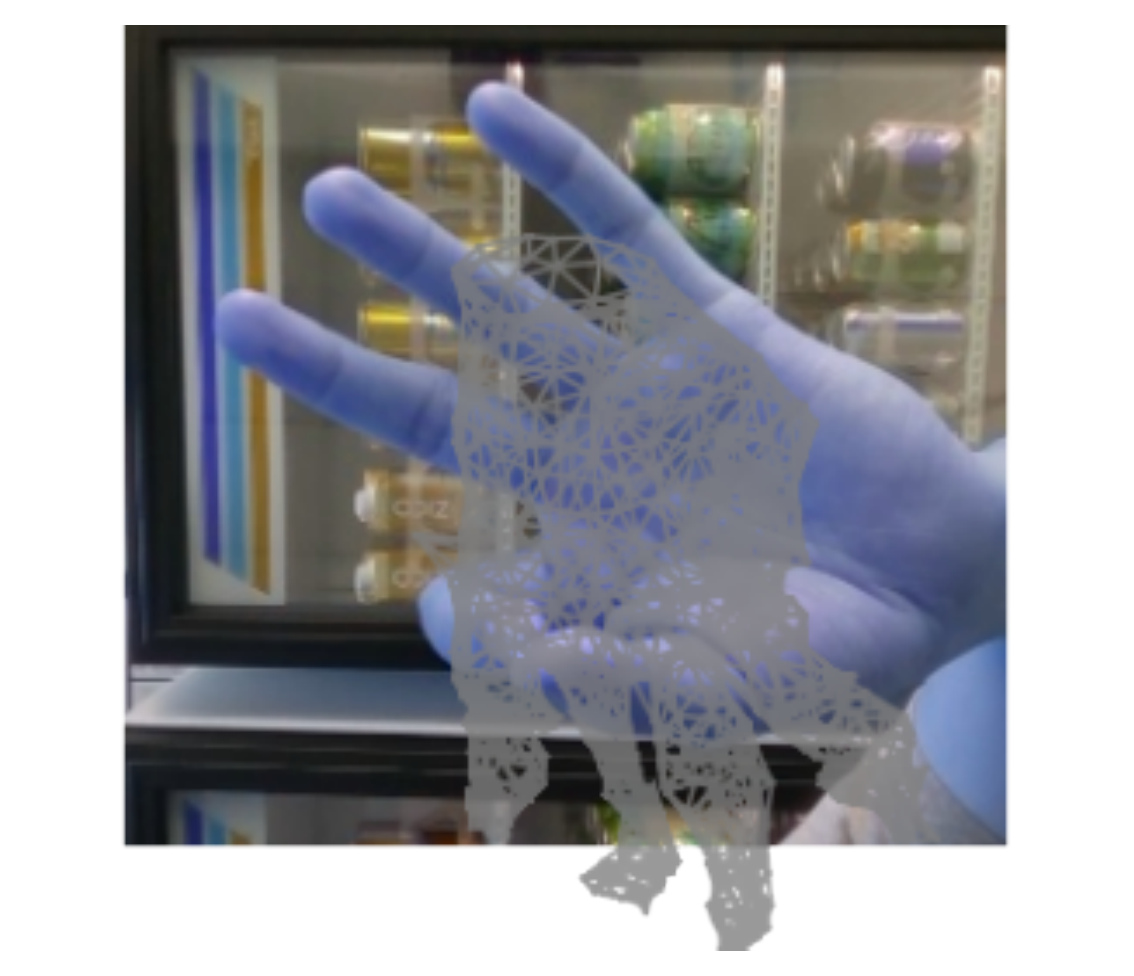}
    \centering
  \end{subfigure}
  \begin{subfigure}{\rotwidth\linewidth}
    \includegraphics[width=\linewidth]{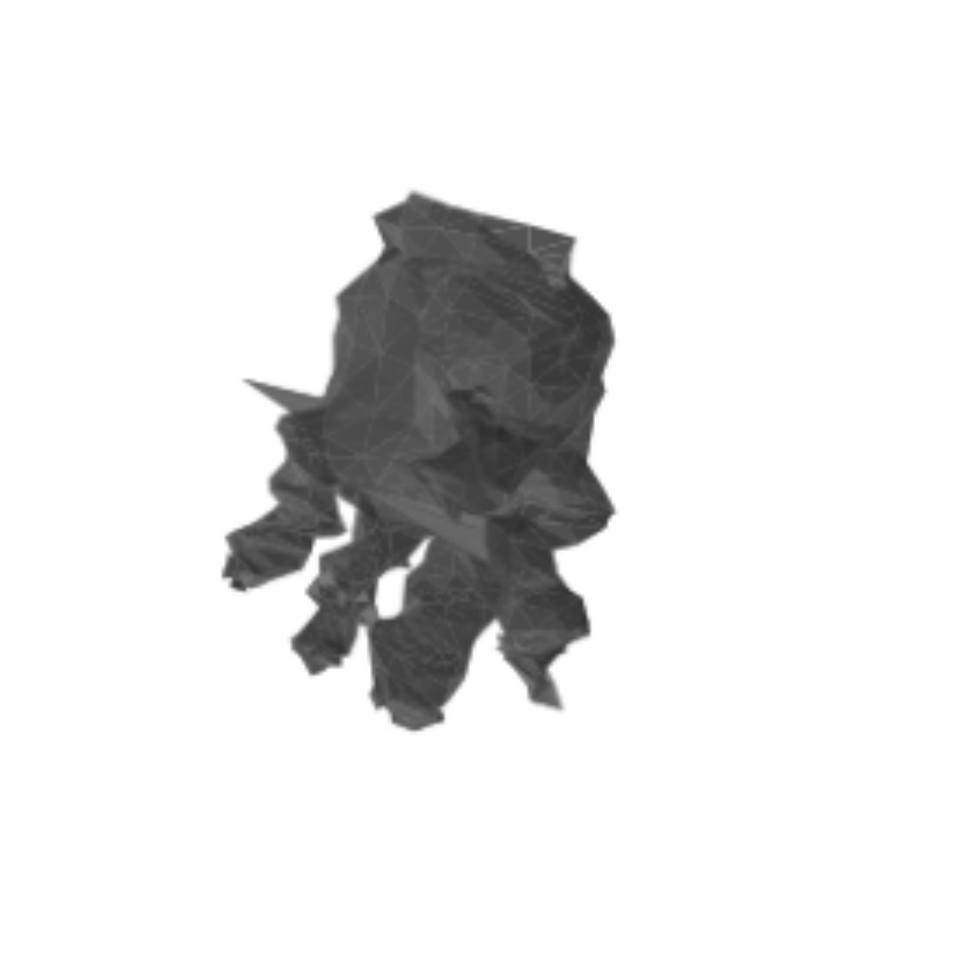}
    \centering
  \end{subfigure}

  \begin{subfigure}{\rotwidth\linewidth}
    \includegraphics[width=\linewidth]{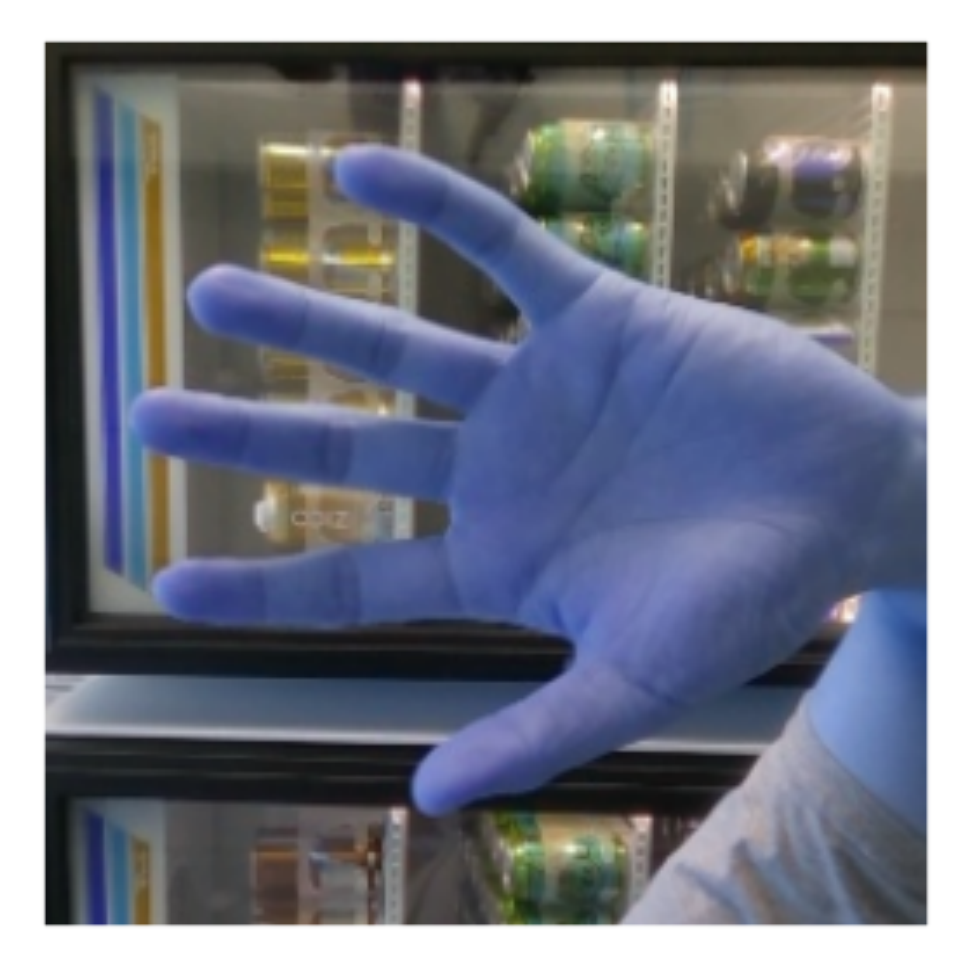}
    \centering
  \end{subfigure}
  \begin{subfigure}{\rotwidth\linewidth}
    \includegraphics[width=\linewidth]{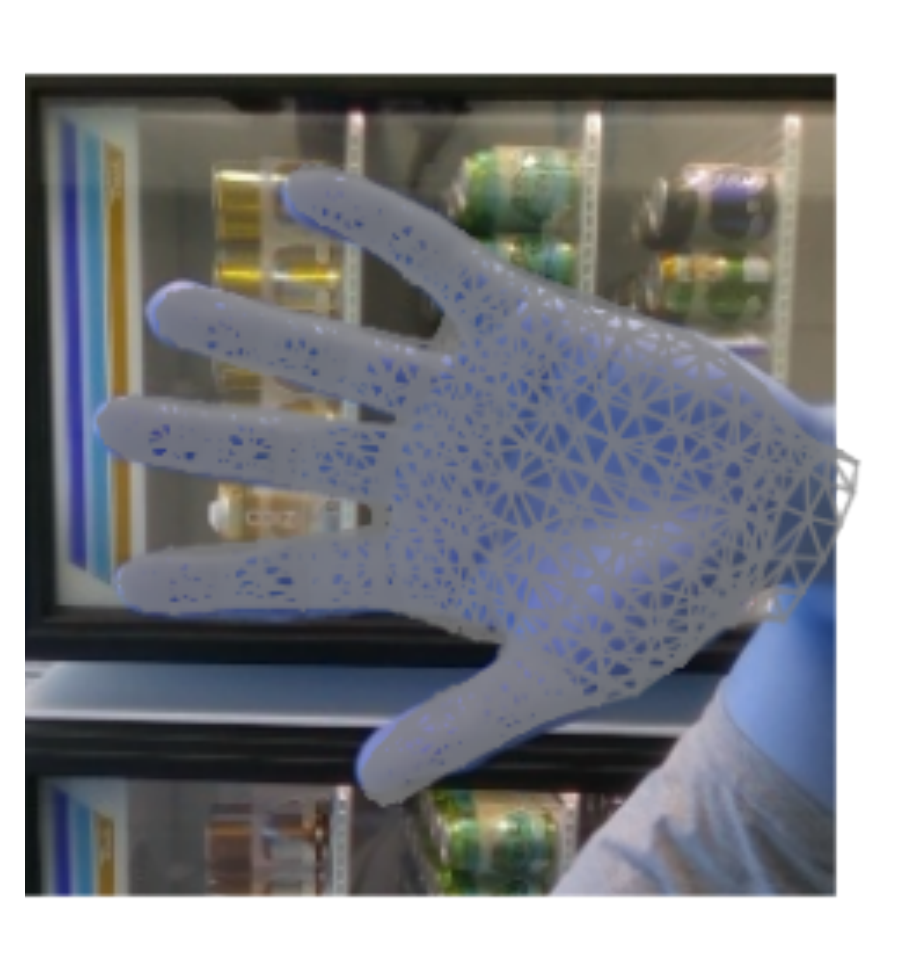}
    \centering
  \end{subfigure}
  \begin{subfigure}{\rotwidth\linewidth}
    \includegraphics[width=\linewidth]{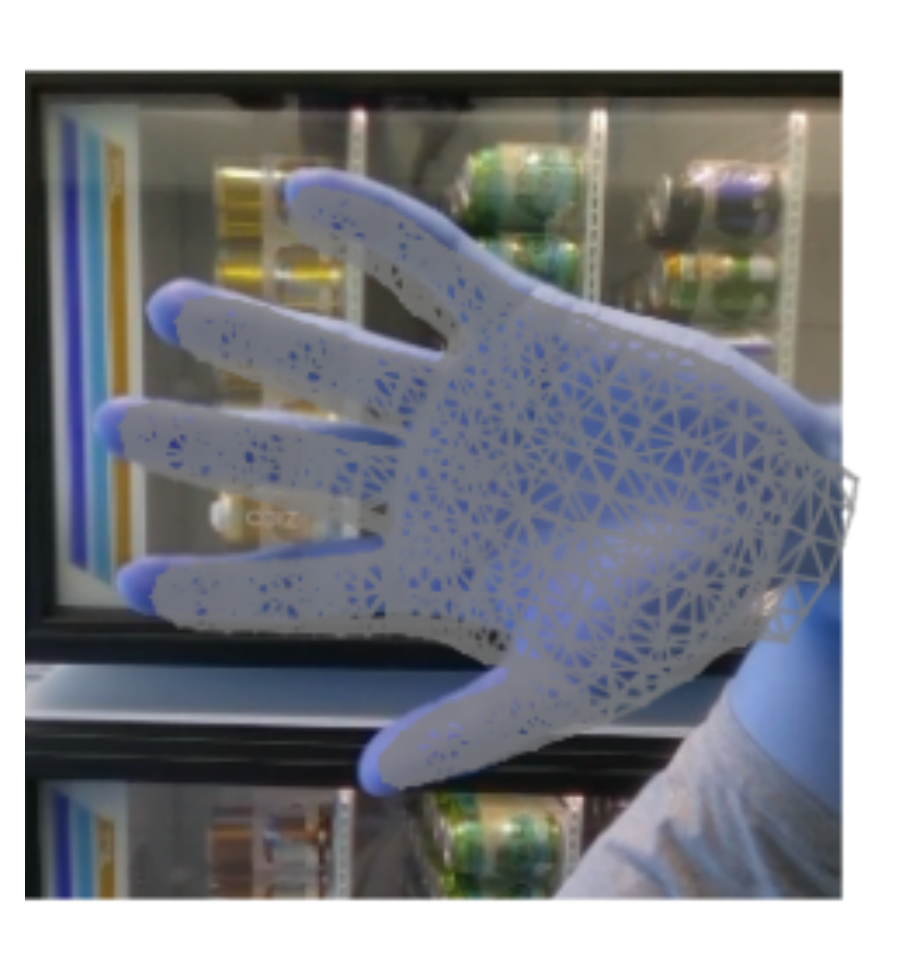}
    \centering
  \end{subfigure}
  \begin{subfigure}{\rotwidth\linewidth}
    \includegraphics[width=\linewidth]{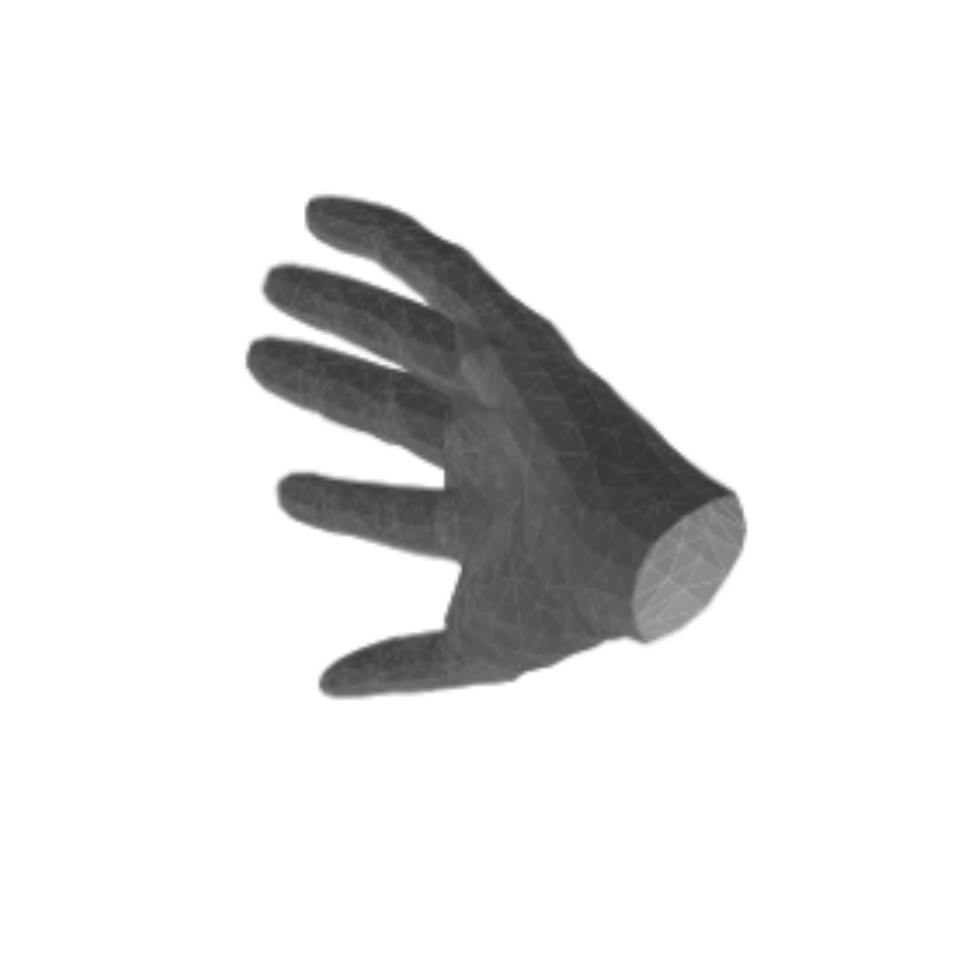}
    \centering
  \end{subfigure}
  \begin{subfigure}{\rotwidth\linewidth}
    \includegraphics[width=\linewidth]{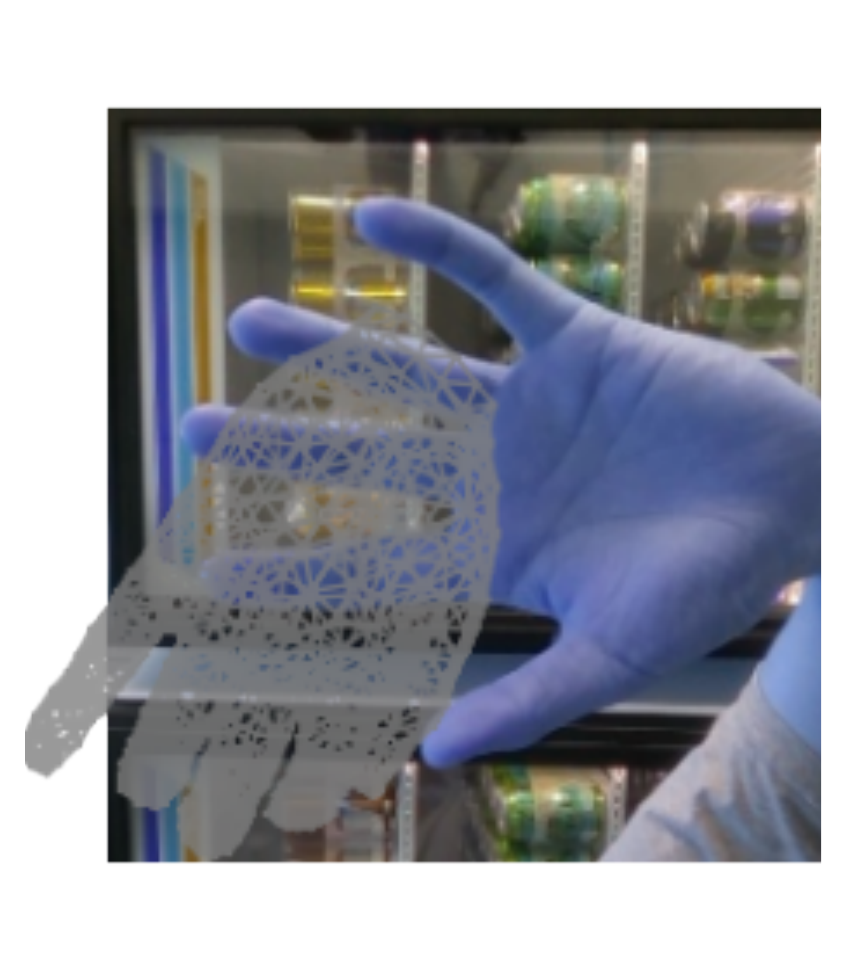}
    \centering
  \end{subfigure}
  \begin{subfigure}{\rotwidth\linewidth}
    \includegraphics[width=\linewidth]{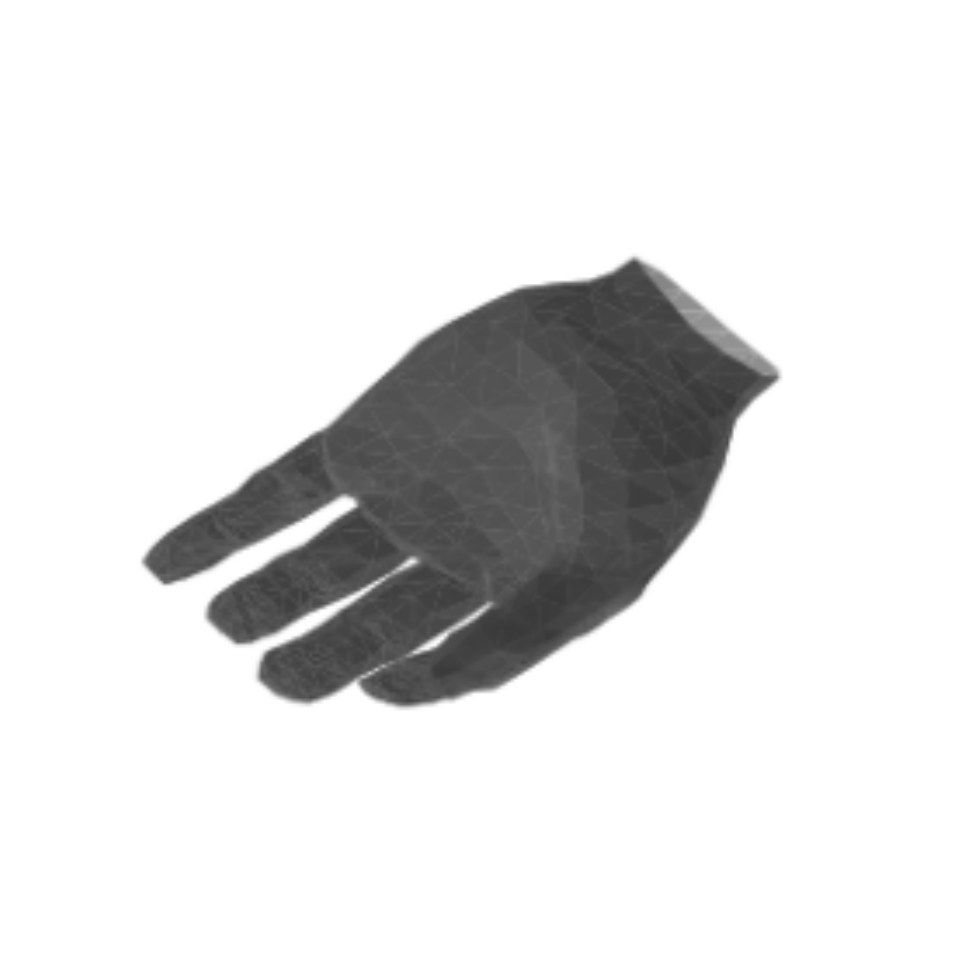}
    \centering
  \end{subfigure}
  \begin{subfigure}{\rotwidth\linewidth}
    \includegraphics[width=\linewidth]{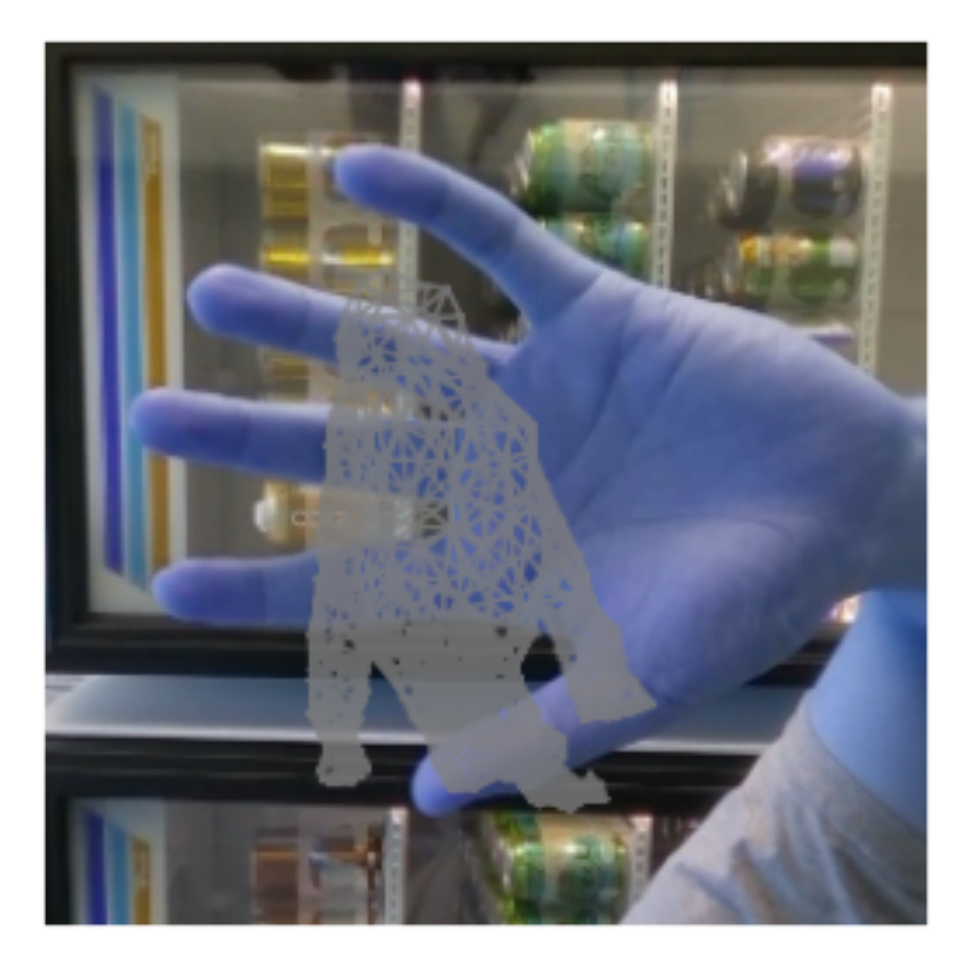}
    \centering
  \end{subfigure}
  \begin{subfigure}{\rotwidth\linewidth}
    \includegraphics[width=\linewidth]{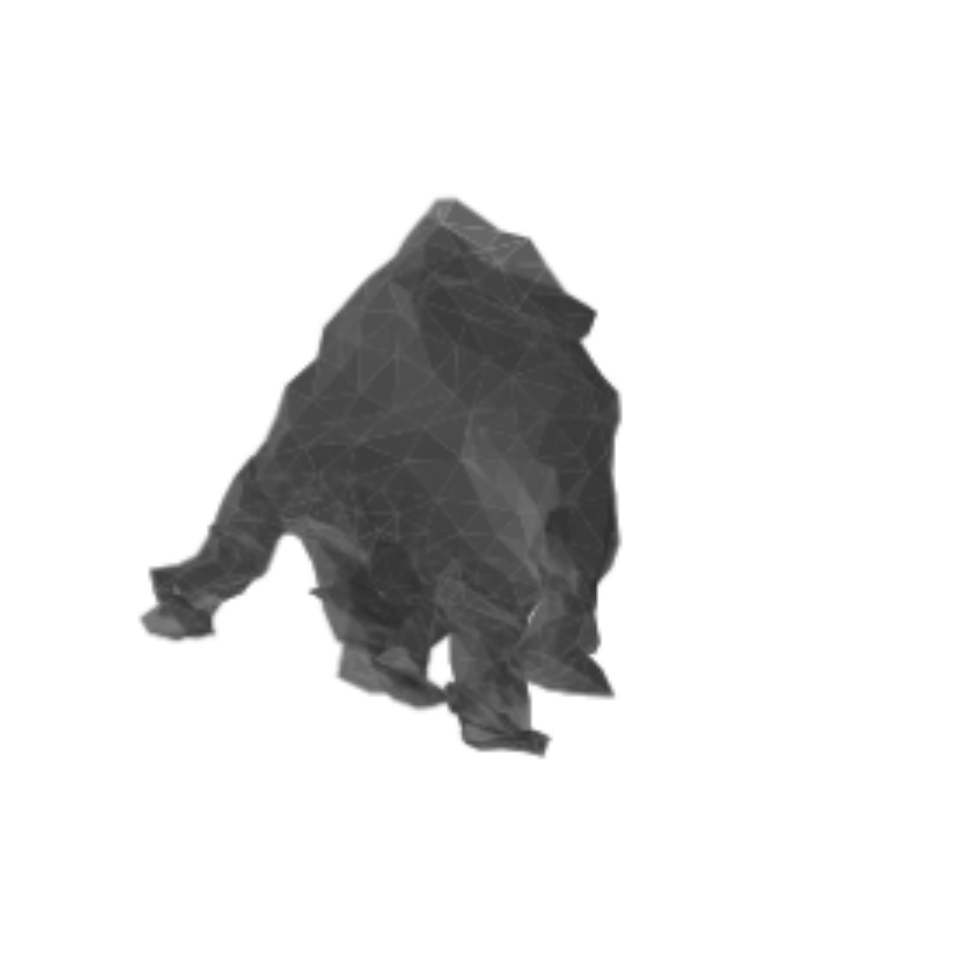}
    \centering
  \end{subfigure}

  \begin{subfigure}{\rotwidth\linewidth}
    \includegraphics[width=\linewidth]{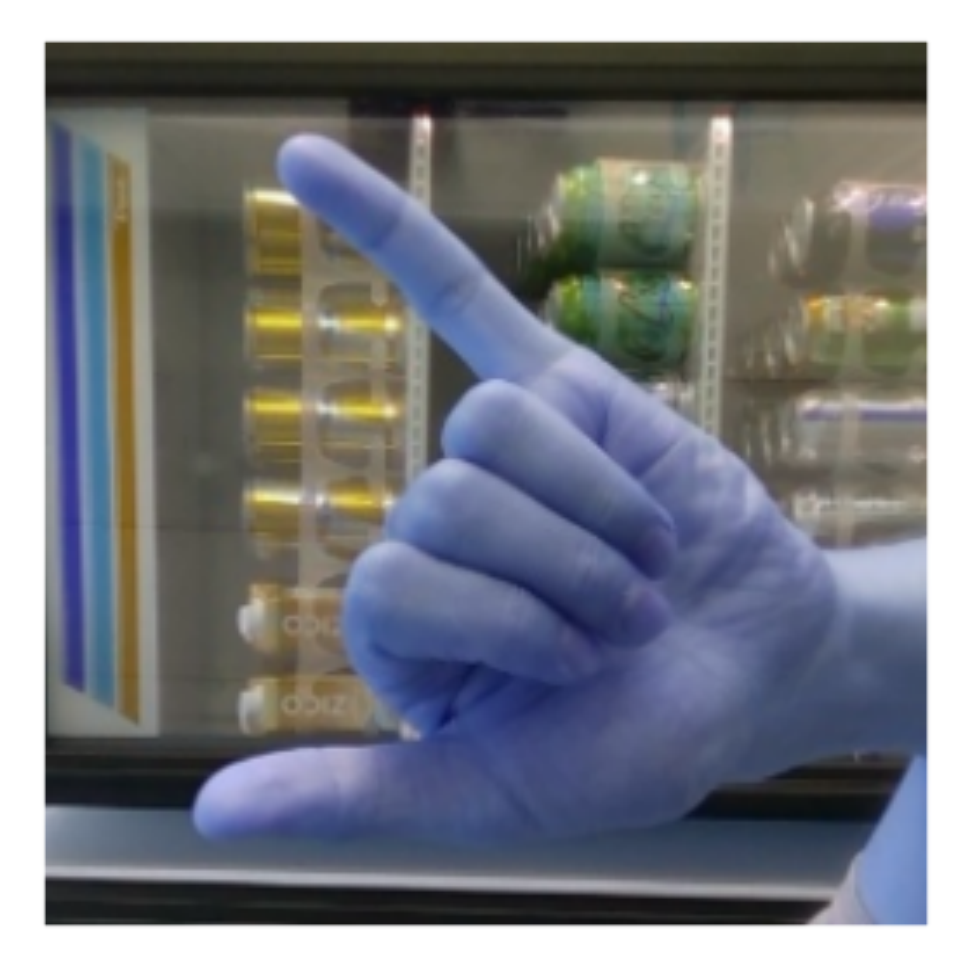}
    \centering
  \end{subfigure}
  \begin{subfigure}{\rotwidth\linewidth}
    \includegraphics[width=\linewidth]{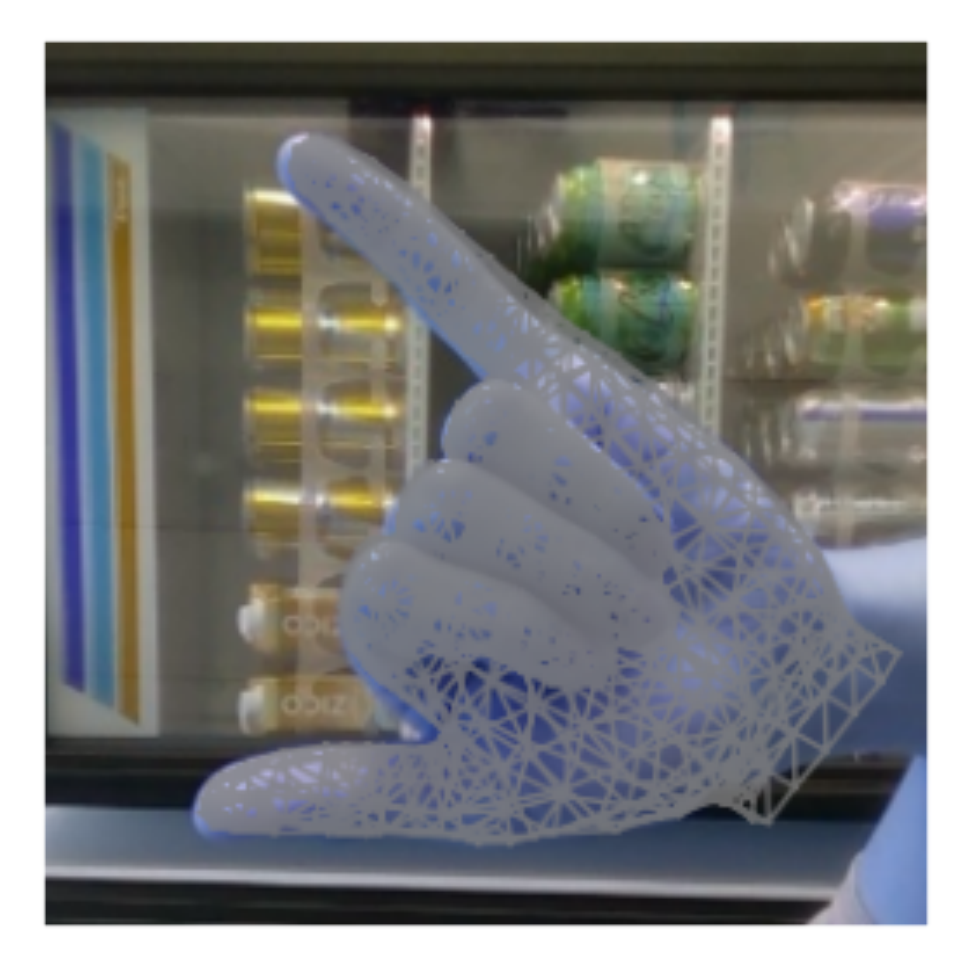}
    \centering
  \end{subfigure}
  \begin{subfigure}{\rotwidth\linewidth}
    \includegraphics[width=\linewidth]{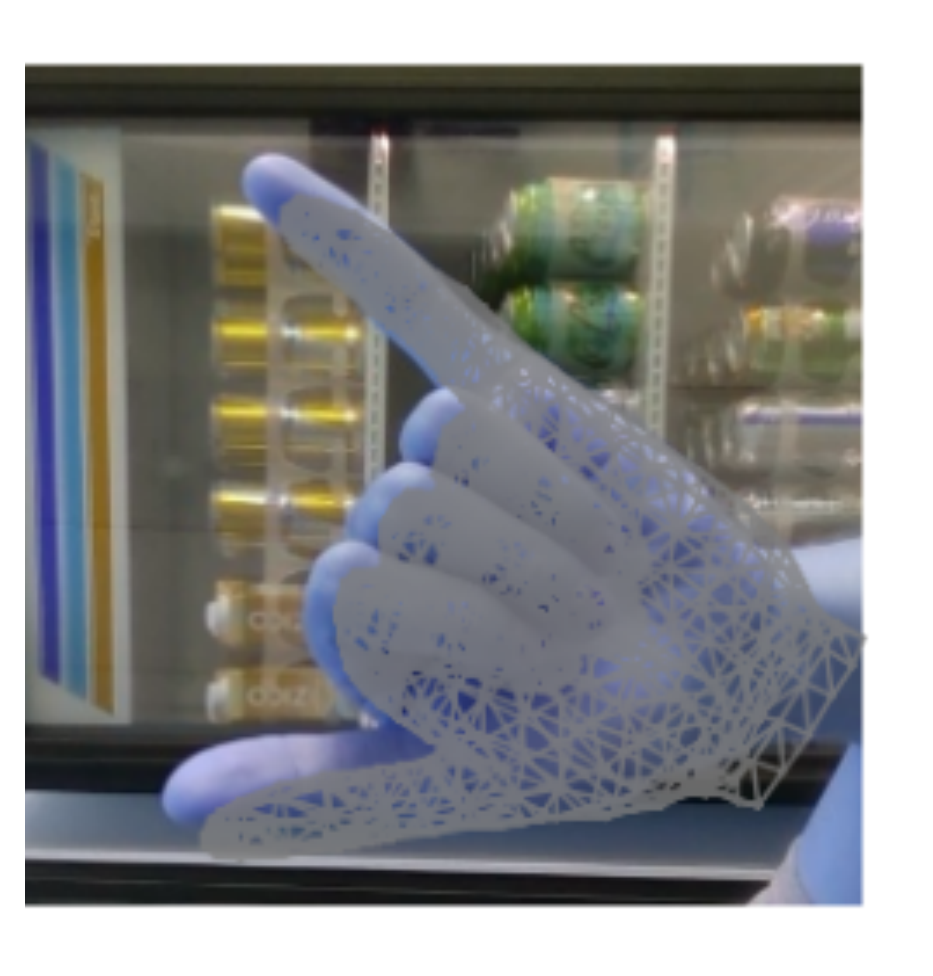}
    \centering
  \end{subfigure}
  \begin{subfigure}{\rotwidth\linewidth}
    \includegraphics[width=\linewidth]{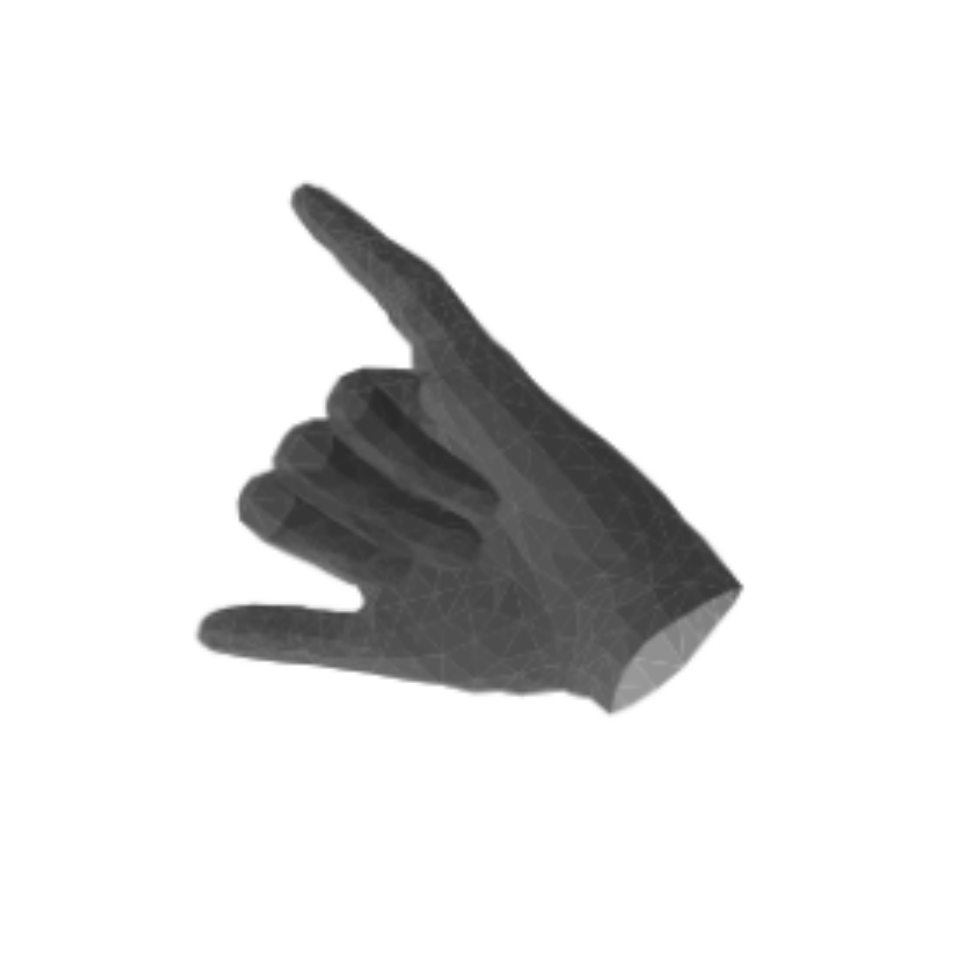}
    \centering
  \end{subfigure}
  \begin{subfigure}{\rotwidth\linewidth}
    \includegraphics[width=\linewidth]{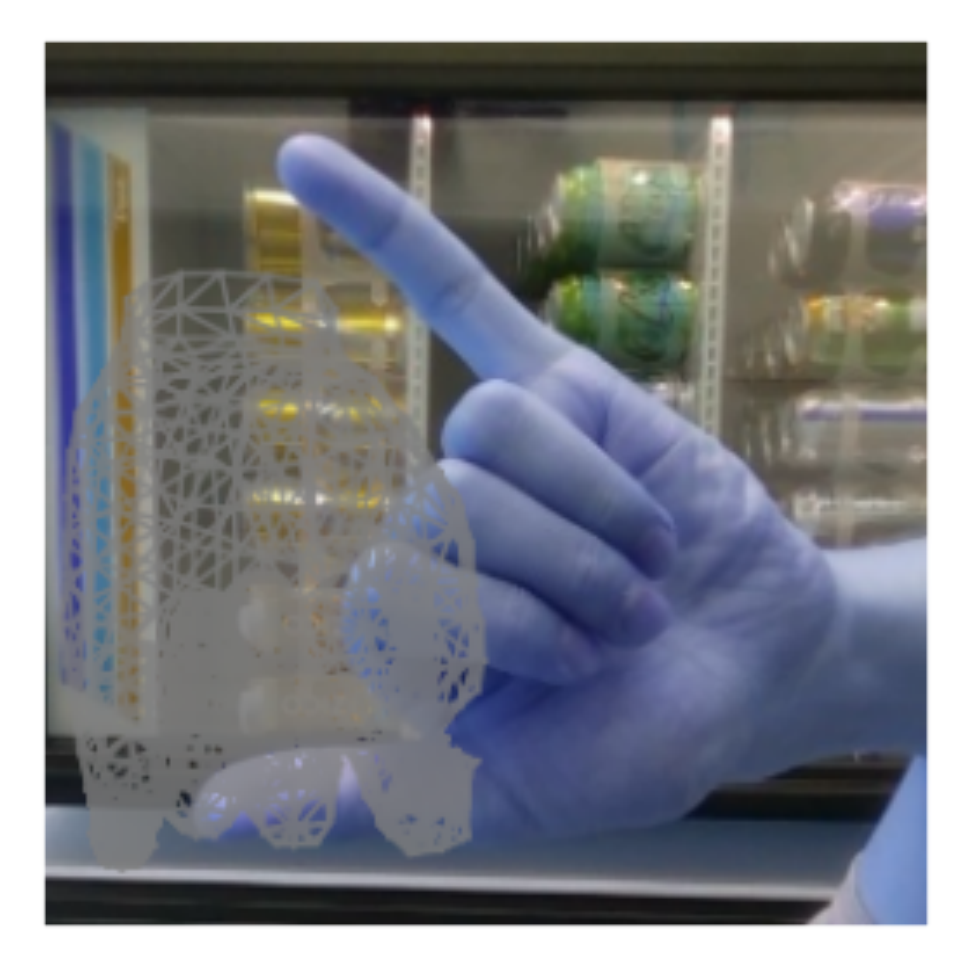}
    \centering
  \end{subfigure}
  \begin{subfigure}{\rotwidth\linewidth}
    \includegraphics[width=\linewidth]{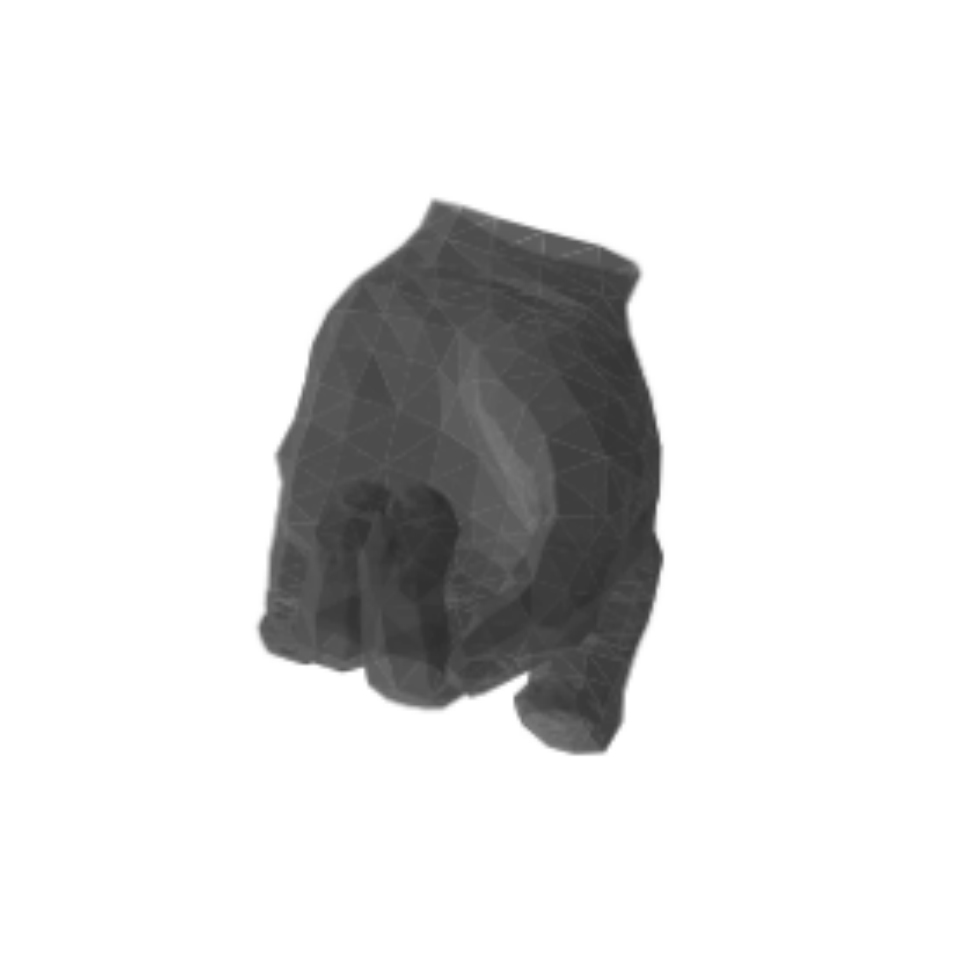}
    \centering
  \end{subfigure}
  \begin{subfigure}{\rotwidth\linewidth}
    \includegraphics[width=\linewidth]{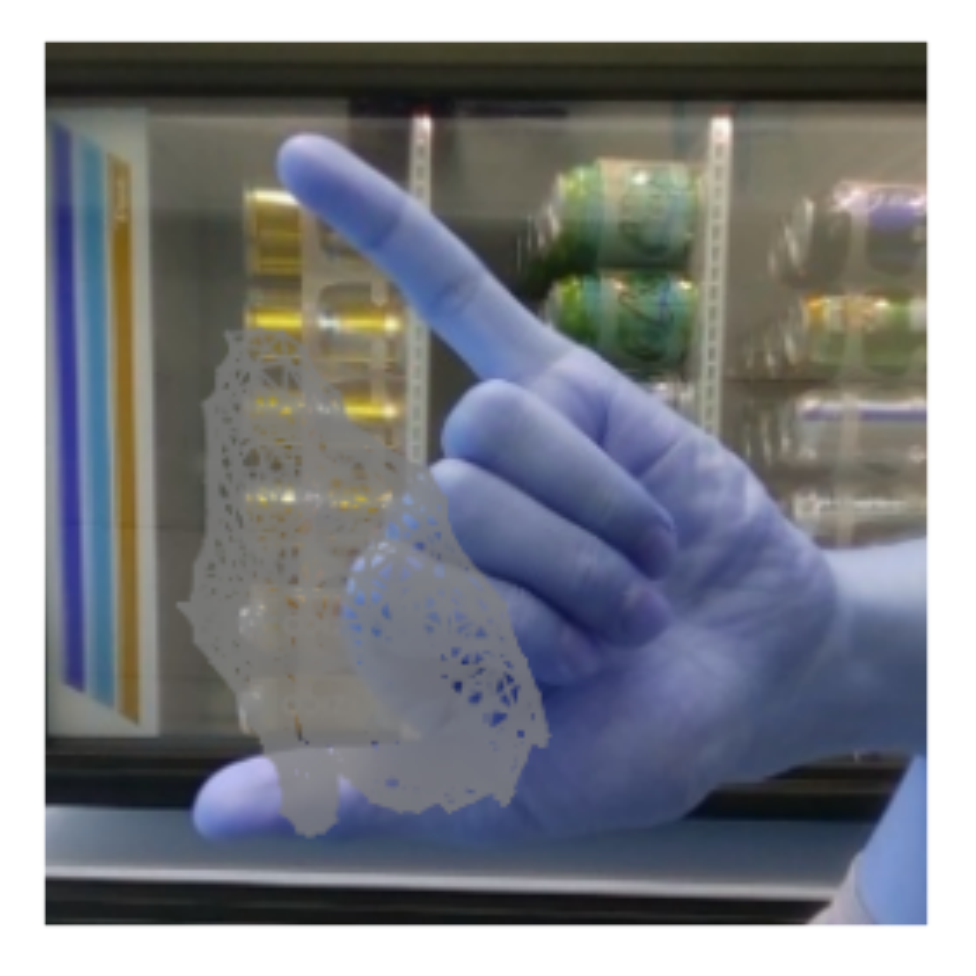}
    \centering
  \end{subfigure}
  \begin{subfigure}{\rotwidth\linewidth}
    \includegraphics[width=\linewidth]{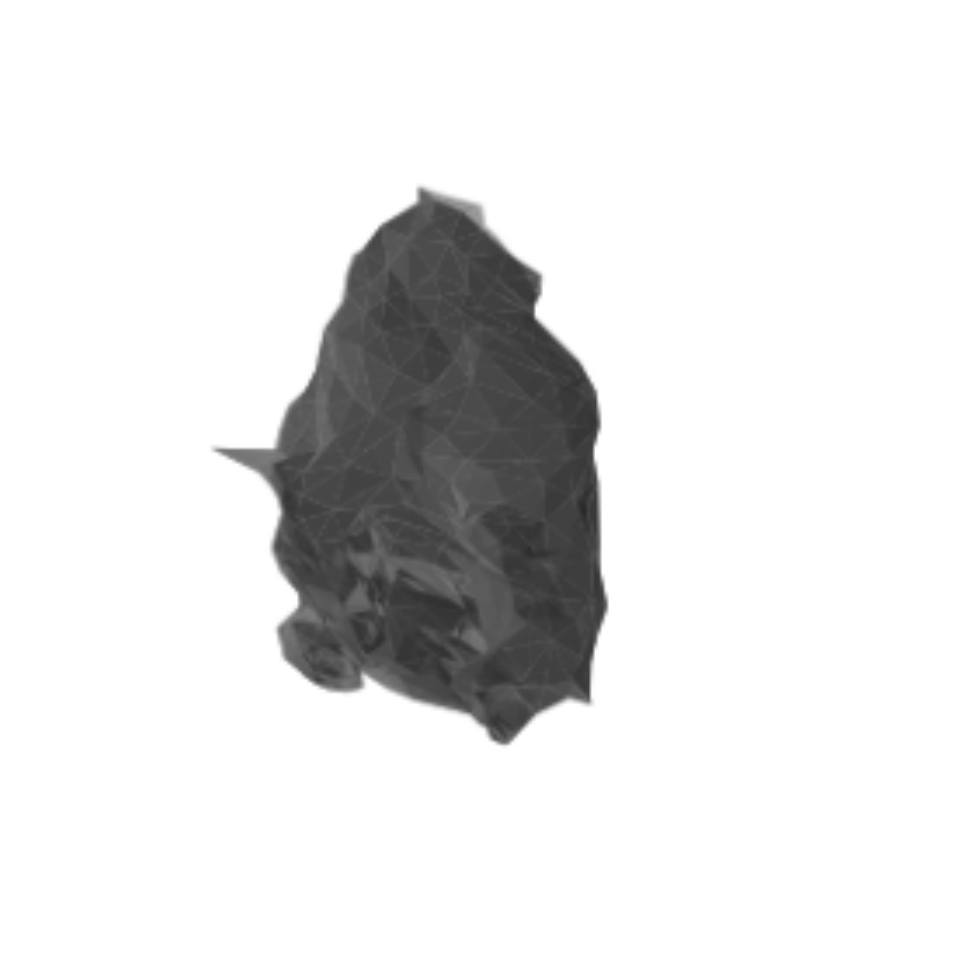}
    \centering
  \end{subfigure}

  \begin{subfigure}{\rotwidth\linewidth}
    \includegraphics[width=\linewidth]{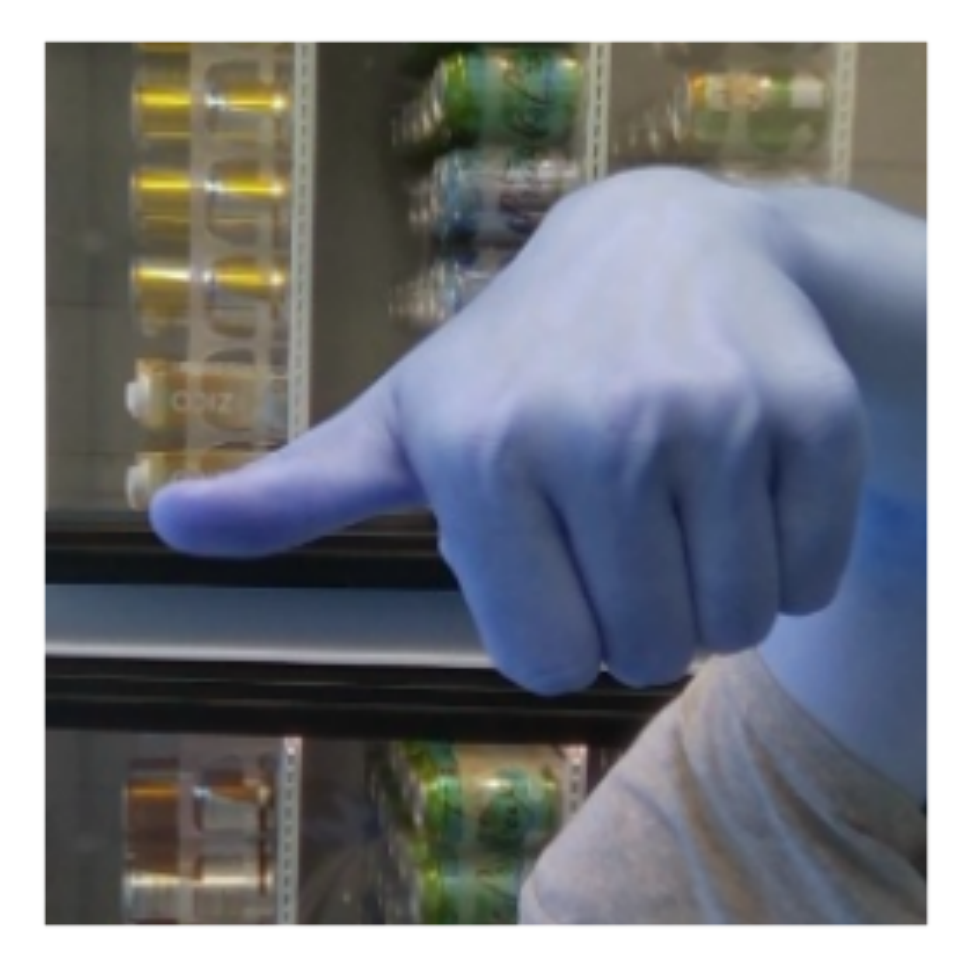}
    \centering
  \end{subfigure}
  \begin{subfigure}{\rotwidth\linewidth}
    \includegraphics[width=\linewidth]{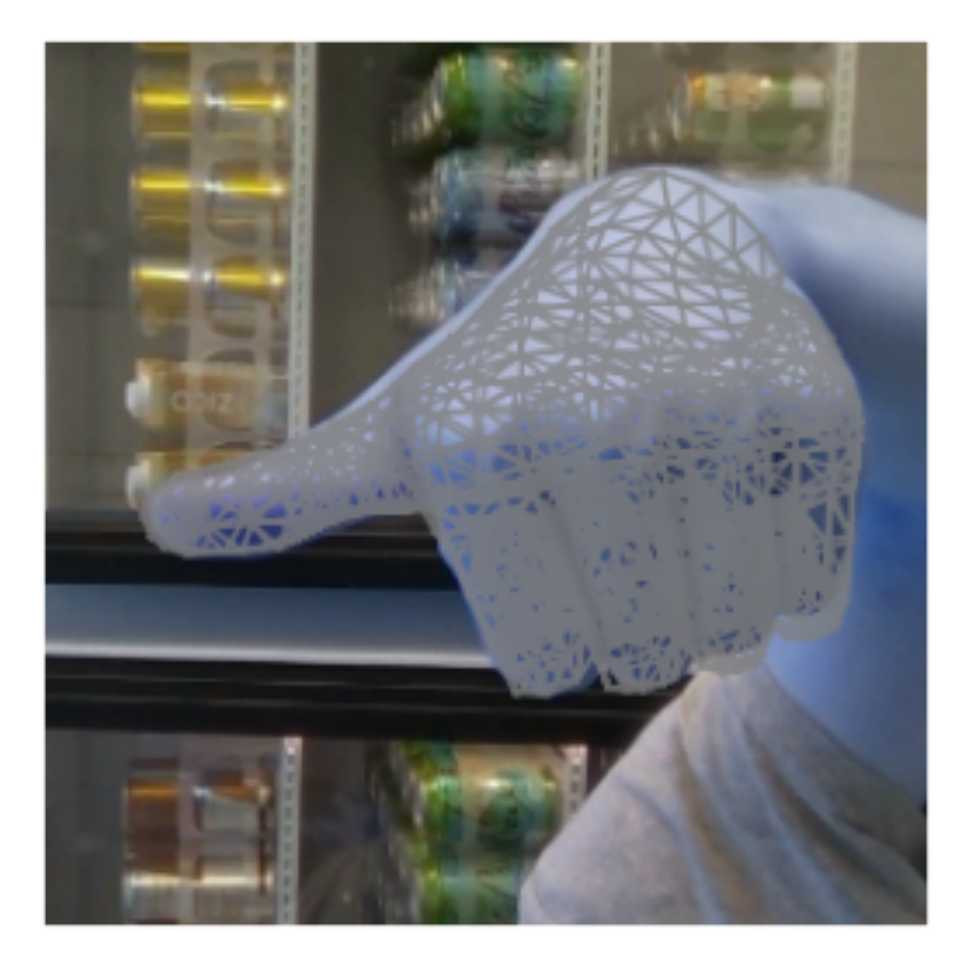}
    \centering
  \end{subfigure}
  \begin{subfigure}{\rotwidth\linewidth}
    \includegraphics[width=\linewidth]{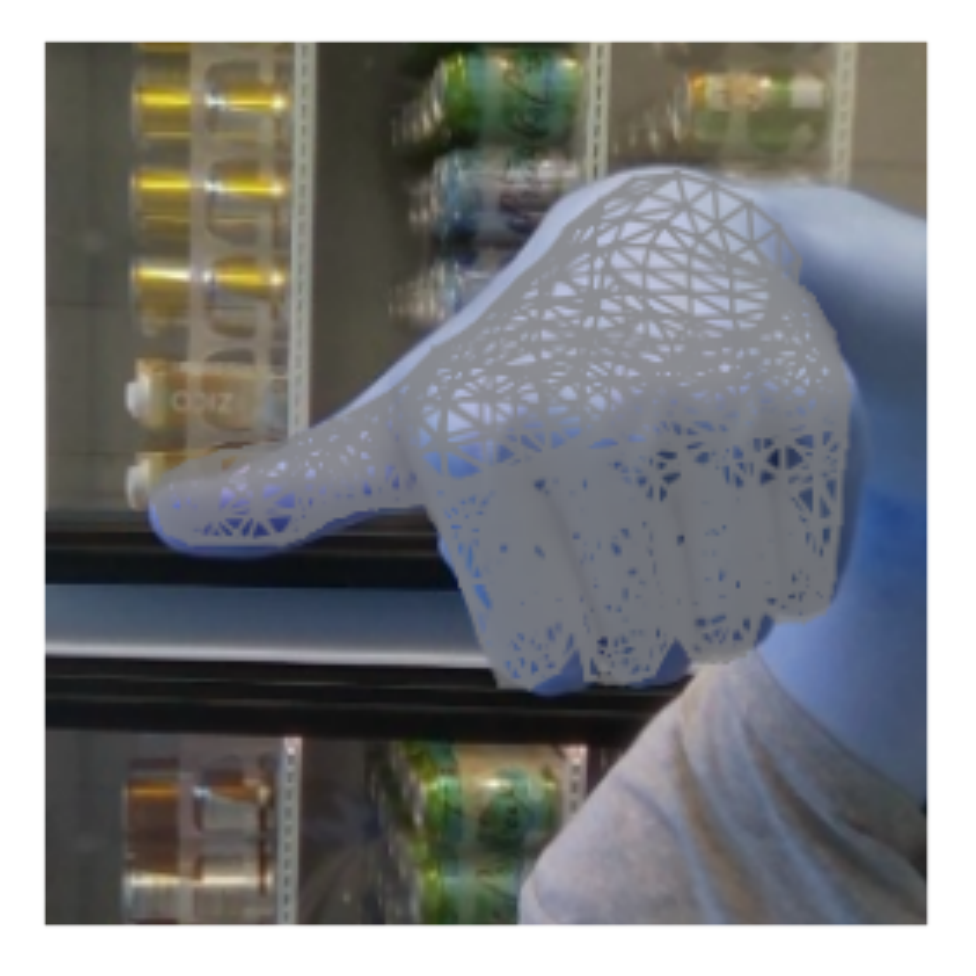}
    \centering
  \end{subfigure}
  \begin{subfigure}{\rotwidth\linewidth}
    \includegraphics[width=\linewidth]{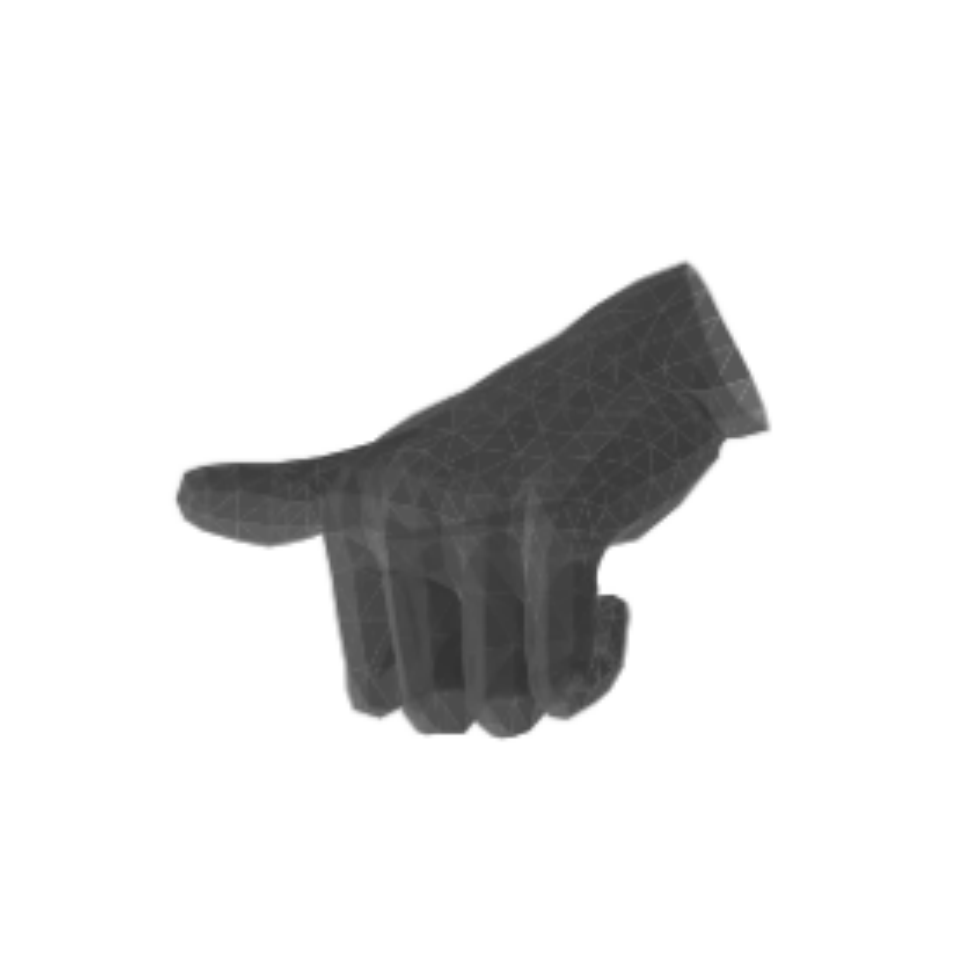}
    \centering
  \end{subfigure}
  \begin{subfigure}{\rotwidth\linewidth}
    \includegraphics[width=\linewidth]{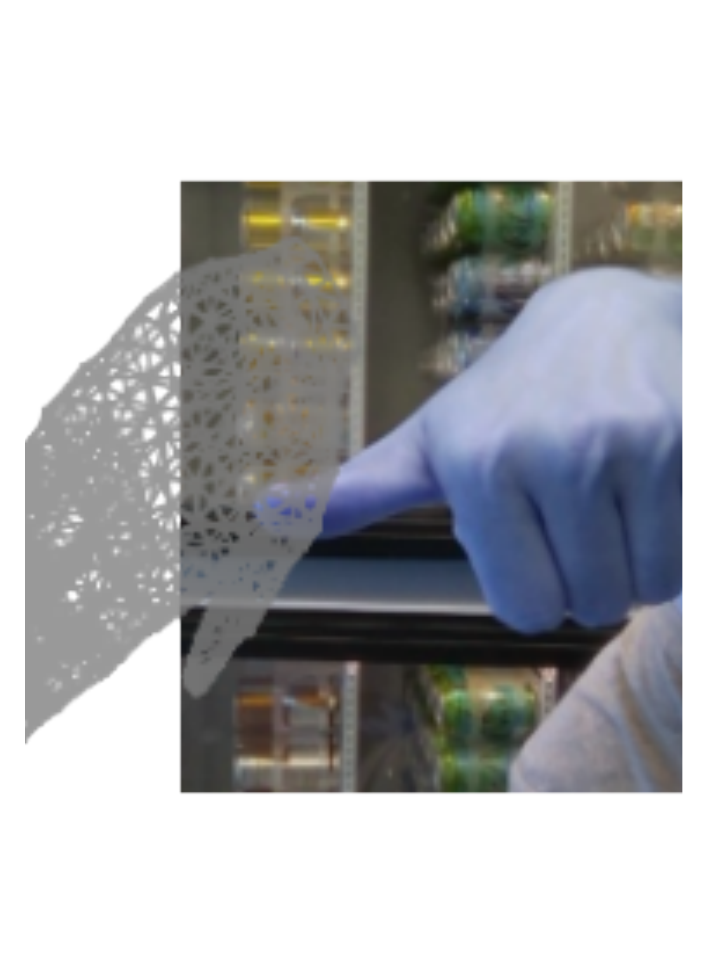}
    \centering
  \end{subfigure}
  \begin{subfigure}{\rotwidth\linewidth}
    \includegraphics[width=\linewidth]{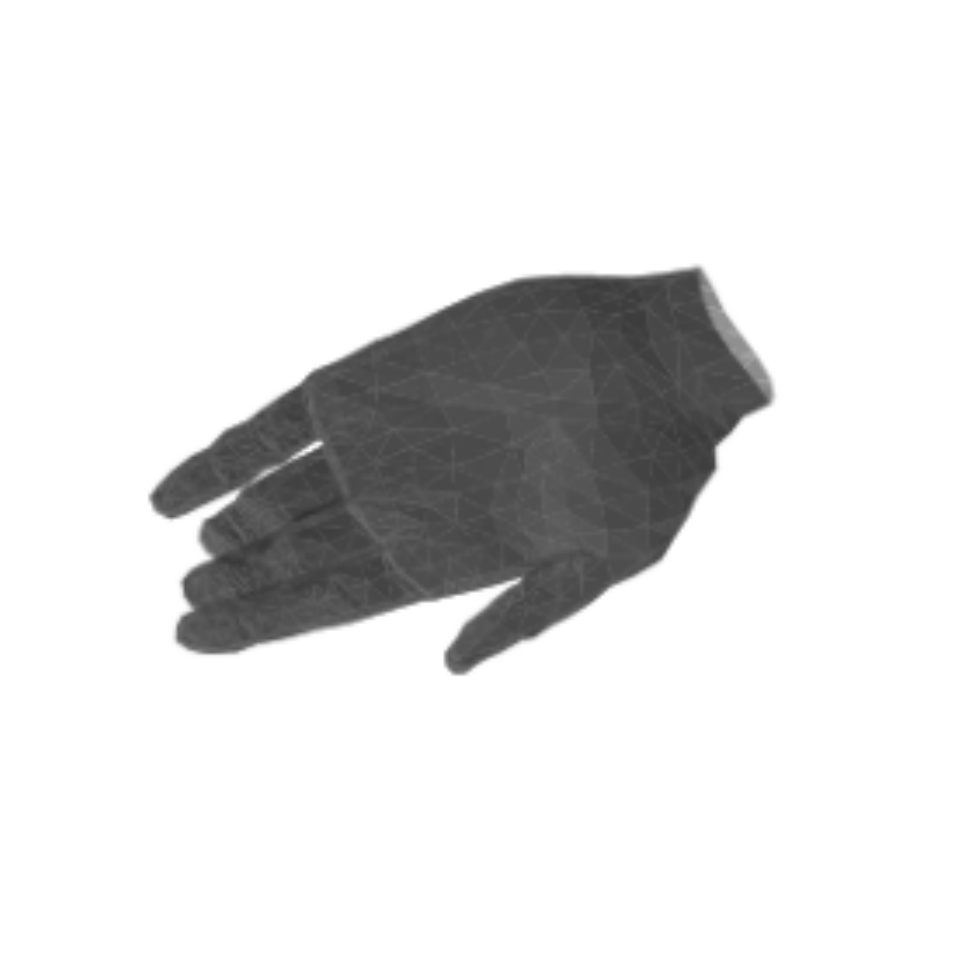}
    \centering
  \end{subfigure}
  \begin{subfigure}{\rotwidth\linewidth}
    \includegraphics[width=\linewidth]{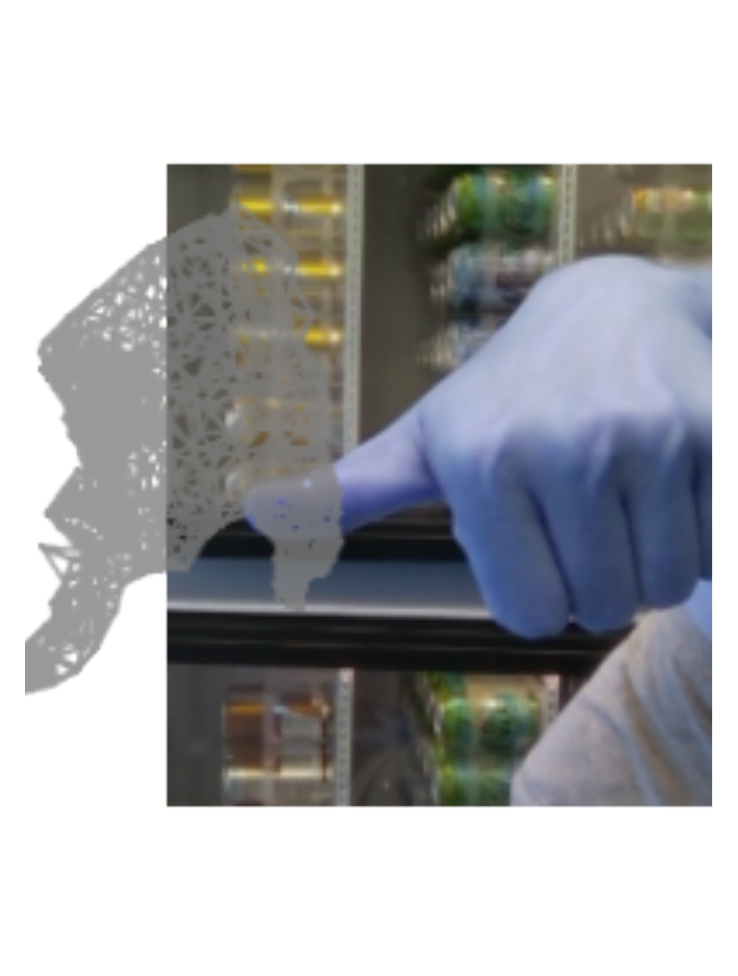}
    \centering
  \end{subfigure}
  \begin{subfigure}{\rotwidth\linewidth}
    \includegraphics[width=\linewidth]{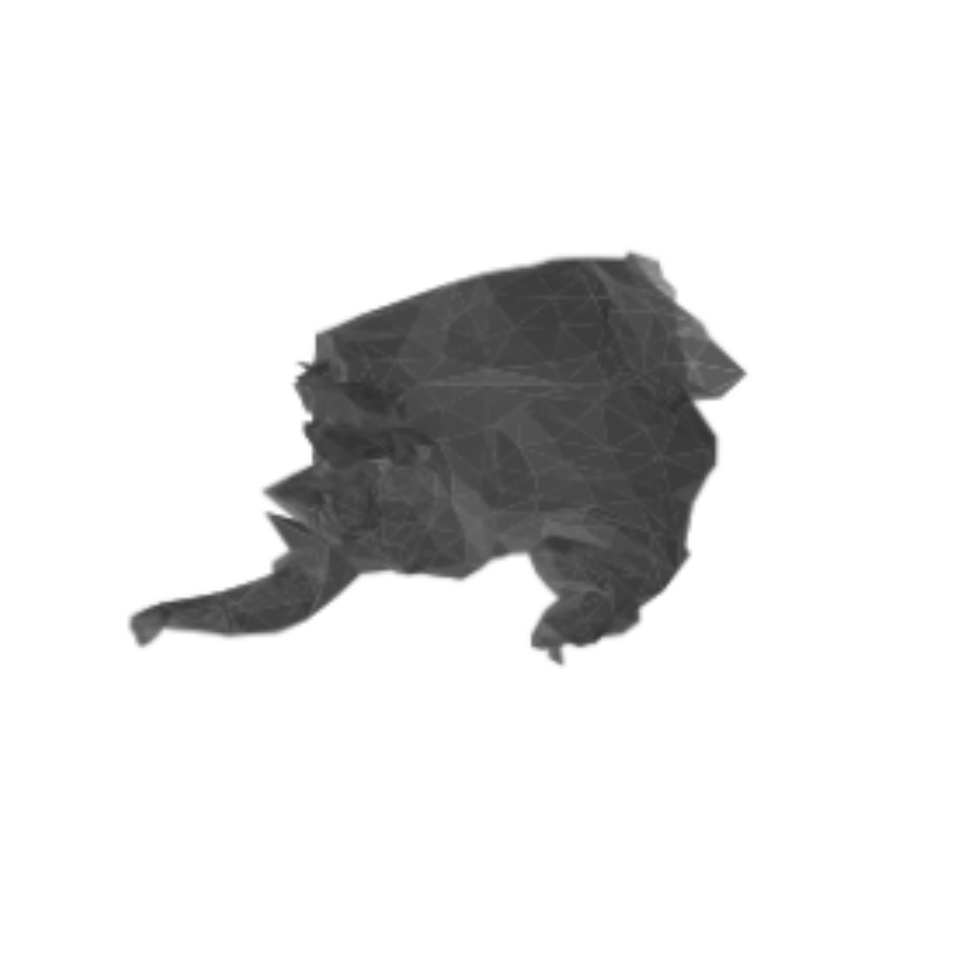}
    \centering
  \end{subfigure}

  \begin{subfigure}{\rotwidth\linewidth}
    \includegraphics[width=\linewidth]{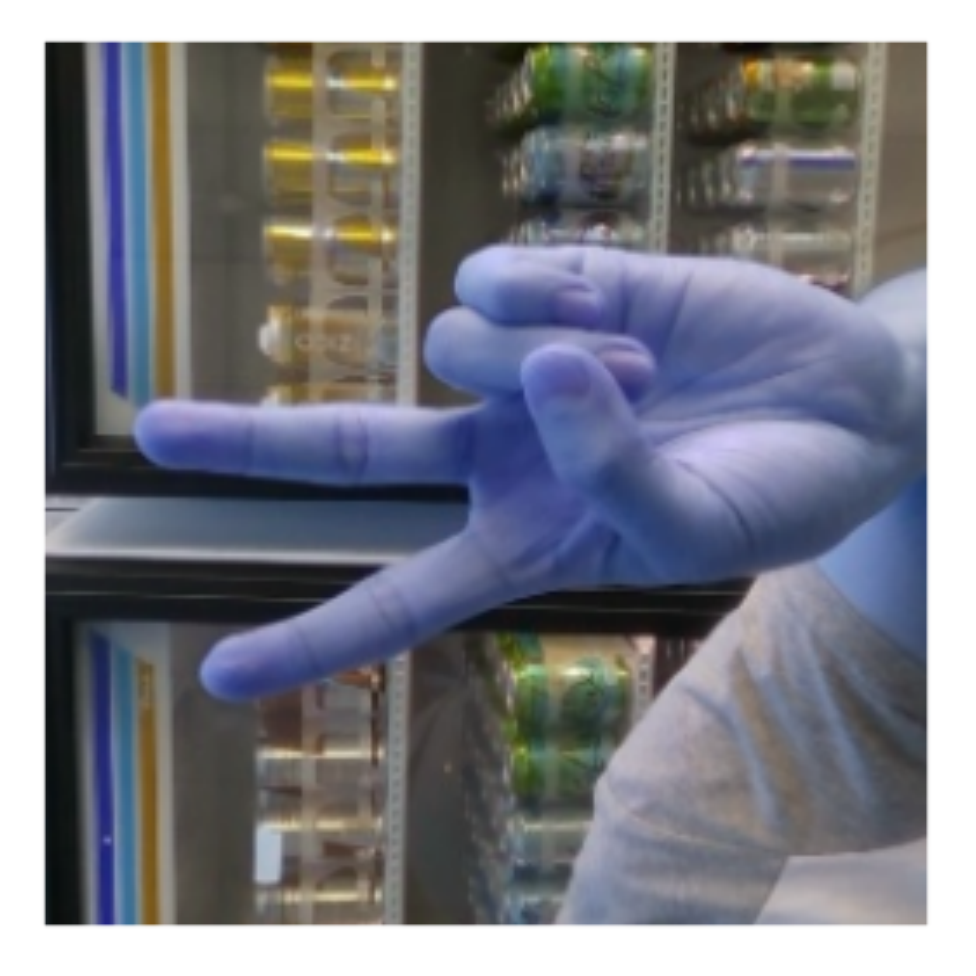}
    \centering
    \caption*{Input}
  \end{subfigure}
  \begin{subfigure}{\rotwidth\linewidth}
    \includegraphics[width=\linewidth]{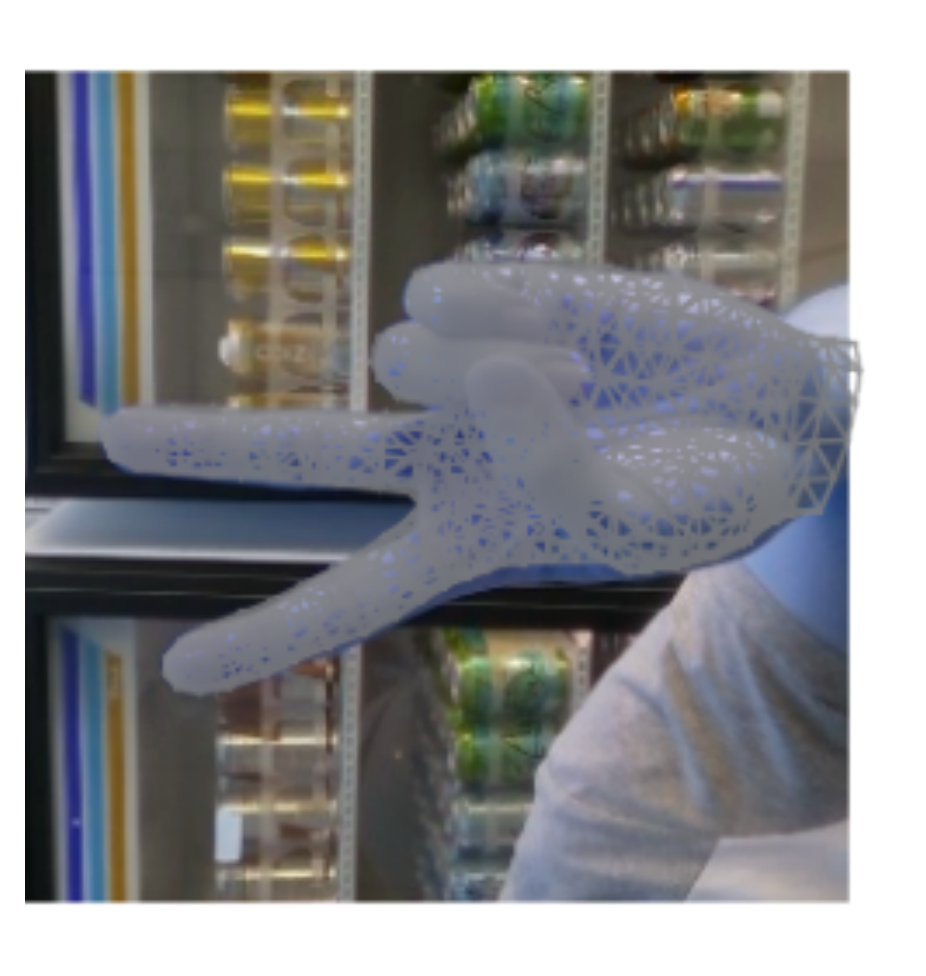}
    \centering
    \caption*{Target}
  \end{subfigure}
  \begin{subfigure}{\rotwidth\linewidth}
    \includegraphics[width=\linewidth]{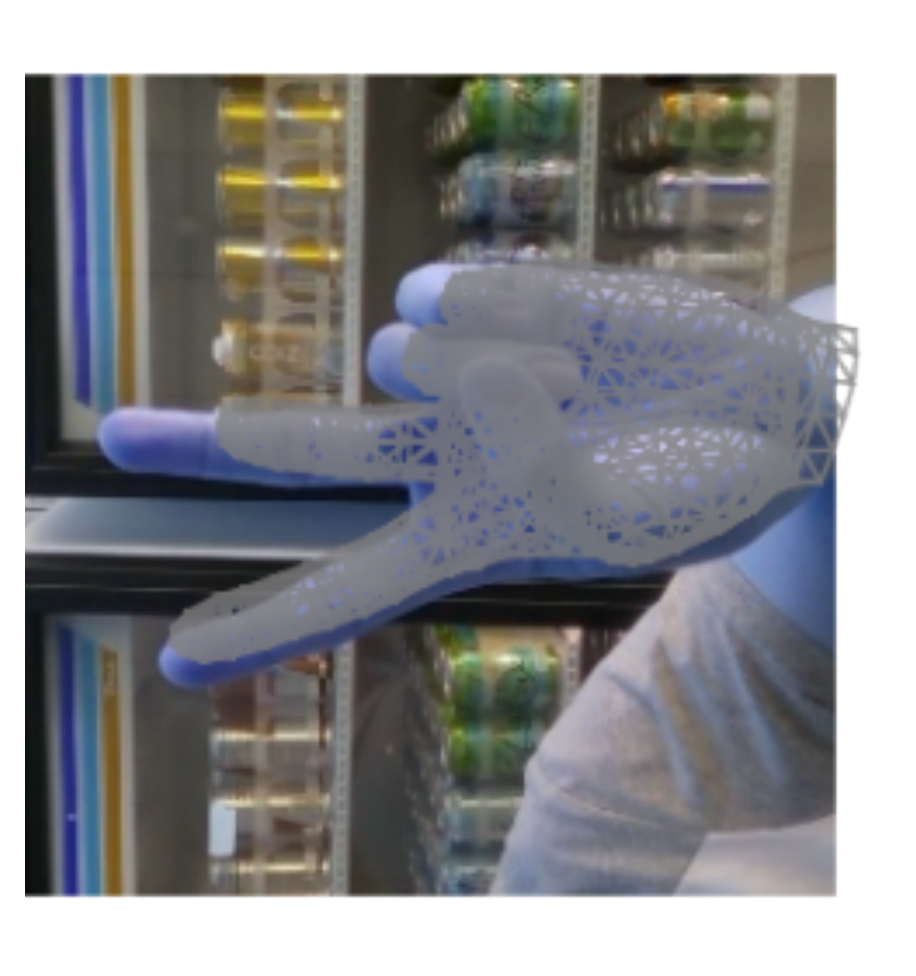}
    \centering
    \caption*{EMLP Prediction}
  \end{subfigure}
  \begin{subfigure}{\rotwidth\linewidth}
    \includegraphics[width=\linewidth]{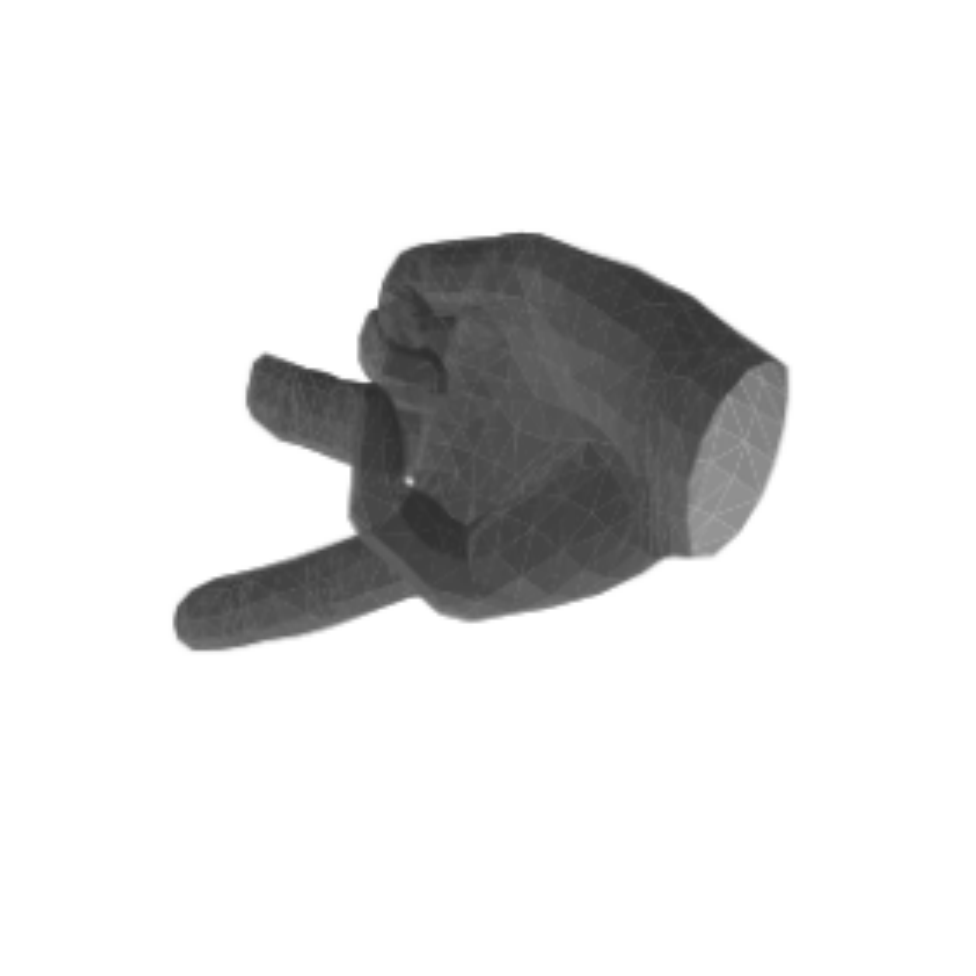}
    \centering
    \caption*{EMLP View A}
  \end{subfigure}
  \begin{subfigure}{\rotwidth\linewidth}
    \includegraphics[width=\linewidth]{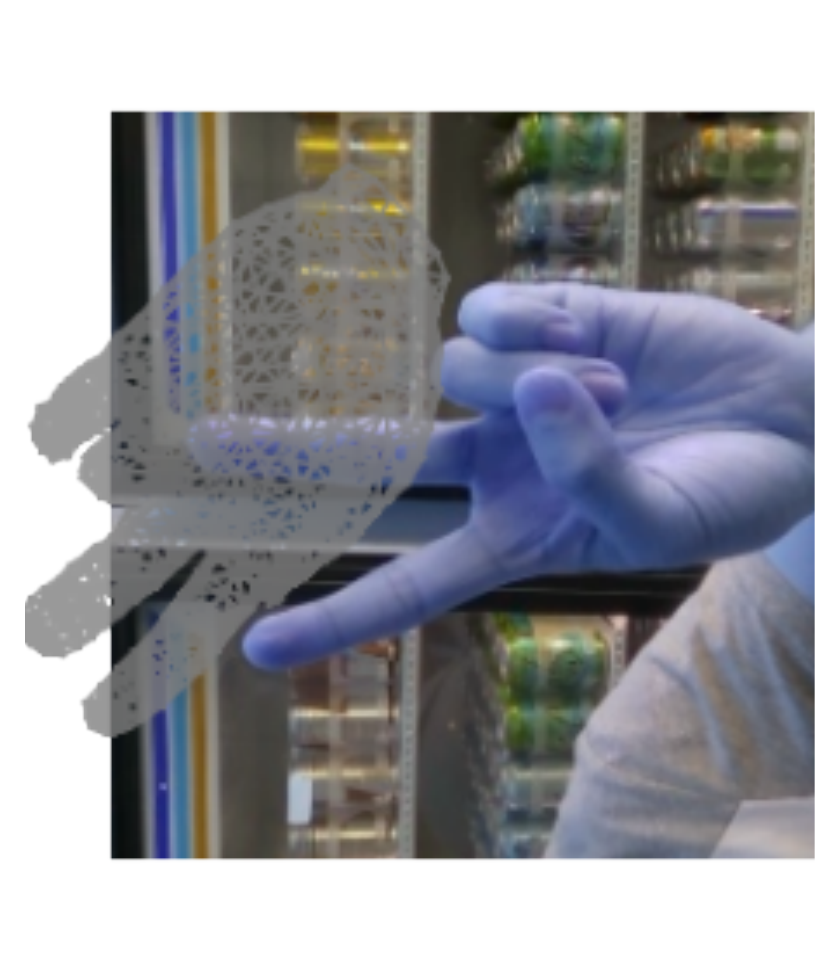}
    \centering
    \caption*{MLP Prediction}
  \end{subfigure}
  \begin{subfigure}{\rotwidth\linewidth}
    \includegraphics[width=\linewidth]{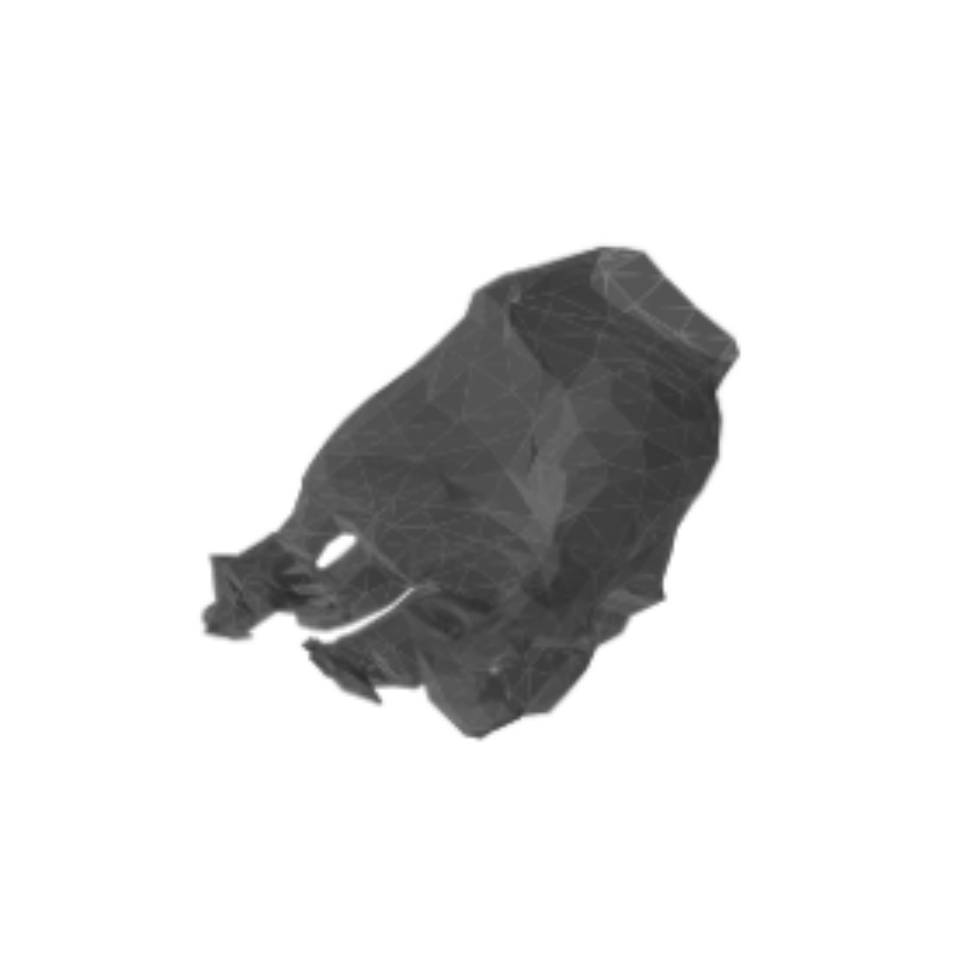}
    \centering
    \caption*{MLP View A}
  \end{subfigure}
  \begin{subfigure}{\rotwidth\linewidth}
    \includegraphics[width=\linewidth]{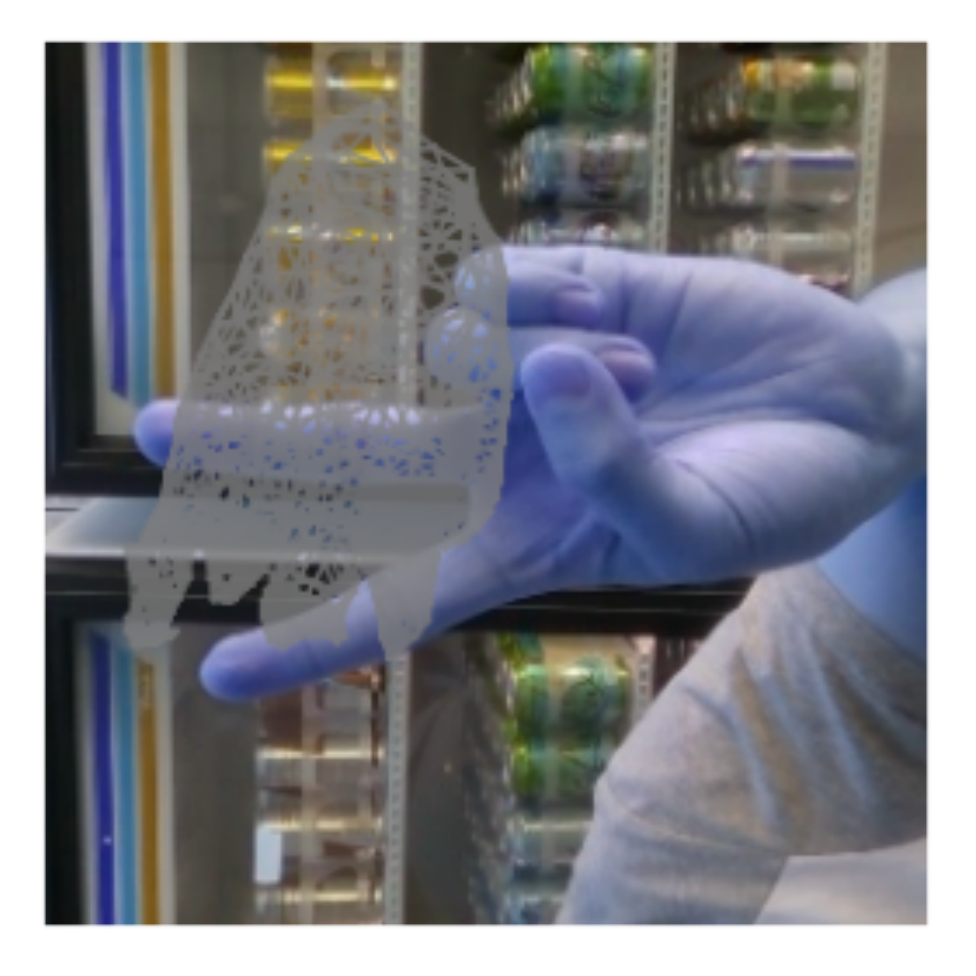}
    \centering
    \caption*{GNN Prediction}
  \end{subfigure}
  \begin{subfigure}{\rotwidth\linewidth}
    \includegraphics[width=\linewidth]{imgsnmeshs/gnn_rot/pred_mesh_5_1.pdf}
    \centering
    \caption*{GNN View A}
  \end{subfigure}
  
  \caption{Qualitative mesh reconstruction results on rotated validation hand images from the validation dataset for the EMLP, MLP, and GNN models.}
  \label{fig:rotmeshs}
\end{figure*}

Finally, we qualitatively validate our method by comparing the target mesh to the mesh predicted by our model in Figure~\ref{fig:meshs}. We present the image input into the model with both the target and predicted meshes projected onto the input image. Further, we show the mesh generated by our model for the image in two different view points. This shows that the generated meshes accurately capture the pose and shape required and the generated meshes are smooth. In addition to qualitative analysis on the validation dataset, we also compare how our model performs to the two non-rotation equivariant models with MLP and GNN decoders on a validation dataset where the hands have been rotated by $90^{\circ}$. As is demonstrated in Figure~\ref{fig:rotmeshs} our EMLP model generates meshes that accurately capture the pose and and shape of the hand. On the contrary, the MLP and GNN model fail to capture the pose correctly and do not predict realistic looking hand meshes in many cases. This result is an effect of the type of inductive bias that we chose to build into our model. Table~\ref{tab:rotations} shows that the GNN model can accurately reconstruct the meshes when the orientation remains fixed, although in reality this will not be the case and therefore the inductive bias of locality hurts the models ability to generalise. Further, the MLP model with no inductive bias appears to generate more realistic looking hands than the GNN model, despite generating hands of incorrect pose. Comparing both models to our EMLP model highlights the value of building useful inductive bias into the model, especially those that reflecting some known symmetry in the data. 

\section{Limitations}
\label{sec:limitations}
We use a discrete group of rotations, $C_{8}$, in the encoder, which means the encoder presented here is equivariant to rotations by $45^{\circ}$. This could be modified to use a larger group consisting of smaller rotations, although this would increase the feature space size. Ideally this would be modified to use a continuous rotation group, which although possible, would require a different choice of vector mapping function. However, we leave this to future work.

\section{Conclusion}
\label{sec:conc}
We present a novel framework for generating 3D hand meshes from 2D RGB images comprising of equivariant layers that ensure the entire model is rotation equivariant. To the best of our knowledge this is the first model for generating 3D meshes from 2D RGB images that considers equivariance. We provide a comparison between and MLP and GNN based decoder, showing that an MLP model generates smoother meshes than a GNN model. This improvement in mesh generation justifies our choice of using an MLP based model. Further, we demonstrate that rotation equivariance is a suitable inductive bias to build into each component of the model by outperforming other leading methods on a real-world dataset. In addition, this improves robustness by removing undesirable deformations of the generated meshes under rotation of the input image, which we both quantitatively and qualitatively compare with non-rotation equivariant models. The theoretical match of prior knowledge about the problem, and the improved empirical performance compared to the state of the art support the use of these new inductive biases in models generating 3D hand meshes from 2D RGB images.

\section{Acknowledgements}

Joshua Mitton is supported by a University of Glasgow Lord Kelvin Adam Smith Studentship. Chaitanya Kaul and Roderick Murray-Smith acknowledge funding from the QuantIC project funded by the EPSRC Quantum Technology Programme (grant EP/MO1326X/1) and the iCAIRD project, funded by Innovate UK (project number 104690). Roderick Murray-Smith acknowledges funding support from EPSRC grant EP/R018634/1, Closed-loop Data Science.

{\small
\bibliographystyle{ieee_fullname}
\bibliography{eqhandmesh}

\begin{thebibliography}{10}\itemsep=-1pt

\bibitem{baek2019pushing}
Seungryul Baek, Kwang~In Kim, and Tae-Kyun Kim.
\newblock {Pushing the envelope for RGB-based dense 3D hand pose estimation via
  neural rendering}.
\newblock In {\em Proceedings of the IEEE/CVF Conference on Computer Vision and
  Pattern Recognition}, pages 1067--1076, 2019.

\bibitem{bekkers2018roto}
Erik~J. Bekkers, Maxime~W. Lafarge, Mitko Veta, Koen~A.J. Eppenhof, Josien~P.W.
  Pluim, and Remco Duits.
\newblock Roto-translation covariant convolutional networks for medical image
  analysis.
\newblock In {\em International conference on medical image computing and
  computer-assisted intervention}, pages 440--448. Springer, 2018.

\bibitem{boukhayma20193d}
Adnane Boukhayma, Rodrigo~de Bem, and Philip~HS Torr.
\newblock 3d hand shape and pose from images in the wild.
\newblock In {\em Proceedings of the IEEE/CVF Conference on Computer Vision and
  Pattern Recognition}, pages 10843--10852, 2019.

\bibitem{bronstein2017geometric}
Michael~M Bronstein, Joan Bruna, Yann LeCun, Arthur Szlam, and Pierre
  Vandergheynst.
\newblock Geometric deep learning: going beyond euclidean data.
\newblock {\em IEEE Signal Processing Magazine}, 34(4):18--42, 2017.

\bibitem{cai2018weakly}
Yujun Cai, Liuhao Ge, Jianfei Cai, and Junsong Yuan.
\newblock {Weakly-supervised 3D hand pose estimation from monocular RGB
  images}.
\newblock In {\em Proceedings of the European Conference on Computer Vision
  (ECCV)}, pages 666--682, 2018.

\bibitem{chatzis2020comprehensive}
Theocharis Chatzis, Andreas Stergioulas, Dimitrios Konstantinidis, Kosmas
  Dimitropoulos, and Petros Daras.
\newblock A comprehensive study on deep learning-based 3d hand pose estimation
  methods.
\newblock {\em Applied Sciences}, 10(19):6850, 2020.

\bibitem{chen2020simple}
Ming Chen, Zhewei Wei, Zengfeng Huang, Bolin Ding, and Yaliang Li.
\newblock Simple and deep graph convolutional networks.
\newblock In {\em International Conference on Machine Learning}, pages
  1725--1735. PMLR, 2020.

\bibitem{cohen2018general}
Taco Cohen, Mario Geiger, and Maurice Weiler.
\newblock A general theory of equivariant {CNNs} on homogeneous spaces.
\newblock {\em arXiv preprint arXiv:1811.02017}, 2018.

\bibitem{cohen2019gauge}
Taco Cohen, Maurice Weiler, Berkay Kicanaoglu, and Max Welling.
\newblock {Gauge equivariant convolutional networks and the icosahedral CNN}.
\newblock In {\em International Conference on Machine Learning}, pages
  1321--1330. PMLR, 2019.

\bibitem{cohen2016group}
Taco Cohen and Max Welling.
\newblock Group equivariant convolutional networks.
\newblock In {\em International conference on machine learning}, pages
  2990--2999. PMLR, 2016.

\bibitem{cohen2018intertwiners}
Taco~S. Cohen, Mario Geiger, and Maurice Weiler.
\newblock Intertwiners between induced representations (with applications to
  the theory of equivariant neural networks).
\newblock {\em arXiv preprint arXiv:1803.10743}, 2018.

\bibitem{cohen2016steerable}
Taco~S. Cohen and Max Welling.
\newblock {Steerable CNNs}.
\newblock {\em International Conference on Learning Representations (ICLR)},
  2017.

\bibitem{defferrard2016convolutional}
Micha{\"e}l Defferrard, Xavier Bresson, and Pierre Vandergheynst.
\newblock Convolutional neural networks on graphs with fast localized spectral
  filtering.
\newblock {\em Advances in neural information processing systems},
  29:3844--3852, 2016.

\bibitem{dieleman2016exploiting}
Sander Dieleman, Jeffrey De~Fauw, and Koray Kavukcuoglu.
\newblock Exploiting cyclic symmetry in convolutional neural networks.
\newblock In {\em International conference on machine learning}, pages
  1889--1898. PMLR, 2016.

\bibitem{duvenaud2015convolutional}
David Duvenaud, Dougal Maclaurin, Jorge Aguilera-Iparraguirre, Rafael
  G{\'o}mez-Bombarelli, Timothy Hirzel, Al{\'a}n Aspuru-Guzik, and Ryan~P
  Adams.
\newblock Convolutional networks on graphs for learning molecular fingerprints.
\newblock {\em arXiv preprint arXiv:1509.09292}, 2015.

\bibitem{feng2019meshnet}
Yutong Feng, Yifan Feng, Haoxuan You, Xibin Zhao, and Yue Gao.
\newblock Meshnet: Mesh neural network for {3D} shape representation.
\newblock In {\em Proceedings of the AAAI Conference on Artificial
  Intelligence}, volume~33, pages 8279--8286, 2019.

\bibitem{finzi2020generalizing}
Marc Finzi, Samuel Stanton, Pavel Izmailov, and Andrew~Gordon Wilson.
\newblock Generalizing convolutional neural networks for equivariance to {Lie}
  groups on arbitrary continuous data.
\newblock In {\em International Conference on Machine Learning}, pages
  3165--3176. PMLR, 2020.

\bibitem{finzi2021practical}
Marc Finzi, Max Welling, and Andrew~Gordon Wilson.
\newblock A practical method for constructing equivariant multilayer
  perceptrons for arbitrary matrix groups.
\newblock {\em arXiv preprint arXiv:2104.09459}, 2021.

\bibitem{fuchs2020se3transformers}
Fabian~B. Fuchs, Daniel~E. Worrall, Volker Fischer, and Max Welling.
\newblock {SE(3)-Transformers: 3D Roto-Translation Equivariant Attention
  Networks}.
\newblock In {\em Advances in Neural Information Processing Systems 34
  (NeurIPS)}, 2020.

\bibitem{ge20193d}
Liuhao Ge, Zhou Ren, Yuncheng Li, Zehao Xue, Yingying Wang, Jianfei Cai, and
  Junsong Yuan.
\newblock {3D hand shape and pose estimation from a single RGB image}.
\newblock In {\em Proceedings of the IEEE/CVF Conference on Computer Vision and
  Pattern Recognition}, pages 10833--10842, 2019.

\bibitem{gilmer2017neural}
Justin Gilmer, Samuel~S. Schoenholz, Patrick~F. Riley, Oriol Vinyals, and
  George~E Dahl.
\newblock Neural message passing for quantum chemistry.
\newblock In {\em International conference on machine learning}, pages
  1263--1272. PMLR, 2017.

\bibitem{gori2005new}
Marco Gori, Gabriele Monfardini, and Franco Scarselli.
\newblock A new model for learning in graph domains.
\newblock In {\em Proceedings. 2005 IEEE International Joint Conference on
  Neural Networks, 2005.}, volume~2, pages 729--734. IEEE, 2005.

\bibitem{gupta2021rotation}
Deepak~K Gupta, Devanshu Arya, and Efstratios Gavves.
\newblock Rotation equivariant siamese networks for tracking.
\newblock In {\em Proceedings of the IEEE/CVF Conference on Computer Vision and
  Pattern Recognition}, pages 12362--12371, 2021.

\bibitem{hanocka2019meshcnn}
Rana Hanocka, Amir Hertz, Noa Fish, Raja Giryes, Shachar Fleishman, and Daniel
  Cohen-Or.
\newblock {MeshCNN}: a network with an edge.
\newblock {\em ACM Transactions on Graphics (TOG)}, 38(4):1--12, 2019.

\bibitem{hanocka2020point2mesh}
Rana Hanocka, Gal Metzer, Raja Giryes, and Daniel Cohen-Or.
\newblock Point2mesh: A self-prior for deformable meshes.
\newblock {\em arXiv preprint arXiv:2005.11084}, 2020.

\bibitem{hasson2019learning}
Yana Hasson, Gul Varol, Dimitrios Tzionas, Igor Kalevatykh, Michael~J Black,
  Ivan Laptev, and Cordelia Schmid.
\newblock Learning joint reconstruction of hands and manipulated objects.
\newblock In {\em Proceedings of the IEEE/CVF Conference on Computer Vision and
  Pattern Recognition}, pages 11807--11816, 2019.

\bibitem{he2016deep}
Kaiming He, Xiangyu Zhang, Shaoqing Ren, and Jian Sun.
\newblock Deep residual learning for image recognition.
\newblock In {\em Proceedings of the IEEE conference on computer vision and
  pattern recognition}, pages 770--778, 2016.

\bibitem{hertz2020pointgmm}
Amir Hertz, Rana Hanocka, Raja Giryes, and Daniel Cohen-Or.
\newblock {PointGMM: A neural GMM network for point clouds}.
\newblock In {\em Proceedings of the IEEE/CVF Conference on Computer Vision and
  Pattern Recognition}, pages 12054--12063, 2020.

\bibitem{hoogeboom2018hexaconv}
Emiel Hoogeboom, Jorn~W.T. Peters, Taco~S. Cohen, and Max Welling.
\newblock Hexaconv.
\newblock {\em International Conference on Learning Representations (ICLR)},
  2018.

\bibitem{ioffe2015batch}
Sergey Ioffe and Christian Szegedy.
\newblock Batch normalization: Accelerating deep network training by reducing
  internal covariate shift.
\newblock In {\em International conference on machine learning}, pages
  448--456. PMLR, 2015.

\bibitem{kipf2016semi}
Thomas~N Kipf and Max Welling.
\newblock Semi-supervised classification with graph convolutional networks.
\newblock {\em arXiv preprint arXiv:1609.02907}, 2016.

\bibitem{kohler2019equivariant}
{K{\"o}hler, Jonas and Klein, Leon and No{\'e}, Frank}.
\newblock {Equivariant flows: sampling configurations for multi-body systems
  with symmetric energies}.
\newblock {\em arXiv preprint arXiv:1910.00753}, 2019.

\bibitem{kondor2018generalization}
Risi Kondor and Shubhendu Trivedi.
\newblock On the generalization of equivariance and convolution in neural
  networks to the action of compact groups.
\newblock In {\em International Conference on Machine Learning}, pages
  2747--2755. PMLR, 2018.

\bibitem{krizhevsky2012imagenet}
Alex Krizhevsky, Ilya Sutskever, and Geoffrey~E. Hinton.
\newblock Imagenet classification with deep convolutional neural networks.
\newblock {\em Advances in neural information processing systems},
  25:1097--1105, 2012.

\bibitem{kulon2020weakly}
Dominik Kulon, Riza~Alp Guler, Iasonas Kokkinos, Michael~M Bronstein, and
  Stefanos Zafeiriou.
\newblock Weakly-supervised mesh-convolutional hand reconstruction in the wild.
\newblock In {\em Proceedings of the IEEE/CVF Conference on Computer Vision and
  Pattern Recognition}, pages 4990--5000, 2020.

\bibitem{kulon2019single}
Dominik Kulon, Haoyang Wang, Riza~Alp G{\"u}ler, Michael Bronstein, and
  Stefanos Zafeiriou.
\newblock Single image {3D} hand reconstruction with mesh convolutions.
\newblock {\em arXiv preprint arXiv:1905.01326}, 2019.

\bibitem{lang2020wigner}
Leon Lang and Maurice Weiler.
\newblock A {Wigner-Eckart} theorem for group equivariant convolution kernels.
\newblock {\em arXiv preprint arXiv:2010.10952}, 2020.

\bibitem{lecun1989backpropagation}
Yann LeCun, Bernhard Boser, John~S. Denker, Donnie Henderson, Richard~E.
  Howard, Wayne Hubbard, and Lawrence~D. Jackel.
\newblock Backpropagation applied to handwritten zip code recognition.
\newblock {\em Neural computation}, 1(4):541--551, 1989.

\bibitem{li2019point}
Shile Li and Dongheui Lee.
\newblock Point-to-pose voting based hand pose estimation using residual
  permutation equivariant layer.
\newblock In {\em Proceedings of the IEEE/CVF Conference on Computer Vision and
  Pattern Recognition}, pages 11927--11936, 2019.

\bibitem{li2015gated}
Yujia Li, Daniel Tarlow, Marc Brockschmidt, and Richard Zemel.
\newblock Gated graph sequence neural networks.
\newblock {\em arXiv preprint arXiv:1511.05493}, 2015.

\bibitem{mitton2021rotation}
Joshua Mitton and Roderick Murray-Smith.
\newblock Rotation equivariant deforestation segmentation and driver
  classification.
\newblock {\em arXiv preprint arXiv:2110.13097}, 2021.

\bibitem{mitton2021graph}
Joshua Mitton, Hans~M Senn, Klaas Wynne, and Roderick Murray-Smith.
\newblock A graph vae and graph transformer approach to generating molecular
  graphs.
\newblock {\em arXiv preprint arXiv:2104.04345}, 2021.

\bibitem{morris2019weisfeiler}
Christopher Morris, Martin Ritzert, Matthias Fey, William~L Hamilton, Jan~Eric
  Lenssen, Gaurav Rattan, and Martin Grohe.
\newblock Weisfeiler and leman go neural: Higher-order graph neural networks.
\newblock In {\em Proceedings of the AAAI Conference on Artificial
  Intelligence}, volume~33, pages 4602--4609, 2019.

\bibitem{nair2010rectified}
Vinod Nair and Geoffrey~E Hinton.
\newblock {Rectified Linear Units improve Restricted Boltzmann machines}.
\newblock In {\em ICML}, 2010.

\bibitem{ravanbakhsh2017equivariance}
Siamak Ravanbakhsh, Jeff Schneider, and Barnabas Poczos.
\newblock Equivariance through parameter-sharing.
\newblock In {\em International Conference on Machine Learning}, pages
  2892--2901. PMLR, 2017.

\bibitem{romero2017embodied}
Javier Romero, Dimitrios Tzionas, and Michael~J Black.
\newblock Embodied hands: Modeling and capturing hands and bodies together.
\newblock {\em ACM Transactions on Graphics (ToG)}, 36(6):1--17, 2017.

\bibitem{satorras2021n}
Victor~Garcia Satorras, Emiel Hoogeboom, and Max Welling.
\newblock E(n) equivariant graph neural networks.
\newblock {\em arXiv preprint arXiv:2102.09844}, 2021.

\bibitem{scarselli2008graph}
Franco Scarselli, Marco Gori, Ah~Chung Tsoi, Markus Hagenbuchner, and Gabriele
  Monfardini.
\newblock The graph neural network model.
\newblock {\em IEEE transactions on neural networks}, 20(1):61--80, 2008.

\bibitem{schutt2017schnet}
{Sch{\"u}tt, Kristof T and Kindermans, Pieter-Jan and Sauceda, Huziel E and
  Chmiela, Stefan and Tkatchenko, Alexandre and M{\"u}ller, Klaus-Robert}.
\newblock {Schnet: A continuous-filter convolutional neural network for
  modeling quantum interactions}.
\newblock {\em arXiv preprint arXiv:1706.08566}, 2017.

\bibitem{simonovsky2017dynamic}
Martin Simonovsky and Nikos Komodakis.
\newblock Dynamic edge-conditioned filters in convolutional neural networks on
  graphs.
\newblock In {\em Proceedings of the IEEE conference on computer vision and
  pattern recognition}, pages 3693--3702, 2017.

\bibitem{thomas2018tensor}
Nathaniel Thomas, Tess Smidt, Steven Kearnes, Lusann Yang, Li Li, Kai Kohlhoff,
  and Patrick Riley.
\newblock {Tensor field networks: Rotation-and translation-equivariant neural
  networks for 3D point clouds}.
\newblock {\em arXiv preprint arXiv:1802.08219}, 2018.

\bibitem{veeling2018rotation}
Bastiaan~S Veeling, Jasper Linmans, Jim Winkens, Taco Cohen, and Max Welling.
\newblock Rotation equivariant {CNNs} for digital pathology.
\newblock In {\em International Conference on Medical image computing and
  computer-assisted intervention}, pages 210--218. Springer, 2018.

\bibitem{velivckovic2017graph}
Petar Veli{\v{c}}kovi{\'c}, Guillem Cucurull, Arantxa Casanova, Adriana Romero,
  Pietro Lio, and Yoshua Bengio.
\newblock Graph attention networks.
\newblock {\em arXiv preprint arXiv:1710.10903}, 2017.

\bibitem{weiler2019general}
Maurice Weiler and Gabriele Cesa.
\newblock {General $ E (2) $-Equivariant Steerable CNNs}.
\newblock {\em arXiv preprint arXiv:1911.08251}, 2019.

\bibitem{weiler20183d}
Maurice Weiler, Mario Geiger, Max Welling, Wouter Boomsma, and Taco Cohen.
\newblock {3D steerable CNNs: Learning rotationally equivariant features in
  volumetric data}.
\newblock {\em Conference on Neural Information Processing Systems (NeurIPS)},
  2018.

\bibitem{weiler2018learning}
Maurice Weiler, Fred~A. Hamprecht, and Martin Storath.
\newblock {Learning steerable filters for rotation equivariant CNNs}.
\newblock In {\em Proceedings of the IEEE Conference on Computer Vision and
  Pattern Recognition}, pages 849--858, 2018.

\bibitem{wu2019pointconv}
Wenxuan Wu, Zhongang Qi, and Li Fuxin.
\newblock {PointConv}: {Deep} convolutional networks on {3D} point clouds.
\newblock In {\em Proceedings of the IEEE/CVF Conference on Computer Vision and
  Pattern Recognition}, pages 9621--9630, 2019.

\bibitem{xu2018powerful}
Keyulu Xu, Weihua Hu, Jure Leskovec, and Stefanie Jegelka.
\newblock How powerful are graph neural networks?
\newblock {\em arXiv preprint arXiv:1810.00826}, 2018.

\bibitem{zhang2019end}
Xiong Zhang, Qiang Li, Hong Mo, Wenbo Zhang, and Wen Zheng.
\newblock End-to-end hand mesh recovery from a monocular rgb image.
\newblock In {\em Proceedings of the IEEE/CVF International Conference on
  Computer Vision}, pages 2354--2364, 2019.

\bibitem{zhao2021point}
Hengshuang Zhao, Li Jiang, Jiaya Jia, Philip~HS Torr, and Vladlen Koltun.
\newblock Point transformer.
\newblock In {\em Proceedings of the IEEE/CVF International Conference on
  Computer Vision}, pages 16259--16268, 2021.

\end{thebibliography}
}

\end{document}